\documentclass[runningheads]{llncs}
\usepackage{graphicx}
\usepackage{multirow}
\begin{document}

\title{Progressive Feature Fusion Network for Realistic Image Dehazing} 
\titlerunning{PFFNet for Realistic Image Dehazing} 


\author{Kangfu Mei\inst{1}\orcidID{0000-0001-8949-9597} \and
Aiwen Jiang \thanks{Corresponding author} \inst{1}\orcidID{0000-0002-5979-7590} \and \\
Juncheng Li\inst{2}\orcidID{0000-0001-7314-6754} \and
Mingwen Wang\inst{1}}
%

\authorrunning{K. Mei, A. Jiang, et al.} 


\institute{School of Computer Information Engineering, Jiangxi Normal University\\
\email{\{meikangfu, jiangaiwen, jhye, mwwang\}@jxnu.edu.cn}\\
\and
Department of Computer Science \& Technology, East China Normal University\\
\email{51164500049@stu.ecnu.edu.cn}}

\maketitle

\begin{abstract}
Single image dehazing is a challenging ill-posed restoration problem. Various prior-based and learning-based methods have been proposed. Most of them follow a classic atmospheric scattering model which is an elegant simplified physical model based on the assumption of single-scattering and homogeneous atmospheric medium. The formulation of haze in realistic environment is more complicated. In this paper, we propose to take its essential mechanism as "black box", and focus on learning an input-adaptive trainable end-to-end dehazing model. An U-Net like encoder-decoder deep network via progressive feature fusions has been proposed to directly learn highly nonlinear transformation function from observed hazy image to haze-free ground-truth. The proposed network is evaluated on two public image dehazing benchmarks. The experiments demonstrate that it can achieve superior performance when compared with popular state-of-the-art methods. With efficient GPU memory usage, it can satisfactorily recover ultra high definition hazed image up to 4K resolution, which is unaffordable by many deep learning based dehazing algorithms.
\end{abstract}
\section{Introduction}

\begin{figure}[!h]
\centering
\includegraphics[width=0.8\textwidth]{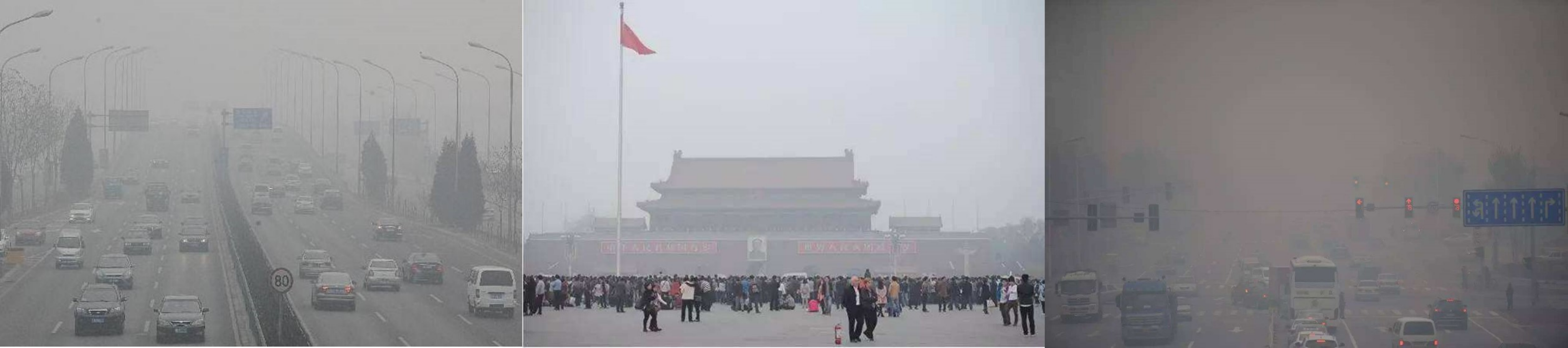}
\caption{Examples of realistic hazy images}
\label{fig:hazyexample}
\end{figure}

Haze is a common atmospheric phenomena produced by small floating particles such as dust and smoke in the air. These floating particles absorb and scatter the light greatly, resulting in degradations on image quality. Under severe hazy conditions, many practical applications such as video surveillance, remote sensing, autonomous driving etc are easily put in jeopardy, as shown in Figure~\ref{fig:hazyexample}. High-level computer vision tasks like detection and recognition are hardly to be completed. Therefore, image dehazing (a.k.a haze removal) becomes an increasingly desirable technique.

Being an ill-posed restoration problem, image dehazing is a very challenging task. Similar to other ill-posed problem like super-resolution, earlier image dehazing methods assumed the availability of multiple images from the same scene. However, in practical settings, dehazing from single image is more realistic and gains more dominant popularity\cite{Fattal2008}. Therefore, in this paper, we focus on the problem of single image dehazing.

Most state-of-the-art single image dehazing methods\cite{DCP:hekaiming2011}\cite{colorprior:dehaze}\cite{Menggaofeng2013}\cite{Codruta2013}\cite{NLD2016:dehaze}\cite{DehazeNet}\cite{mscnn:dehaze} are based on a classic atmospheric scattering model\cite{scattermodel1978} which is formulated as following Equation \ref{eq:scattemodel}:
\begin{equation}
\label{eq:scattemodel}
I\left( x \right) = J\left( x \right)t\left( x \right) + A \cdot \left( {1 - t\left( x \right)} \right)
\end{equation}
where, $I\left( x \right)$ is the observed hazy image, $J\left( x \right)$ is the clear image. $t\left( x \right)$ is called medium transmission function. $A$ is the global atmospheric light. $x$ represents pixel locations.

The physical model explained the degradations of a hazy image. The medium transmission function $t\left( x \right) = {e^{ - \beta  \cdot d\left( x \right)}}$ is a distance dependent factor that reflects the fraction of light reaching camera sensor. The atmospheric light $A$ indicates the intensity of ambient light. It is not difficult to find that haze essentially brings in non-uniform, signal-dependent noise, as the scene attenuation caused by haze is correlated with the physical distance between object's surface and the camera.

Apart from a few works that focused on estimating the atmospheric light\cite{atomLight1:2014}, most of popular algorithms concentrate more on accurately estimation of transmission function $t\left( x \right)$ with either prior knowledge or data-driven learning. Based on the estimated  $\hat t\left( x \right)$ and $\hat A$, the clear image $\hat J$ is then recovered by using following Equation~\ref{eq:recover} .
\begin{equation}
\label{eq:recover}
J\left( x \right) = \frac{{I\left( x \right) - \hat A \cdot \left( {1 - \hat t\left( x \right)} \right)}}{{\hat t\left( x \right)}} = \frac{1}{{\hat t\left( x \right)}}I\left( x \right) - \frac{{\hat A}}{{\hat t\left( x \right)}} + \hat A
\end{equation}

Though tremendous improvements have been made, as we know, the traditional separate pipeline does not directly measure the objective reconstruction errors. The inaccuracies resulted from both transmission function and atmospheric light estimation would potentially amplify each other and hinder the overall dehazing performance.

The recently proposed AOD-Net\cite{ADONet2017} was the first end-to-end trainable image dehazing model. It reformulated a new atmospheric scattering model from the classic one by leveraging a linear transformation to integrate both the transmission function and the atmospheric light into an unified map $K\left( x \right)$, as shown in Equation~\ref{eq:reformua}.
\begin{equation}
\label{eq:reformua}
J\left( x \right) = K\left( x \right)I\left( x \right) - K\left( x \right) + b
\end{equation}
where the $K\left( x \right)$ was an input-dependent transmission function. A light-weight CNN was built to estimate the $K\left( x \right)$ map, and jointly trained to further minimize the reconstruction error between the recovered output $J\left( x \right)$ and the ground-truth clear image.

Going deeper, we consider the general relationship between observed input $I$ and recovered output $J$ as $J\left( x \right) = \Phi \left( {I\left( x \right);\theta } \right)$, where $\Phi \left(  *  \right)$ represents some potential highly nonlinear transformation function whose parameters set is $\theta$. Then the relationship represented by AOD-Net could be viewed as a specific case of the general function $\Phi$.

In this paper, we argue that the formation of hazy image has complicated mechanism, and the classic atmospheric scattering model\cite{scattermodel1978} is just an elegant simplified physical model based on the assumption of single-scattering and homogeneous atmospheric medium. There potentially exists some highly nonlinear transformation between the hazy image and its haze-free ground-truth. With that in mind, instead of limitedly learning the intermediate transmission function or its reformulated one from classic scattering model as AOD-Net did, we propose to build a real complete end-to-end deep network from an observed hazy image $I$ to its recovered clear image $J$. To avoid making efforts on find "real" intermediate physical model, our strict end-to-end network pay much concerns on the qualities of dehazed output.

We employ an encoder-decoder architecture similar to the U-Net\cite{U-Net2015} to directly learn the input-adaptive restoration model $\Phi$. The encoder convolves input image into several successive spatial pyramid layers. The decoder then successively recovers image details from the encoded feature mappings. In order to make full use of input information and accurately estimate structural details, progressive feature fusions are performed on different level mappings between encoder and decoder. We evaluate our proposed network on two public image dehazing benchmarks. The experimental results have shown that our method can achieve great improvements on final restoration performance, when compared with several state-of-the-art methods.

The contributions of this paper are two-fold:\\
\begin{itemize}
  \item We have proposed an effective trainable U-Net like end-to-end network for image dehazing. The encoder-decoder architecture via progressive feature fusion directly learns the input-adaptive restoration model. The essential formulation mechanism of a hazy image is taken as "black box", and efforts are made on restoring the final high quality, clear output. At this viewpoint, our proposed network is in a real sense the first end-to-end deep learning based image dehazing model.

  \item Our proposed network can directly process ultra high-definition realistic hazed image up to 4K resolution with superior restoration performance at a reasonable speed and memory usage. Many popular deep learning based image dehazing network can not afford image of such high resolution on a single TITAN X GPU. We owe our advantage to the effective encoder-decoder architecture.
\end{itemize}

\section{Related work}
Single image dehazing is a very challenging ill-posed problem. In the past, various prior-based and learning-based methods have been developed to solve the problem. On basis of the classic atmospheric scattering model proposed by Cantor\cite{scattermodel1978}, most of image dehazing methods followed a three-step pipeline: (a) estimating transmission map $t\left(x\right)$; (b) estimating global atmospheric light $A$; (c) recovering the clear image $J$ via computing Equation~\ref{eq:recover}. In this section, we would focus on some representative methods. More related works can be referred to surveys~\cite{review1:zidonghuaxuebao}\cite{review2:IEEEAccess}\cite{Youshaodi2017}.

A milestone work was the effective dark channel prior (DCP) proposed by He Kaiming et al.\cite{DCP:hekaiming2011} for outdoor images. They discovered that the local minimum of the dark channel of a haze-free image was close to zero. Base on the prior, transmission map could be reliably calculated. Zhu et al\cite{colorprior:dehaze} proposed a color attenuation prior by observing that the concentration of the haze was positively correlated with the difference between the brightness and the saturation. They created a linear model of scene depth for the hazy image. Based the recovered depth information, a transmission map was well estimated for haze removal. Dana et al\cite{NLD2016:dehaze} proposed a non-local prior that colors of a haze-free image could be well approximated by a few hundred distinct color clusters in RGB space. On assumption that each of these color clusters became a line in the presence of haze, they recovered both the distance map and the haze-free image.

With the success of convolutional neural network in computer vision area, several recent dehazing algorithms directly learn transmission map $t\left(x\right)$ fully from data, in order to avoid inaccurate estimation of physical parameters from a single image. Cai et al. \cite{DehazeNet} proposed a DehazeNet, an end-to-end CNN network for estimating the transmission with a novel BReLU unit. Ren et al. \cite{mscnn:dehaze} proposed a multi-scale deep neural network to estimate the transmission map. The recent AOD-Net\cite{ADONet2017} introduced a newly defined transmission variable to integrate both classic transmission map and atmospheric light. As AOD-Net needed learn the new intermediate transmission map, it still fell into a physical model. The latest proposed Gated Fusion Network(GFN)\cite{GFN2018:dehaze} learned confidence maps to combine several derived input images into a single one by keeping only the most significant features of them. We should note that, for GFN, handcrafted inputs were needed to be specifically derived for fusion and intermediate confidence maps were needed to be estimated. In contrast, our proposed network directly learns the transformation from input hazy image to output dehazed image, needn't learning any specific intermediate maps.

\section{Progressive Feature Fusion Network (PFFNet) for Image Dehazing}
\begin{figure}
\centering
\includegraphics[width=\textwidth]{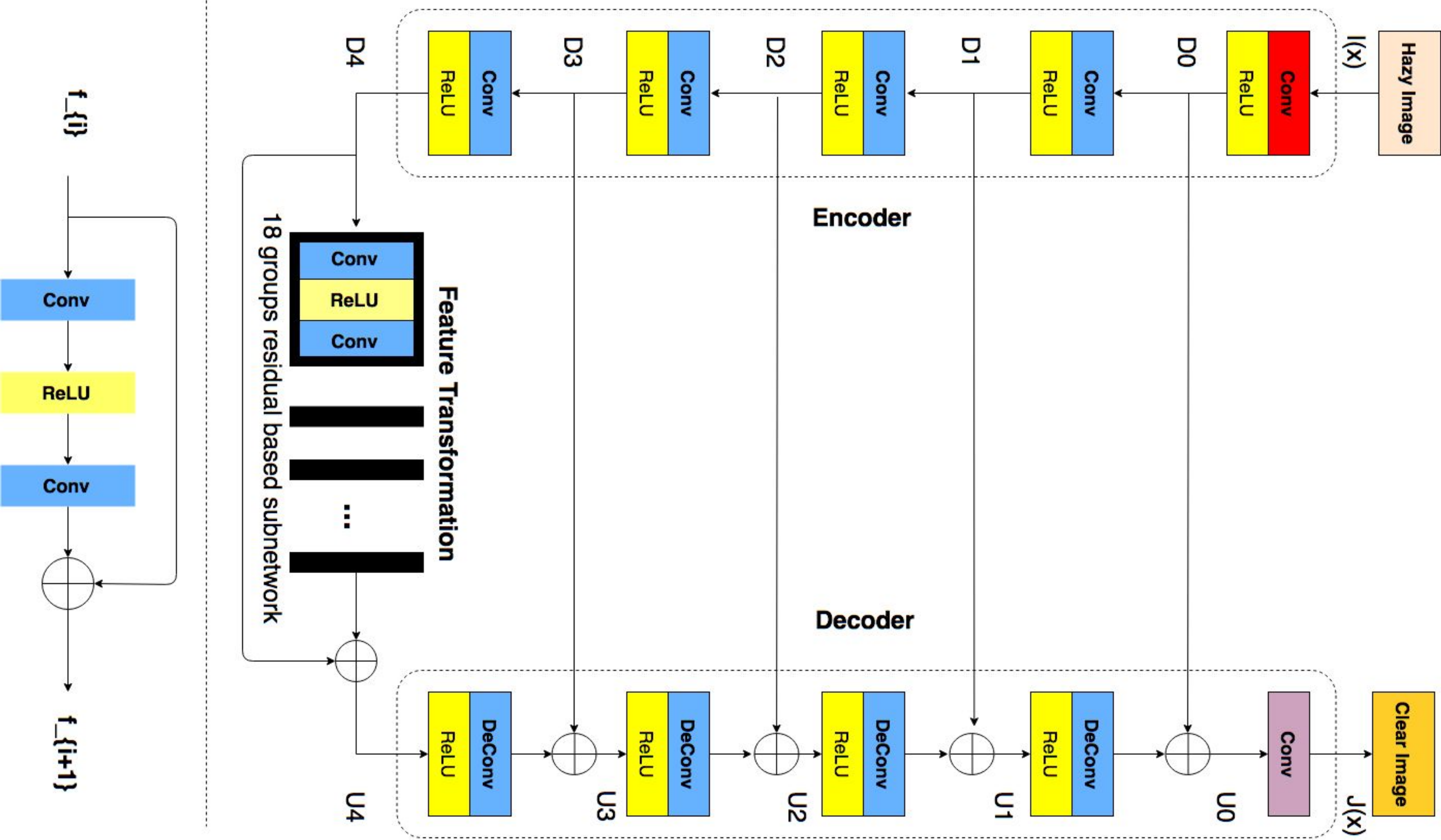}
\caption{The architecture of the progressive feature fusion network for image dehazing}
\label{fig:PFFNet}
\end{figure}

In this section, we will describe our proposed end-to-end image dehazing network in details. The architecture of our progressive feature fusion network is illustrated in Figure~\ref{fig:PFFNet}. It consists of three modules: encoder, feature transformation and decoder.

The encoder module consists of five convolution layers, each of which is followed by a nonlinear ReLU activation. For the convenience of description, we denote the $i$-{th} "conv+relu" layer as $Conv_{en}^i, i = \left\{ {0,1,2,3,4} \right\}$. The first layer $Conv_{en}^0$ is for aggregating informative features on a relatively large local receptive field from original observed hazy image $I$. The following four layers then sequentially perform down-sampling convolutional operations to encode image's information in pyramid scale.
\begin{equation}
{D_i} = Conv_{en}^i\left( {{D_{i - 1}}} \right),i = \left\{ {0,1,2,3,4} \right\},{\rm{where, }}{D_{ - 1}} = I
\end{equation}

We denote $k_i, s_i, c_i$ as the receptive field size, step size, and output channels of layer $Conv_{en}^i$ respectively.
In this paper, empirically, for $conv_{en}^0$, we set $k_0=11, s_0=1, c_i=16$. Consequently, the corresponding output $D_0$ keeps the same spatial size as input $I$. For $conv_{en}^i,i=\left\{ {1,2,3,4} \right\}$ layers, we keep their receptive field size and step size the same. And each one learns feature mappings with double channels more than its previous layer. The super-parameters are set $k_i=3, s_i=2, {c_i} = 2{c_{i - 1}},i = \left\{ {1,2,3,4} \right\}$. As a result, we can easily calculate that if the size of input hazy image is $w \times h \times c$, the size of the output of encoder $D_4$ is consequently ${\textstyle{1 \over {16}}}w \times {\textstyle{1 \over {16}}}h \times 256$, where $w,h,c$ are image width, image height, image channels in respective.
That means if an input image is with 4K resolution level, the resulted feature map is with 256 spatial resolution level after encoder module, which benefits greatly the following processing stages for reducing memory usage.

The feature transformation module denoted as $\Psi\left(*\right)$ consists of residual based subnetworks. As we know, the main benefit of a very deep network is that it can represent very complex functions and also learn features at many different levels of abstraction. However, traditional deep networks often suffer gradient vanish or expansion disaster. The popular residual networks \cite{resnet:2016}\cite{wideresnet:2016} explicitly reformulated network layers as learning residual functions with reference to the layer inputs, instead of learning unreferenced functions. It allows training much deeper networks than were previously practically feasible. Therefore, to balance between computation efficiency and GPU memory usage, in this module, we empirically employed eighteen wide residual blocks for feature learning.

Let $B\left(M\right)$ denotes the structure of a residual block, where $M$ lists the kernel sizes of the convolutional layers in a block. In this paper, we accept $B\left(3,3\right)$ as the basic residual block, as shown in the left part of Figure~\ref{fig:PFFNet}. The channels of convolution layer are all 256, which are the same as the channels of the feature map $D_4$ from encoder module. The step size is constantly kept to be 1.

The decoder module consists of four deconvolution layers followed by a convolution layer. In opposite to encoder, the deconvolution layers of decoder are sequentially to recover image structural details. Similarly, we denote the $j$-{th} "relu+deconv" layer as $DeConv_{dec}^j,j = \left\{ {4,3,2,1} \right\}$.
\begin{equation}
{F_{j - 1}} = DeConv_{dec}^j\left( {{U_j}} \right),j = \left\{ {4,3,2,1} \right\}
\end{equation}
where, $U_j$ is an intermediate feature map.

Through deconvolution (a.k.a transposed) layer, the $DeConv_{dec}^j$ performs up-sample operations to obtain intermediate feature mappings with double spatial size and half channels than its previous counterpart. Concretely, the receptive field size, step size and output channels are set $k_j=3, s_j=2, {c_{j - 1}} = {\textstyle{1 \over 2}}{c_j}, j = \left\{ {4,3,2,1} \right\}$. It is not difficult to find that, in our network setting, the output map $F_{j}$ from $DeConv_{dec}^j$ enjoys the same feature dimensions as corresponding input $D_i$ of $Conv_{en}^i$ has, when $i=j \in \left\{ {3,2,1,0} \right\}$.

In order to maximize information flow along multi-level layers and guarantee better convergence, skip connections are employed between corresponding layers of different level from encoder and decoder. A global shortcut connection is applied between input and output of the feature transformation module, as shown in Figure~\ref{fig:PFFNet}.
\begin{equation}
\begin{array}{l}
{U_i} = {D_i} \oplus {F_i},i = \left\{ {3,2,1,0} \right\}\\
{U_4} = {D_4} \oplus {\mathop{\Psi}\nolimits} \left( {{D_4}} \right)
\end{array}
\end{equation}
where $\oplus$ is an channel-wise addition operator.

The dimension of the transposed feature map $U_0$ is therefore $w \times h \times 16$, as the same as $D_0$. A convolution operation is further applied on $U_0$, and generates the final recovered clear image $J$. Herein, for this convolution layer, the kernel size $k$ is 3; the step size is 1; and the channels is the same as $J$.

The proposed image dehazing network progressively performs feature fusion on spatial pyramid mappings between encoder and decoder, which enables maximally preserved structural details from inputs for deconvolution layers, and further makes the dehazing network more input-adaptive.

\section{Experiments}
\subsection{Dataset}
We evaluate the effectiveness of our proposed method on two public dehazing benchmarks. The source code is available on GitHub\footnote{source code: \url{https://github.com/MKFMIKU/PFFNet}}.
\subsubsection{NTIRE2018 Image Dehazing Dataset}  The dataset was distributed by NTIRE 2018 Challenge on image dehazing\cite{NTIRE2018}. Two novel subsets (I-HAZE\cite{I-HAZE} and O-HAZE\cite{O-HAZE}) with real haze and their ground-truth haze-free images were included. Hazy images were both captured in presence of real haze generated by professional haze machines. The I-HAZE dataset contains 35 scenes that correspond to indoor domestic environments, with objects of different colors and speculates. The O-HAZE contains 45 different outdoor scenes depicting the same visual content recorded in haze-free and hazy conditions, under the same illumination parameters. All images are ultra high definition images on 4K resolution level.

\subsubsection{RESIDE\cite{RESIDE:2018}} The REISDE is a large scale synthetic hazy image dataset. The training set contains 13990 synthetic hazy images generated by using images from existing indoor depth datasets such as NYU2\cite{NYU2012} and Middlebury\cite{Middlebury2003}. Specifically, given a clear image $J$, random atmospheric lights $A \in \left[ {0.7,1.0} \right]$ for each channel, and the corresponding ground-truth depth map $d$, function $t\left( x \right) = {e^{ - \beta  \cdot d\left( x \right)}}$ is applied to synthesize transmission map first, then a hazy image is generated by using the physical model in Equation~(\ref{eq:scattemodel}) with randomly selected scattering coefficient $\beta  \in \left[ {0.6,1.8} \right]$. In RESIDE dataset, images are on ${\rm{620}} \times {\rm{460}}$ resolution level.

The \emph{Synthetic Objective Testing Set} (SOTS) of RESIDE is used as our test dataset. The SOTS contains 500 indoor images from NYU2\cite{NYU2012} (non-overlapping with training images), and follows the same process as training data to synthesize hazy images.

\subsection{Comparisons and Analysis}
Several representative state-of-the-art methods are compared in our experiment: Dark-Channel Prior (DCP)\cite{DCP:hekaiming2011}, Color Attenuation Prior (CAP)\cite{colorprior:dehaze}, Non-Local Dehazing(NLD)\cite{NLD2016:dehaze}, DehazeNet\cite{DehazeNet}, Multi-scale CNN(MSCNN)\cite{mscnn:dehaze}, AOD-Net\cite{ADONet2017}, and Gated Fusion Network(GFN)\cite{GFN2018:dehaze}. The popular full-reference PSNR and SSIM metrics are accepted to evaluate the dehazing performance.

\subsubsection{Training details} As our PFF-Net was initially proposed to take part in the NTIRE2018 challenge on image dehazing, the GPU memory usage of our network is efficient so that we can directly recover an ultra high definition realistic hazy image on 4K resolution level on a single TITAN X GPU.

In this paper, we train our network both on I-HAZE and O-HAZE training images, which has 80 scenes in all. Based on these scenes, we further perform data augmentation for training. We first use sliding window to extract image crops of $520 \times 520$ size from the realistic hazy images. The stride is 260 pixels. For each image crop, we obtain its 12 variants at four angles $\left\{ {0,\frac{\pi }{2},\pi ,\frac{3}{2}\pi } \right\}$ and three mirror flip cases $\left\{ No Flip, Horizontal Flip, Vertical Flip\right \}$. In consequence, about 190K patches are augmented as the training dataset.

\begin{figure}
\begin{center}
\begin{tabular}{cc}
\includegraphics[width=0.48\textwidth]{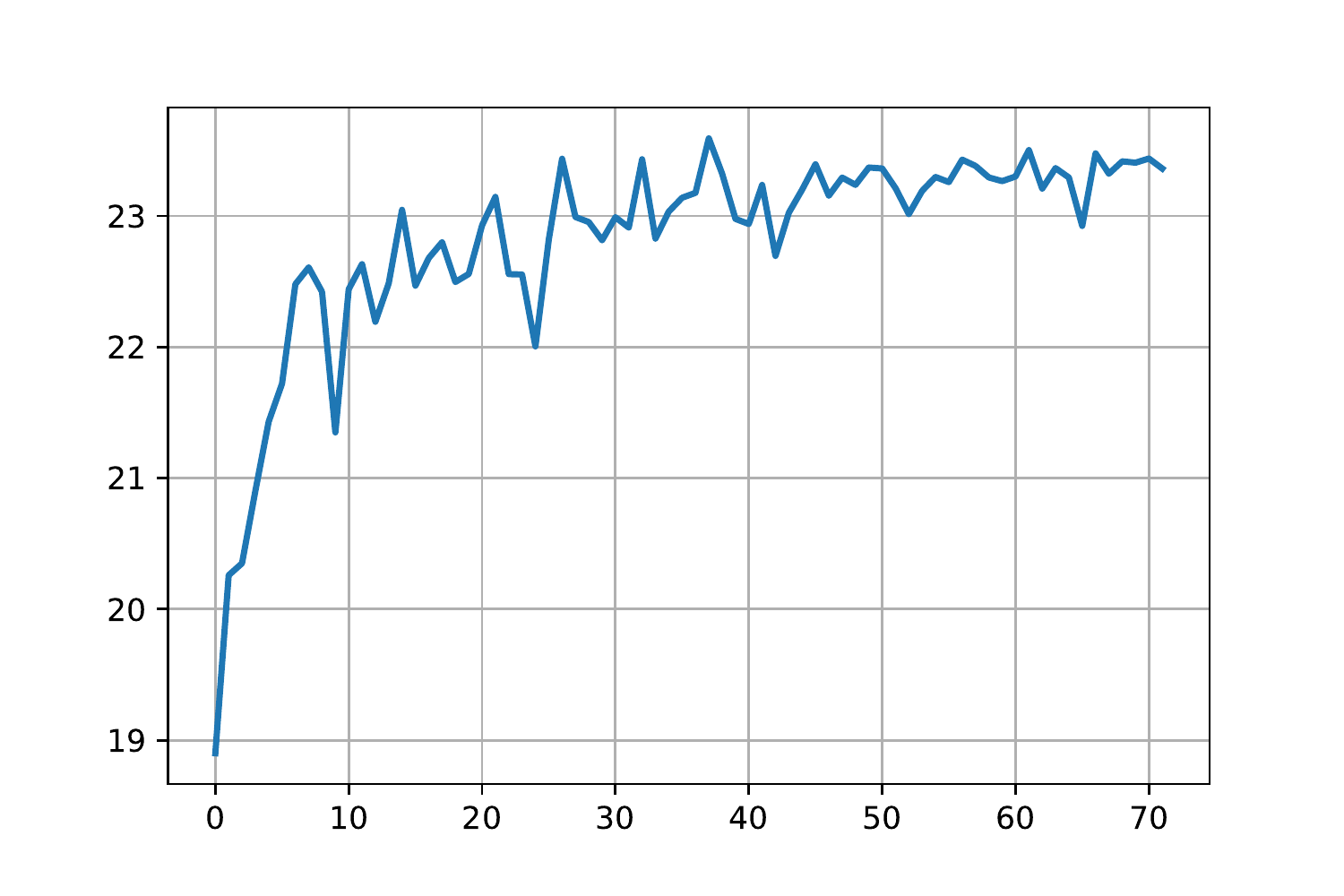} &
\includegraphics[width=0.48\textwidth]{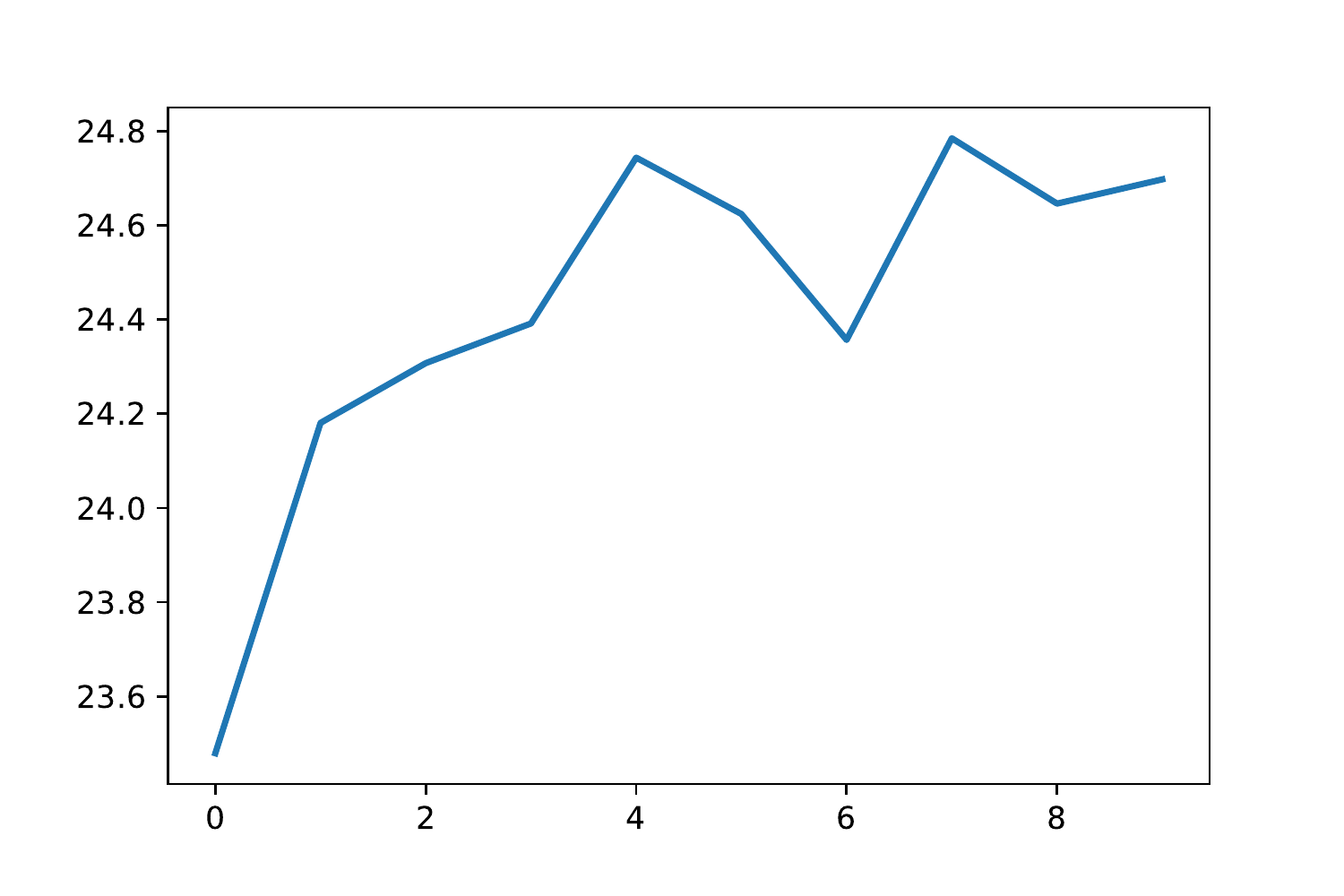} \\
NTIRE2018 & RESIDE\\
\end{tabular}
\caption{The testing curves of our proposed PFF-Net on NTIRE2018 (\emph{Left}) and RESIDE (\emph{Right}). In both sub-figures, the horizontal axis shows training epoches. The vertical axis shows PSNR performance tested on training model at corresponding epoch.}
\label{fig:traincurve}
\end{center}
\end{figure}

The ADAM\cite{Adam2015} is used as the optimizer. The initial learning rate we set is $\eta  = 0.0001$, and kept a constant during training. Mean Square Errors (MSE) between recovered clear image and haze-free ground-truth is taken as our objective loss. The batch-size is 32. During training, we recorded every 2000 iterations as an epoch and the total num of training epoches is empirically 72 in practice. The testing curve on PSNR performance is shown in Figure~\ref{fig:traincurve}. We found that the network started to converge at the last 10 epoches.

\subsubsection{Ablation parameter comparisons on networks settings} Before fixing the architecture of our PFFNet in this paper, we have done several ablation experiments on parematers setting. We have experimented four different blocks sizes in feature transformation module: $\{6, 12, 18, 24\}$. The testing performances are shown in following Figure~\ref{fig:ablation}-Left. Increasing the size of residual blocks would improve the testing performance. By considering the balance between the performance and the available computing resources, we finally adopted the feature transformation module with 18 residual blocks in this paper.

We have also compared the performance of networks with / without skip connections between encoder module and encoder module. The comparisons was experimented through training our network with 12 residual blocks in feature transformation module. In terms of the speed of convergence and the performance, network without skip connections is much worse than network with skip connections, as shown in Figure~\ref{fig:ablation}-Right. The conclusion is consistent with observations in many residual based learning models and also validates the necessaries of progressive feature fusion between encoder and decoder stages.
\begin{figure}
\centering
\begin{tabular}{cc}
\includegraphics[width=0.48\textwidth]{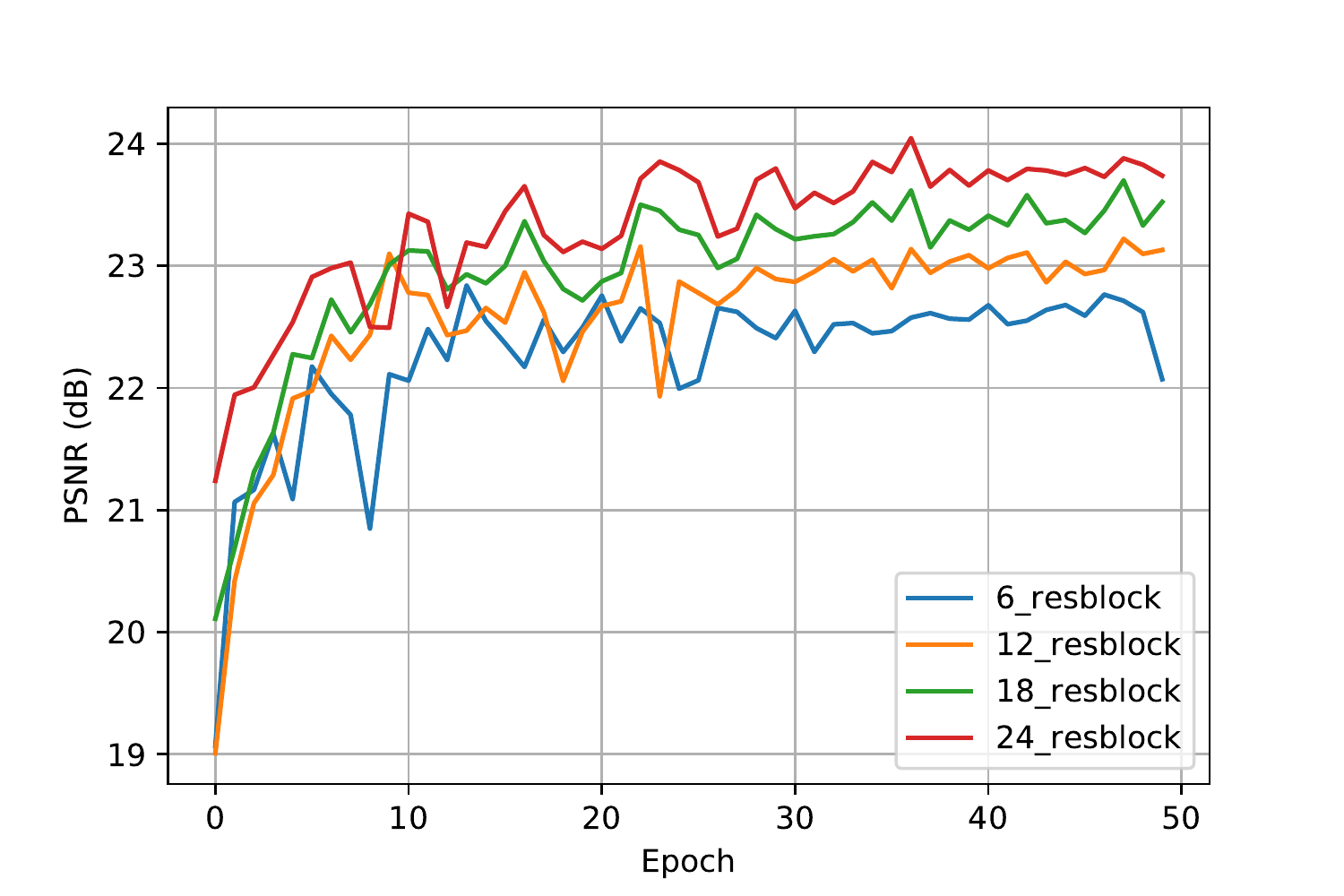} &
\includegraphics[width=0.48\textwidth]{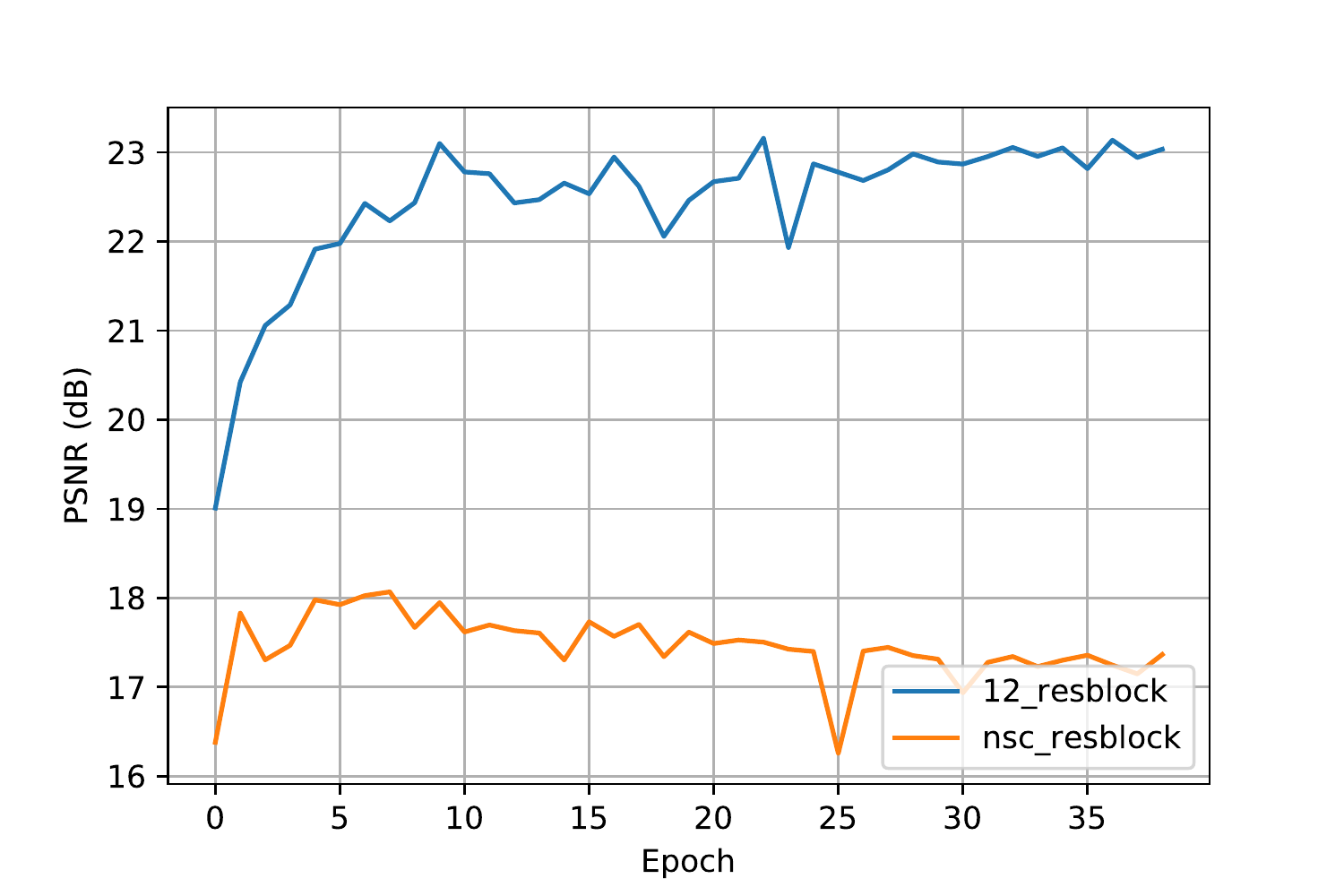}
\end{tabular}
\caption{\textbf{\emph{Left:}} The testing performance comparisons on NTIRE2018 outdoor scenes in different block size cases.
\textbf{\emph{Right:}} The testing performance comparisons of network with/without skip connection between encoder and decoder module on NTIRE2018 outdoor scenes. The "12\_resblock" represents network with skip connections; "nsc\_resblock" represents network without skip connections. The network is trained about 40 epoches.}
\label{fig:ablation}
\end{figure}

\subsubsection{Experiment results}
We have taken part in the NTIRE2018 challenge on image dehazing based on the proposed network. In the final testing phase, our network has achieved top 6 ranking out of 21 teams on I-HAZE track without using any data from O-HAZE and won the NTIRE 2018 honorable mentioned award. It should be noted that our network is very straight-forward and we haven't applied any specific training trick to further boost performance during training period and we haven't re-trained our model for O-HAZE track at that submission time.

Compared with other top methods in NTIRE Dehazing Challenge, our proposed network has several distinguished differences: (1) Most of other top methods use denseblock in their networks while we just use simple residual block then. Empirically, denseblock has better learning power and will have much potentials to boost output performance. (2) Multi-scale or multi-direction ensemble inferences at testing stage are used in some top methods to achieve better performance. In contrary, we haven't applied this tricky strategy. We just use the single output for testing. Using ensemble inference strategy empirically has great potentials to achieve better performance. (3) The last is not the least. As the images used in NTIRE Dehazing Challenge are very large with 4K high-resolution, all these top methods use patch based training strategy without taking entire image as input. Their network can not afford such large image. In contrary, our proposed network can directly process the ultra high-definition realistic hazed image up to 4K resolution with superior restoration performance at a reasonable speed and memory usage.

\begin{figure}
\centering
\includegraphics[width=0.9\textwidth]{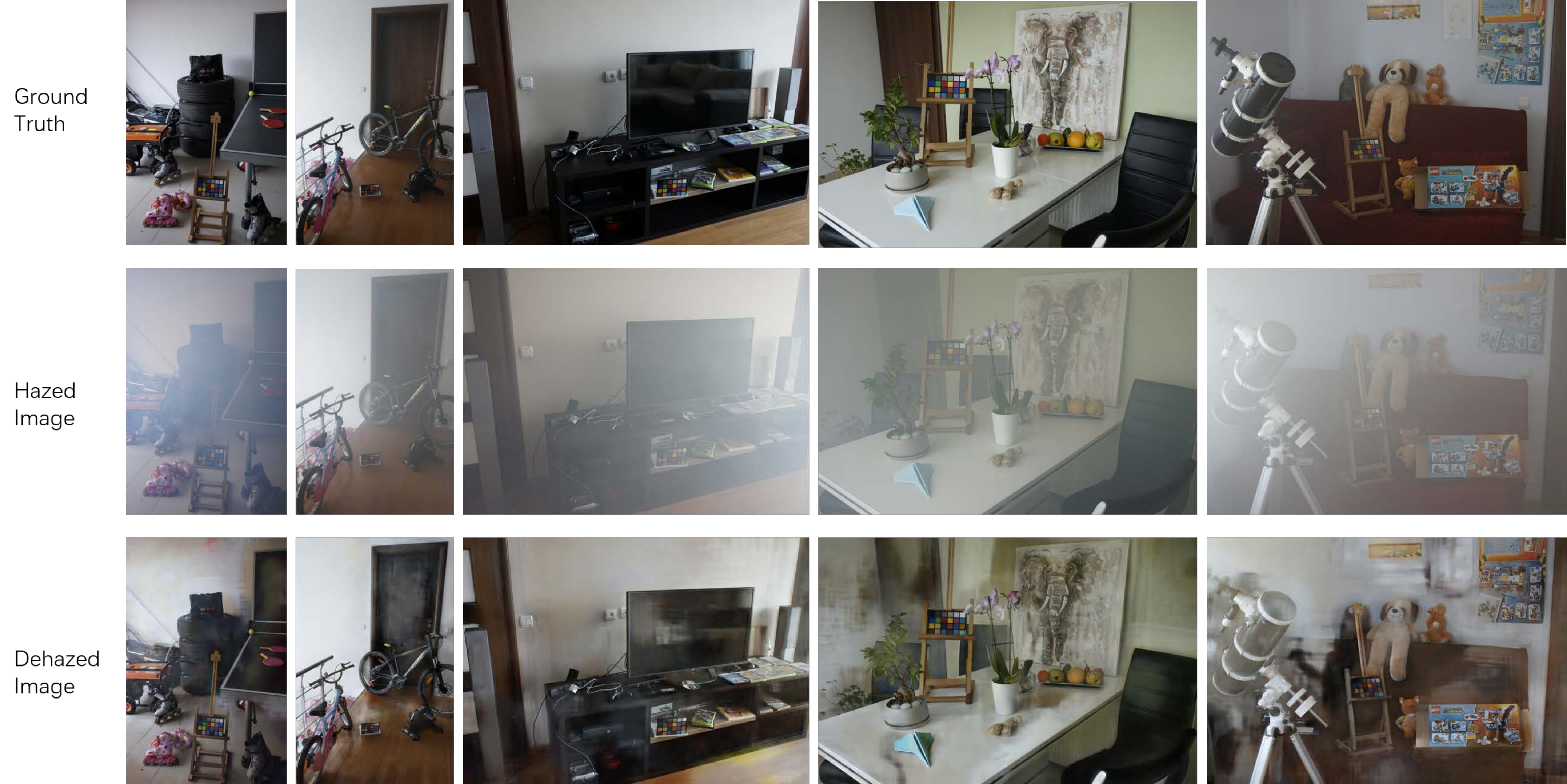}
\caption{Several challenging realistic dehazed examples from I-HAZE by using PFF-Net}
\label{fig:Indoor}
\end{figure}

\begin{figure}
\centering
\includegraphics[width=0.9\textwidth]{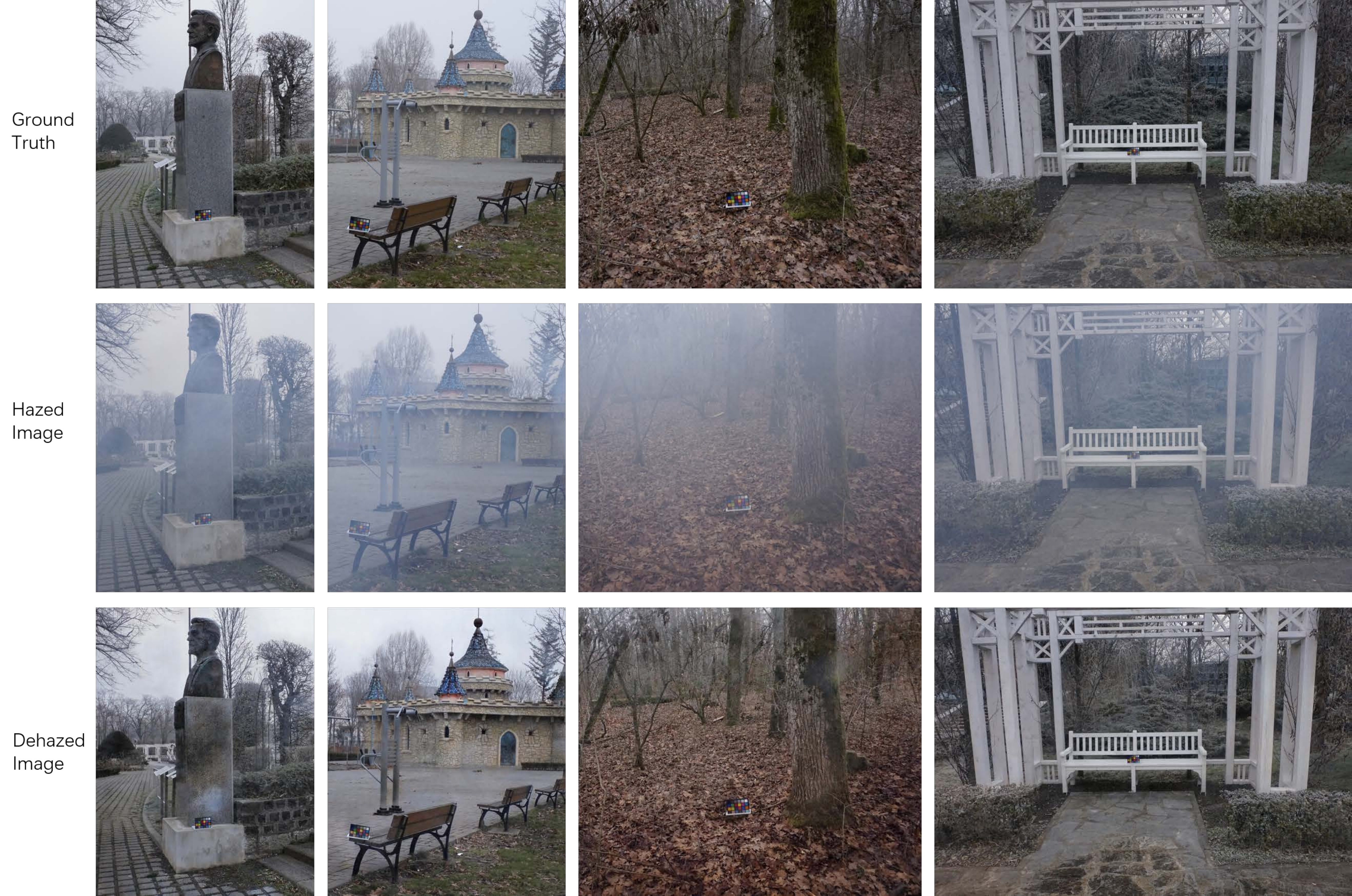}
\caption{Several challenging realistic dehazed examples from O-HAZE by using PFF-Net}
\label{fig:Outdoor}
\end{figure}

Several dehazed examples on realistic images from NTIRE2018 are shown in Figure~\ref{fig:Indoor} and Figure~\ref{fig:Outdoor}. All these images are at 4K resolution level which most current dehazed model cannot afford. Though challenging these examples are, our network still can obtain relatively satisfactory dehazed results with natural color saturation and acceptable perceptual quality.

With the aim to compare with state-of-the-art methods, we evaluate our method on the commonly referred public benchmark. We first pre-train our network on DIV2K\cite{timofte2017ntire}, then fine-tune the pre-trained network on RESIDE training data without data augmentation. The training curve of fine-tuning process on RESIDE is shown in Figure~\ref{fig:traincurve}. After about 8 epoches, the network begins to converge.

We evaluate the performance of our network on the SOTS. The comparison results on SOTS are shown in Table~\ref{tab:sots}. From the experimental comparisons, it has been demonstrated that our proposed network outperforms the current state-of-the-art methods, and achieves superior performance with great improvements.
\begin{table}[h]
\small
	\centering
	\caption{The dehazing performance evaluated on SOTS of RESIDE}
	\begin{tabular}{|c|c|c|c|c|c|c|c|c|}
		\hline
	        & DCP & CAP & NLD &DehazeNet & MSCNN & AOD-Net & GFN & PFF-Net(ours)\\
        \hline
		 PSNR & 16.62 & 19.05 & 17.29 & 21.14 & 17.57 & 19.06 & 22.30 & \textbf{24.78} \\
        \hline
         SSIM & 0.8179 & 0.8364 & 0.7489 & 0.8472 & 0.8102 & 0.8504 & 0.88 &\textbf{0.8923} \\
        \hline
	\end{tabular}
    \label{tab:sots}
\end{table}

\begin{figure*}
\begin{center}
\begin{tabular}{ccccccccc}
\includegraphics[width=0.1\textwidth]{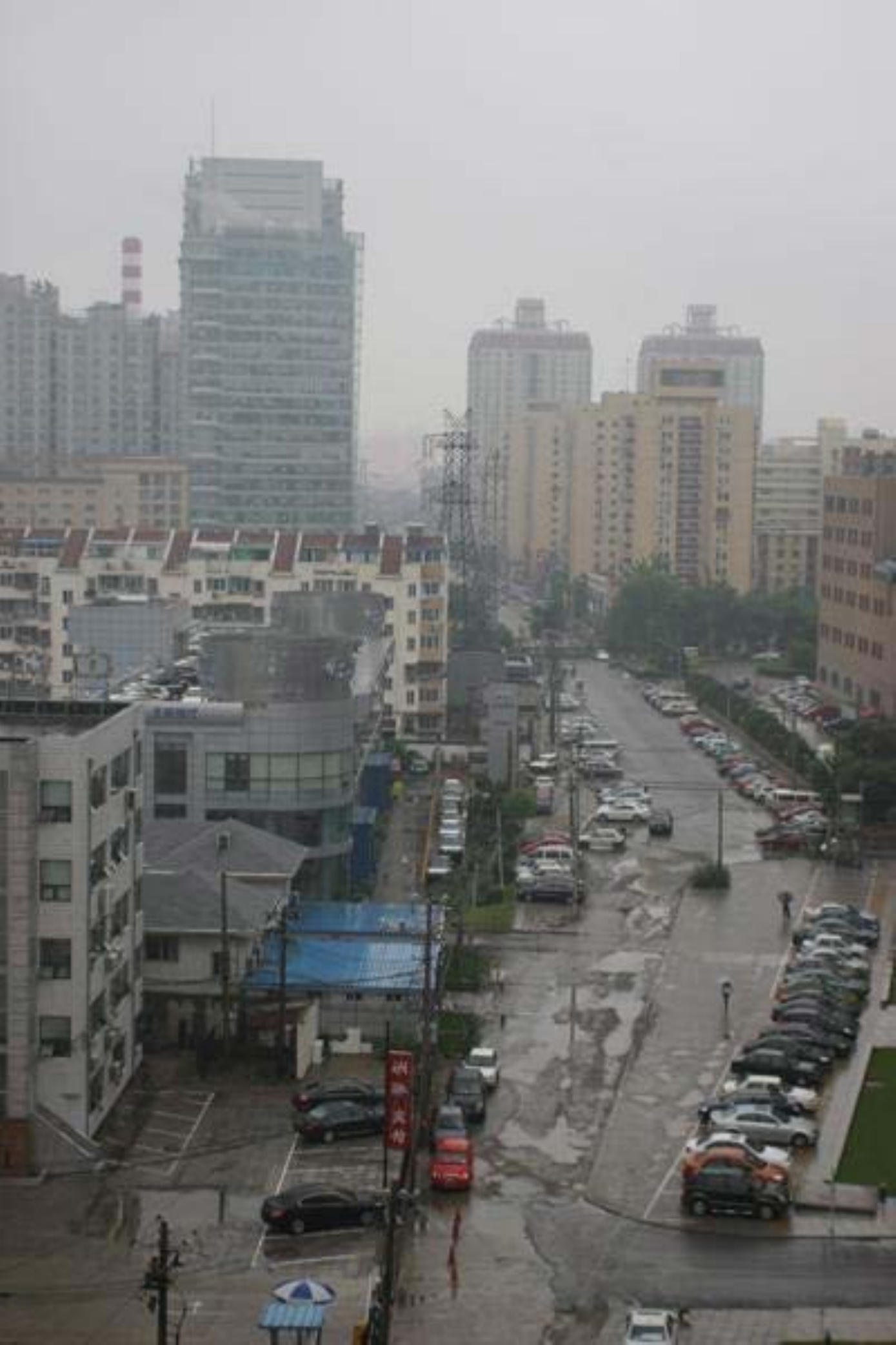} &
\includegraphics[width=0.1\textwidth]{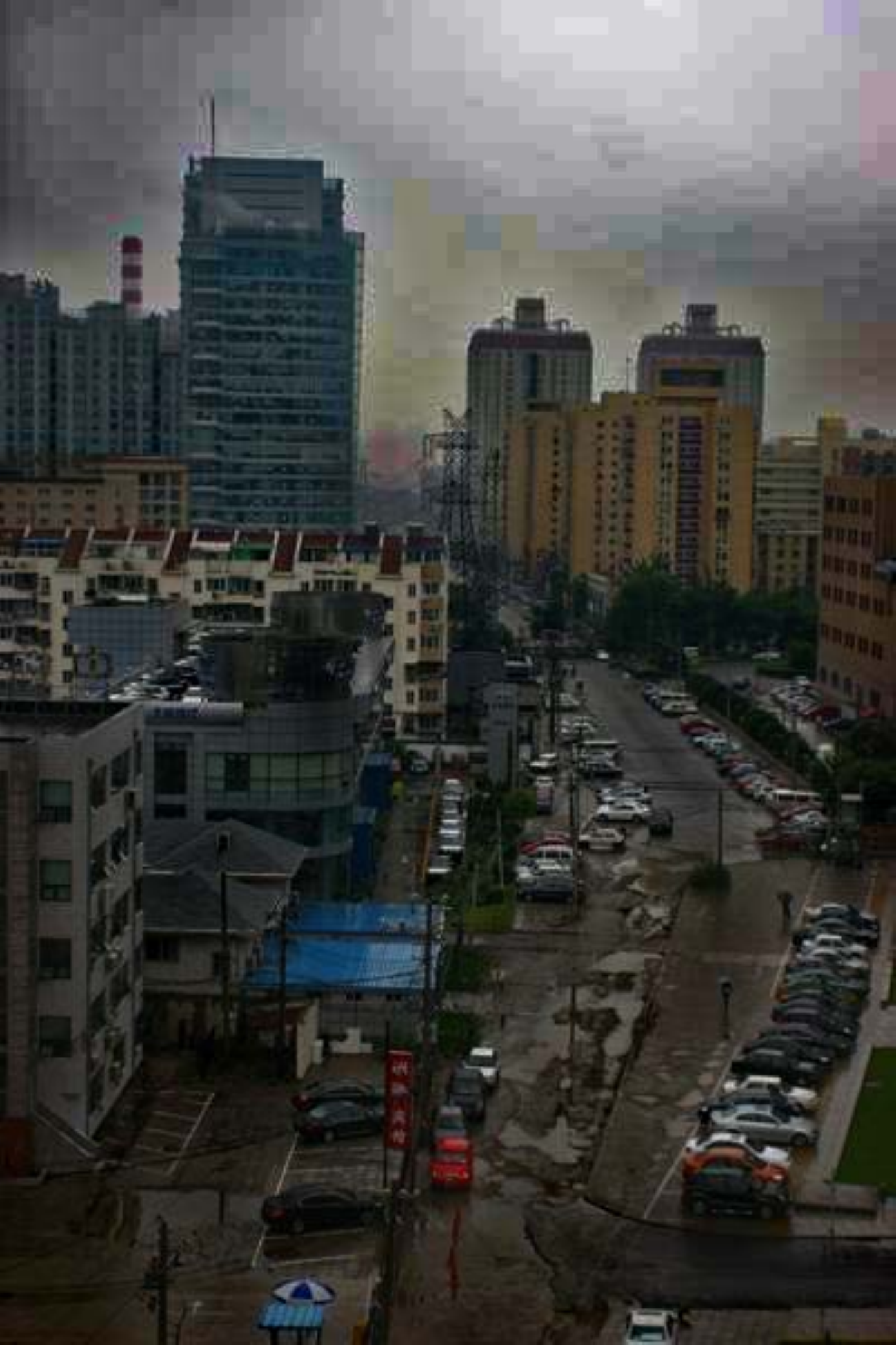} &
\includegraphics[width=0.1\textwidth]{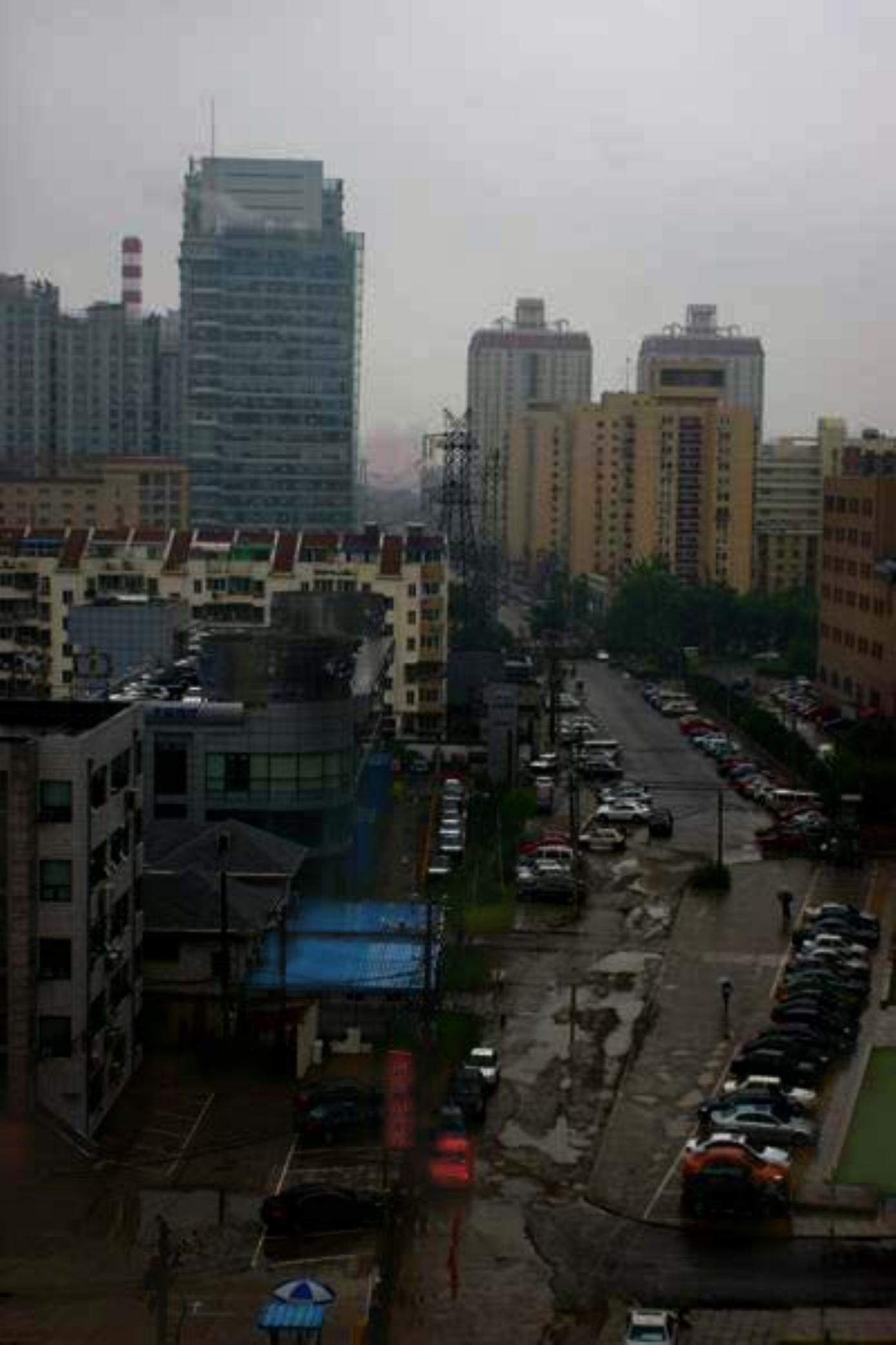} &
\includegraphics[width=0.1\textwidth]{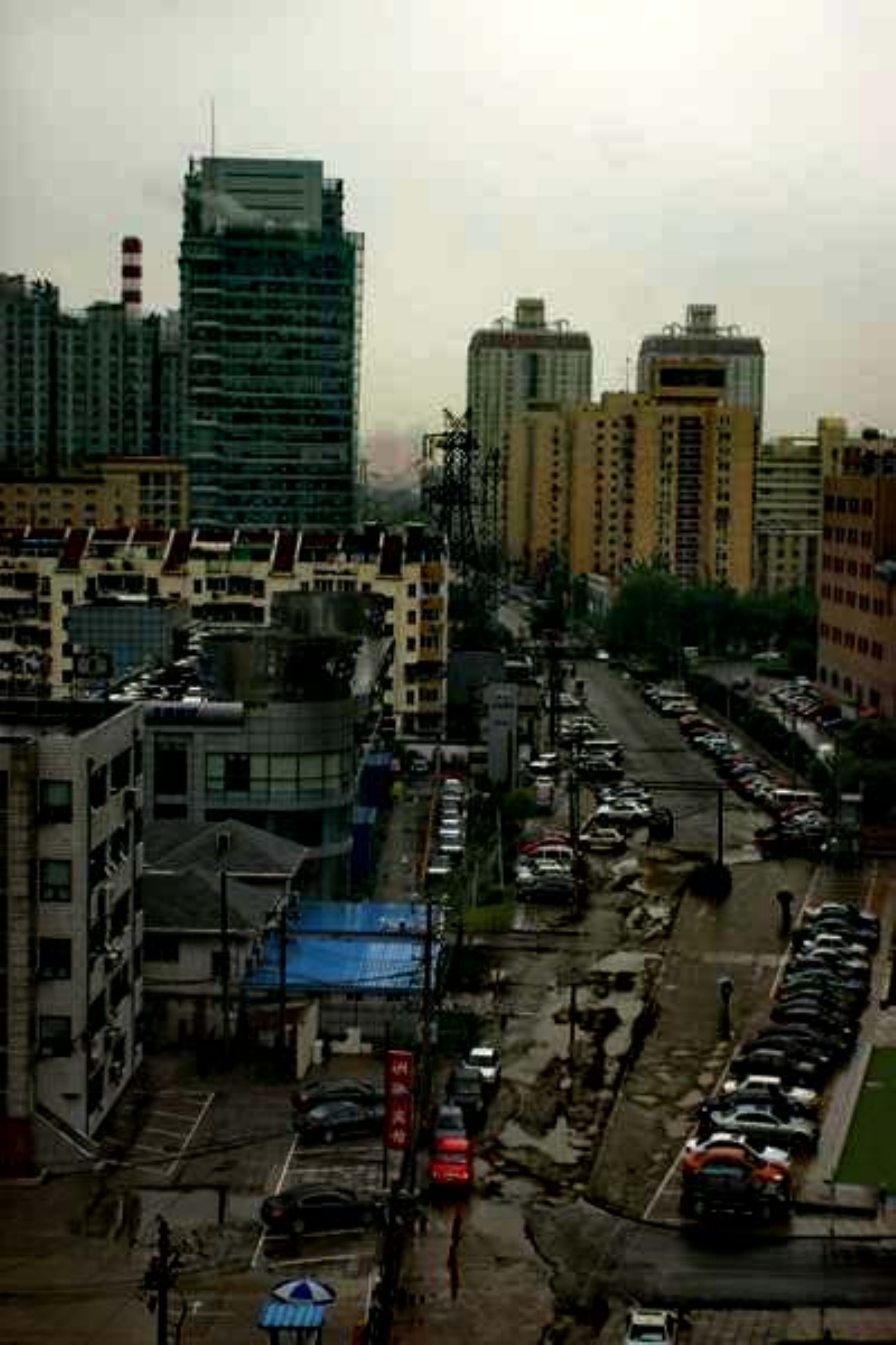} &
\includegraphics[width=0.1\textwidth]{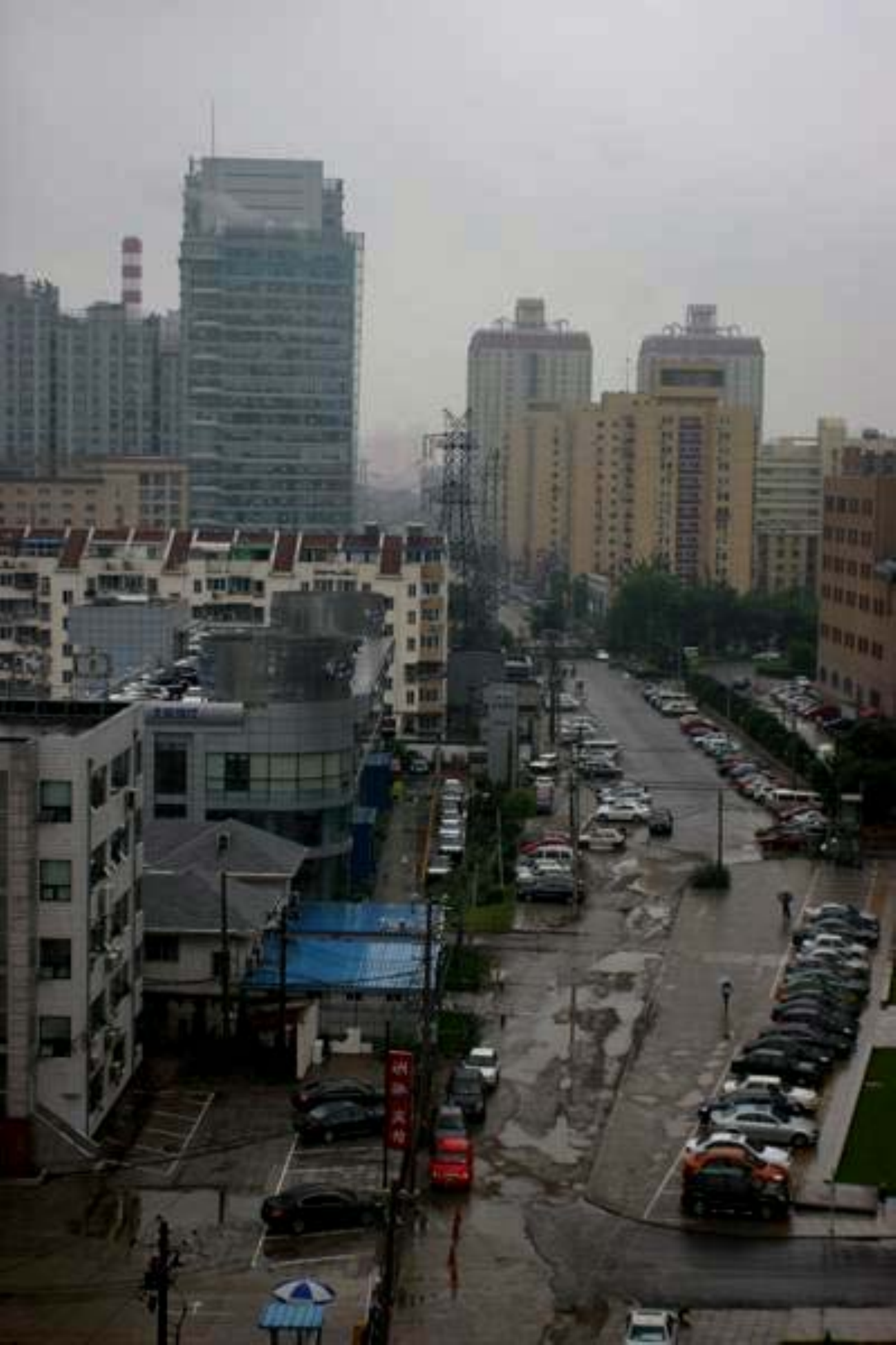} &
\includegraphics[width=0.1\textwidth]{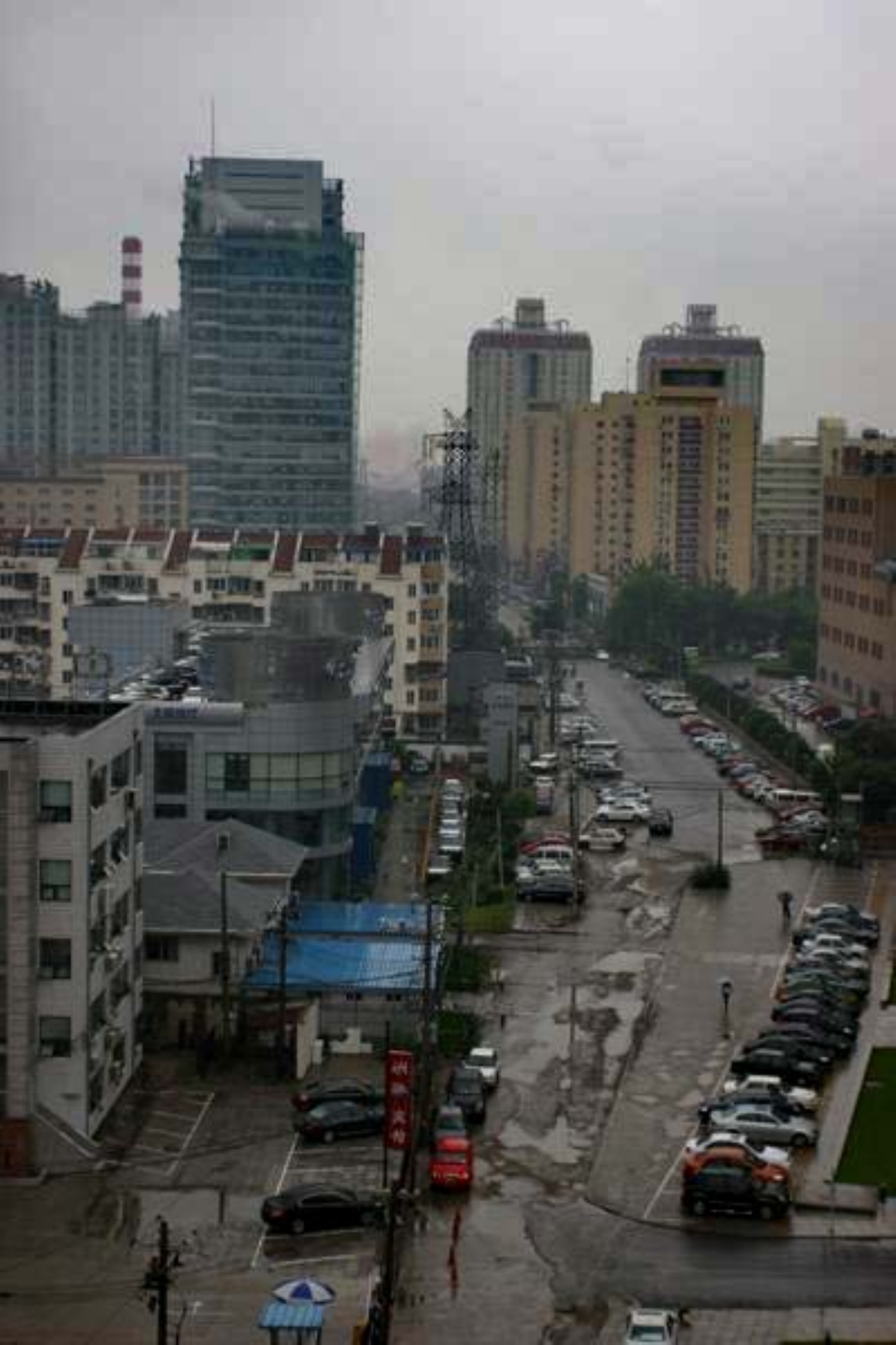} &
\includegraphics[width=0.1\textwidth]{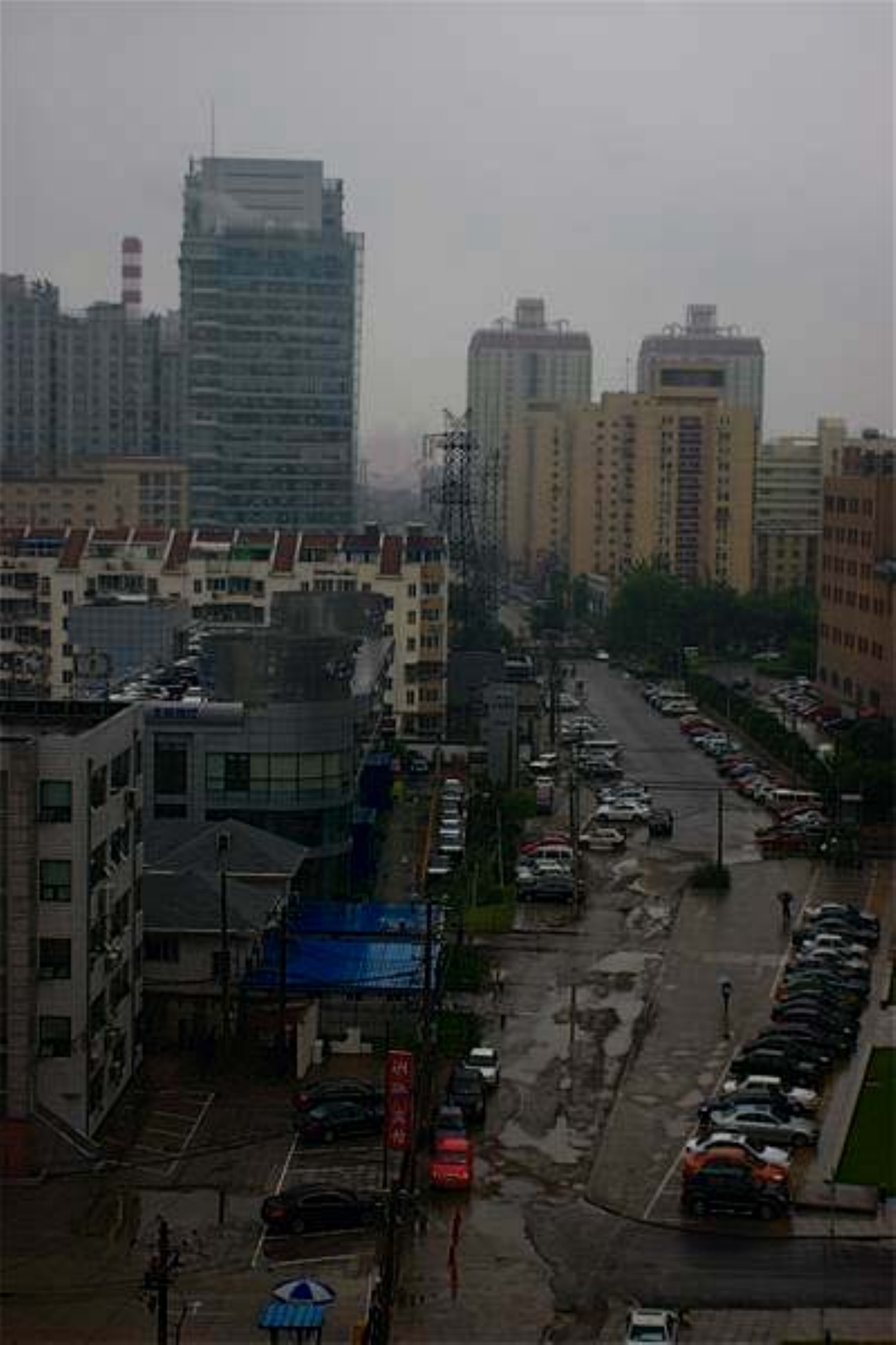} &
\includegraphics[width=0.1\textwidth]{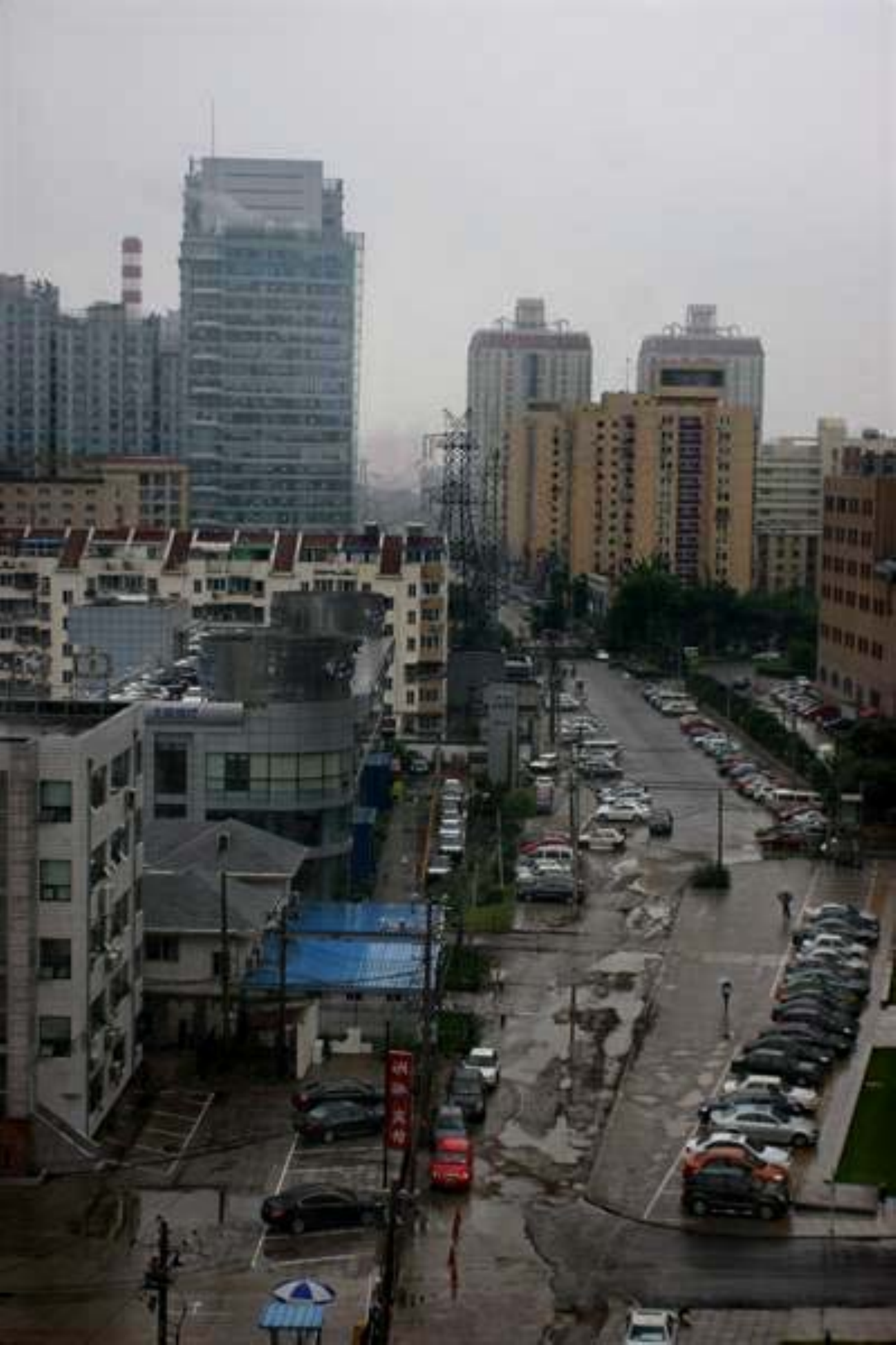} &
\includegraphics[width=0.1\textwidth]{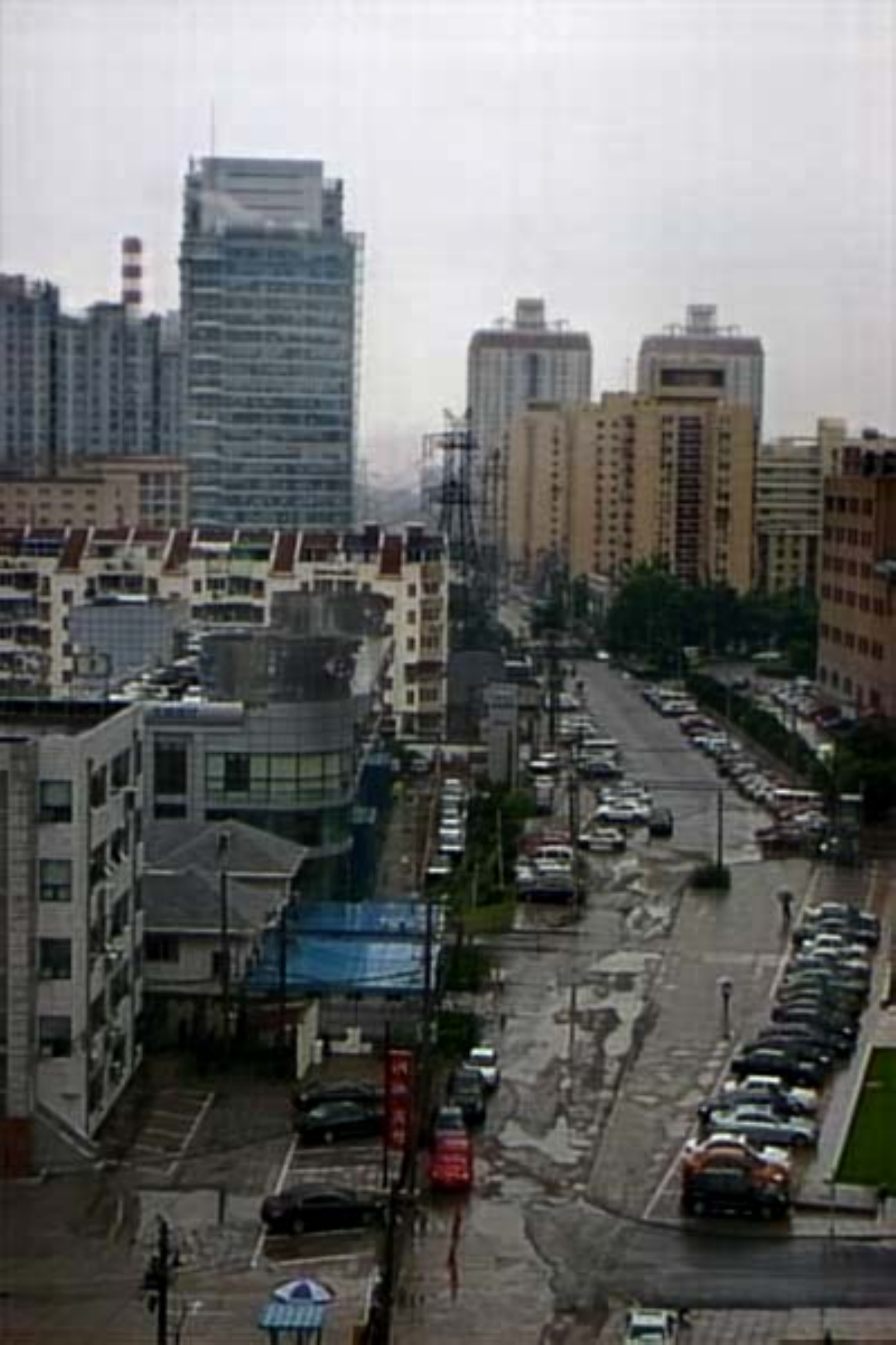} \\

\includegraphics[width=0.1\textwidth]{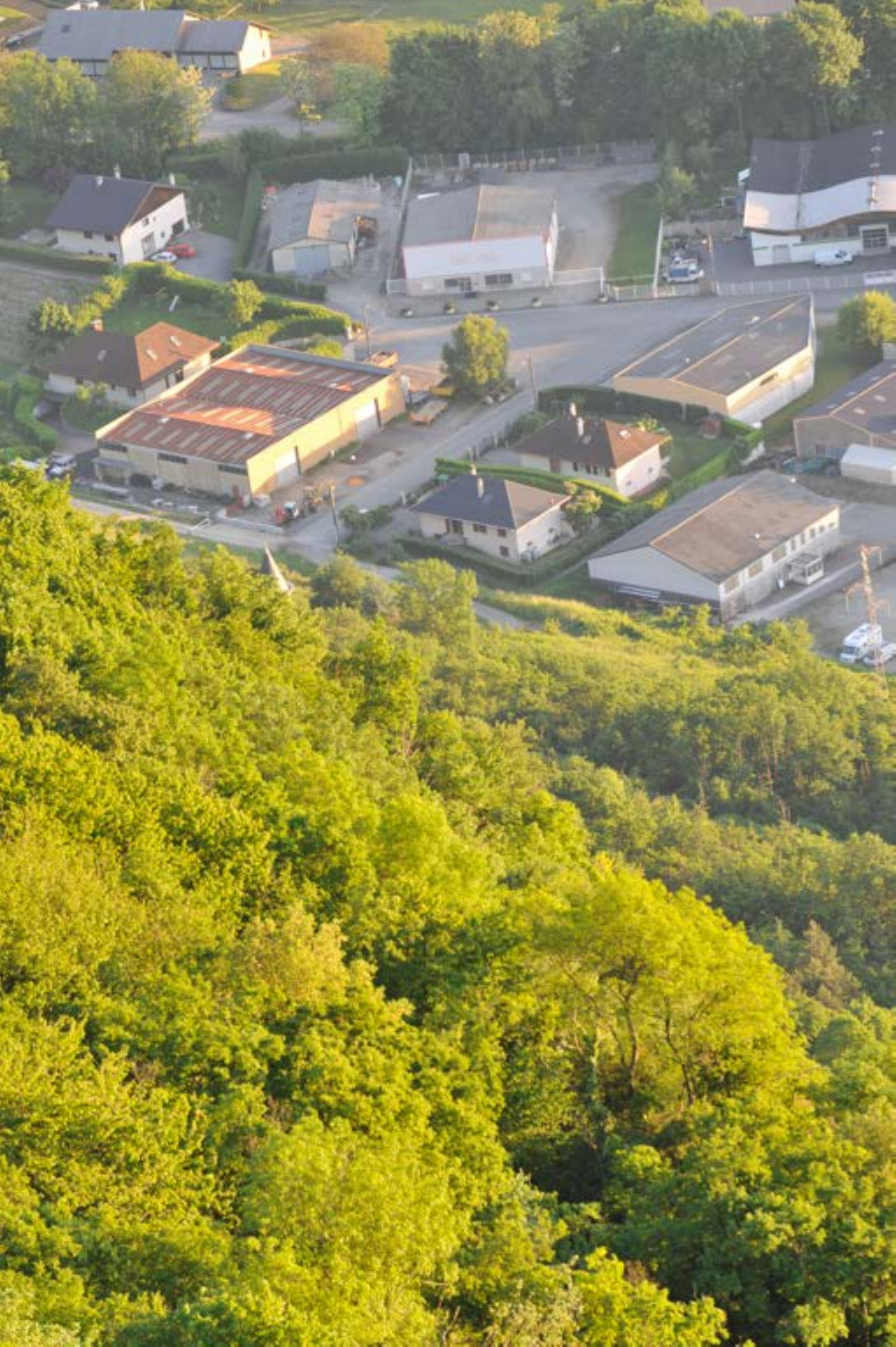} &
\includegraphics[width=0.1\textwidth]{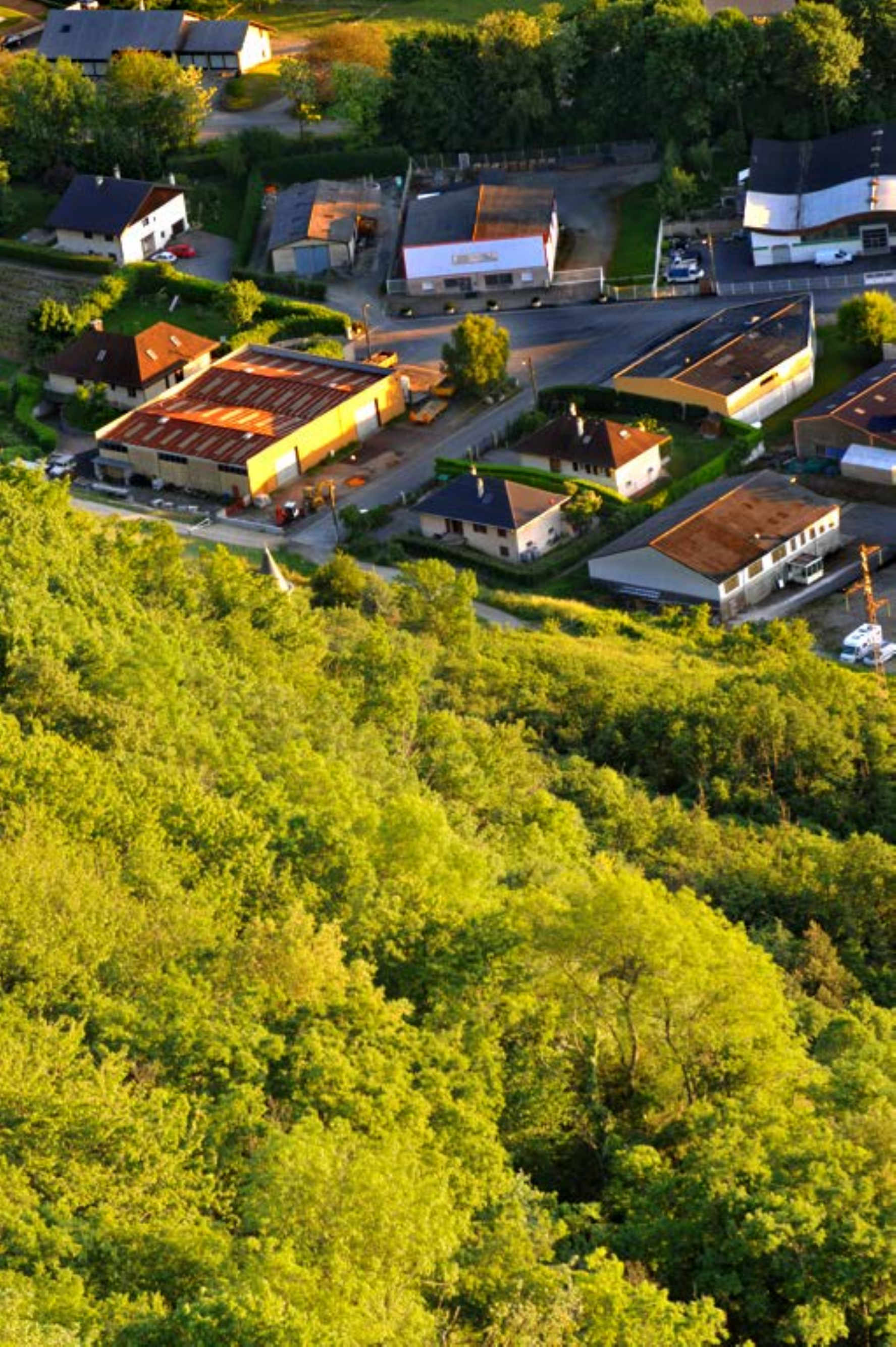} &
\includegraphics[width=0.1\textwidth]{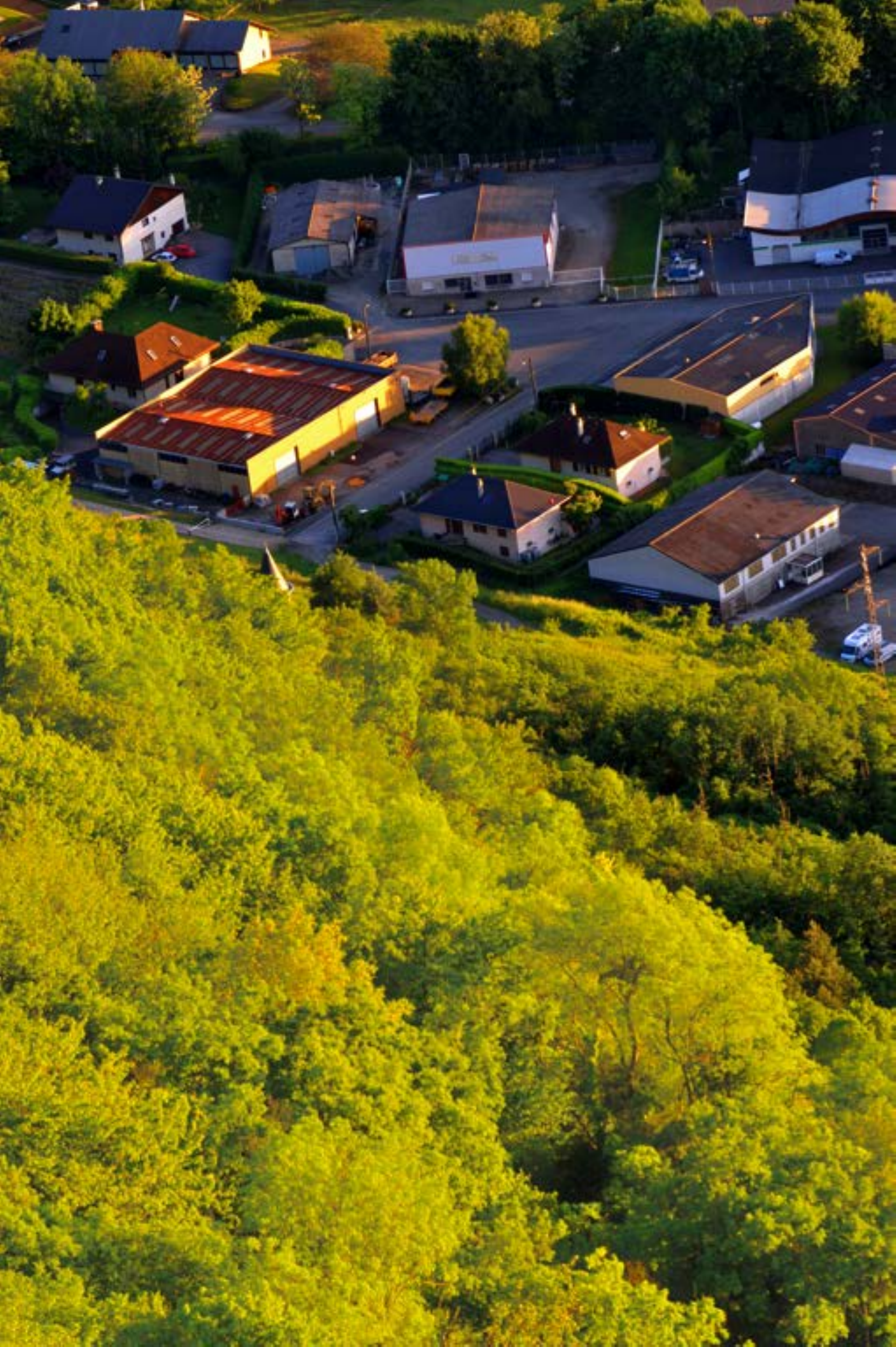} &
\includegraphics[width=0.1\textwidth]{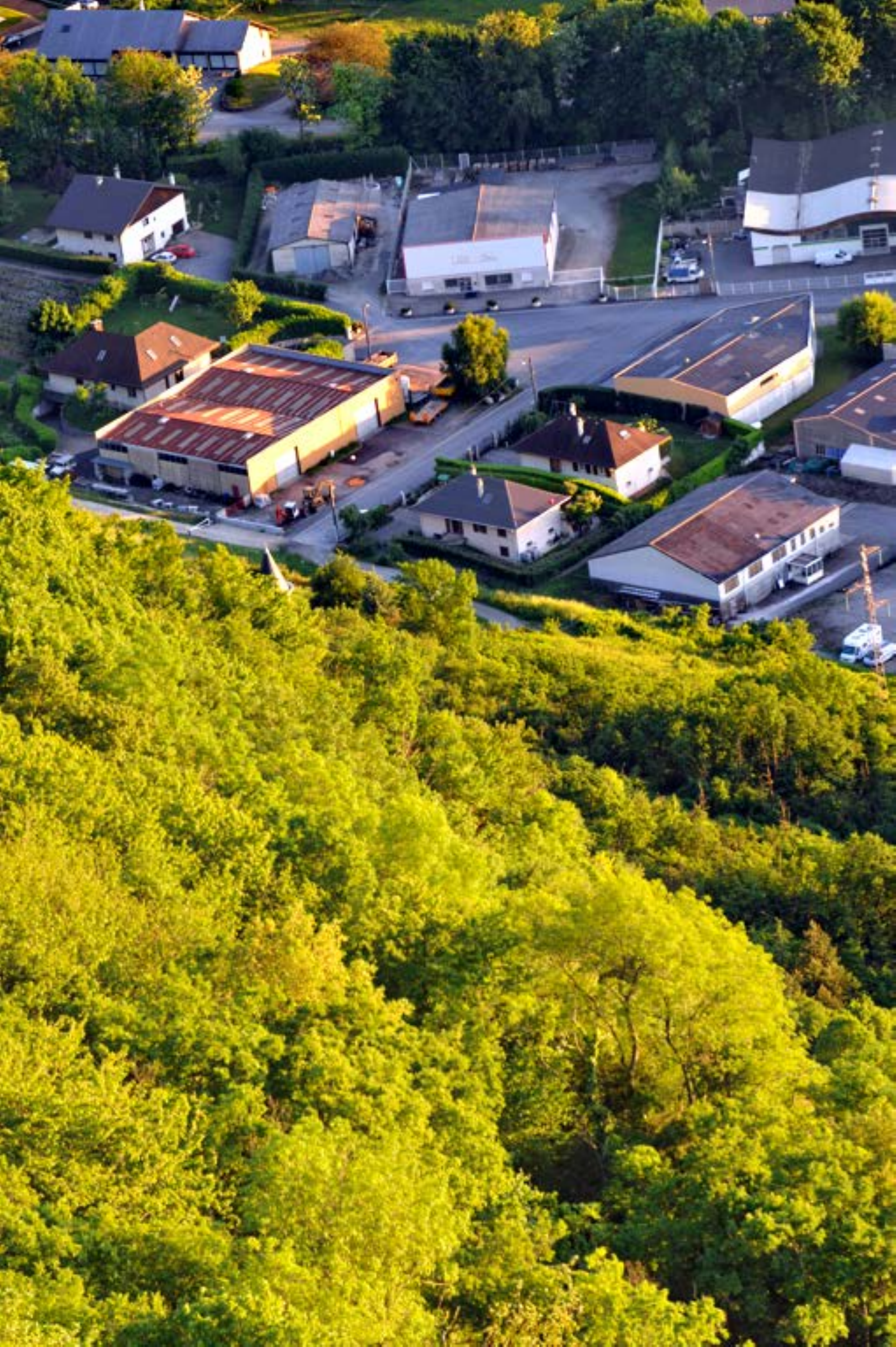} &
\includegraphics[width=0.1\textwidth]{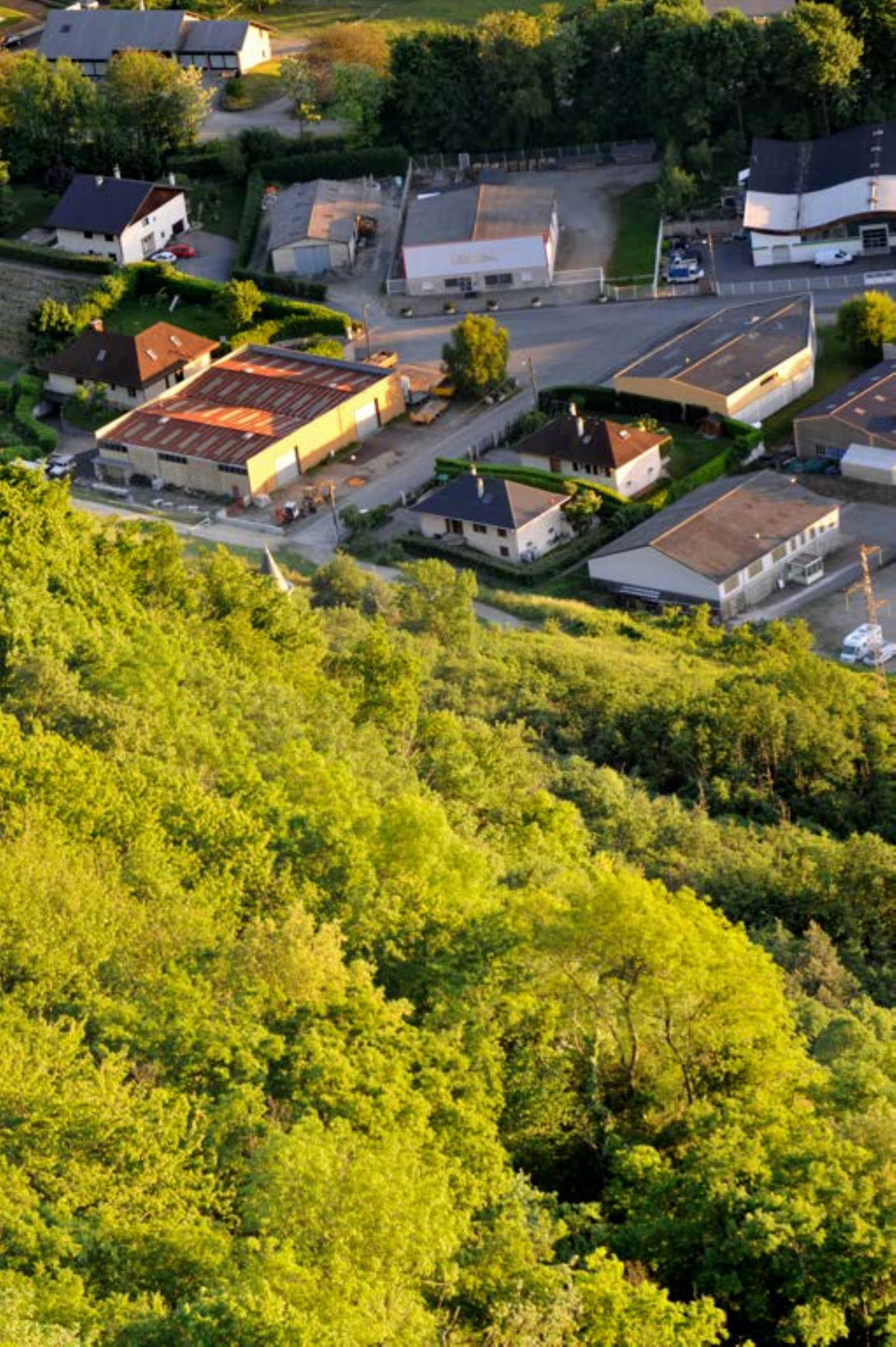} &
\includegraphics[width=0.1\textwidth]{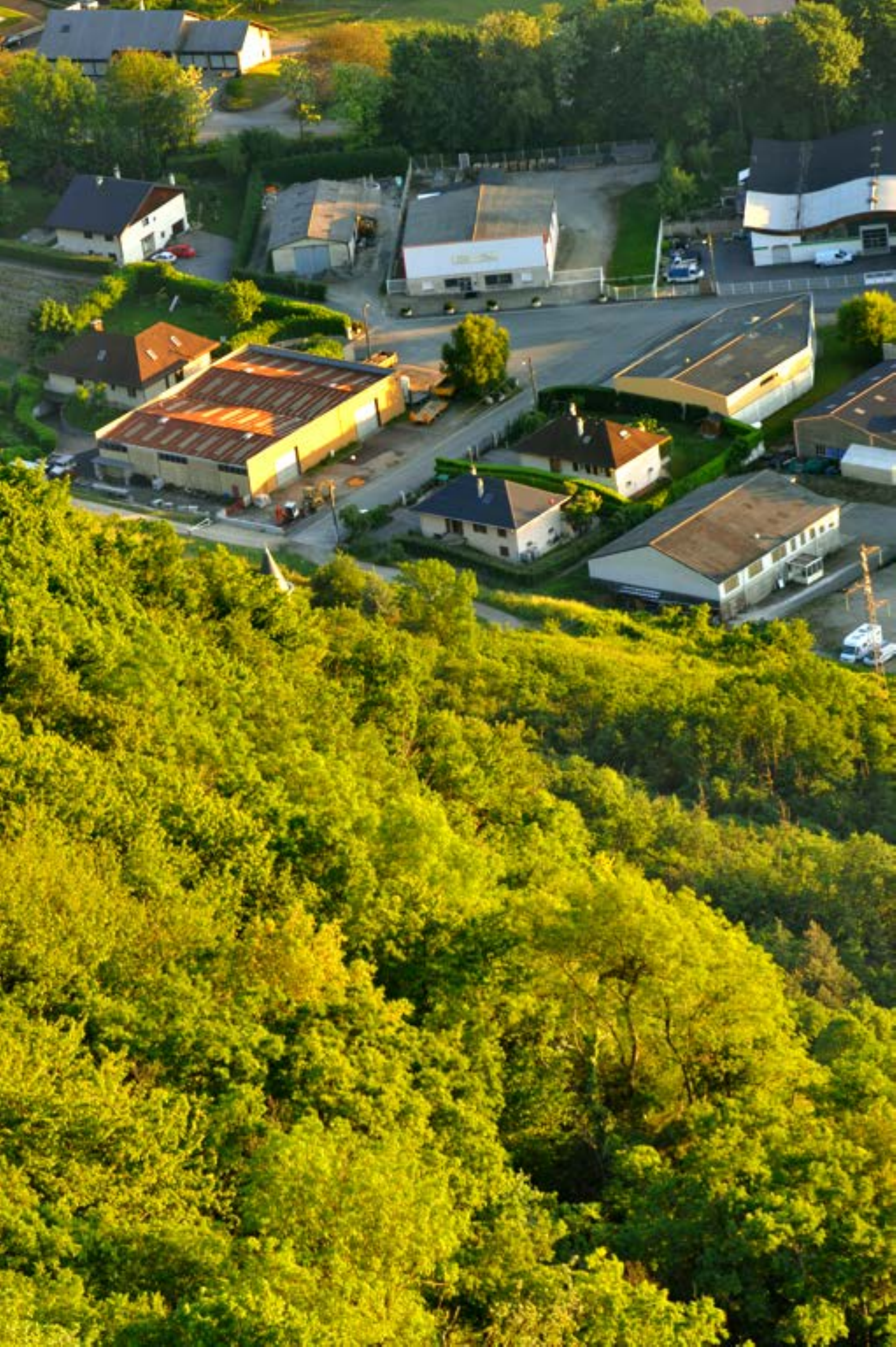} &
\includegraphics[width=0.1\textwidth]{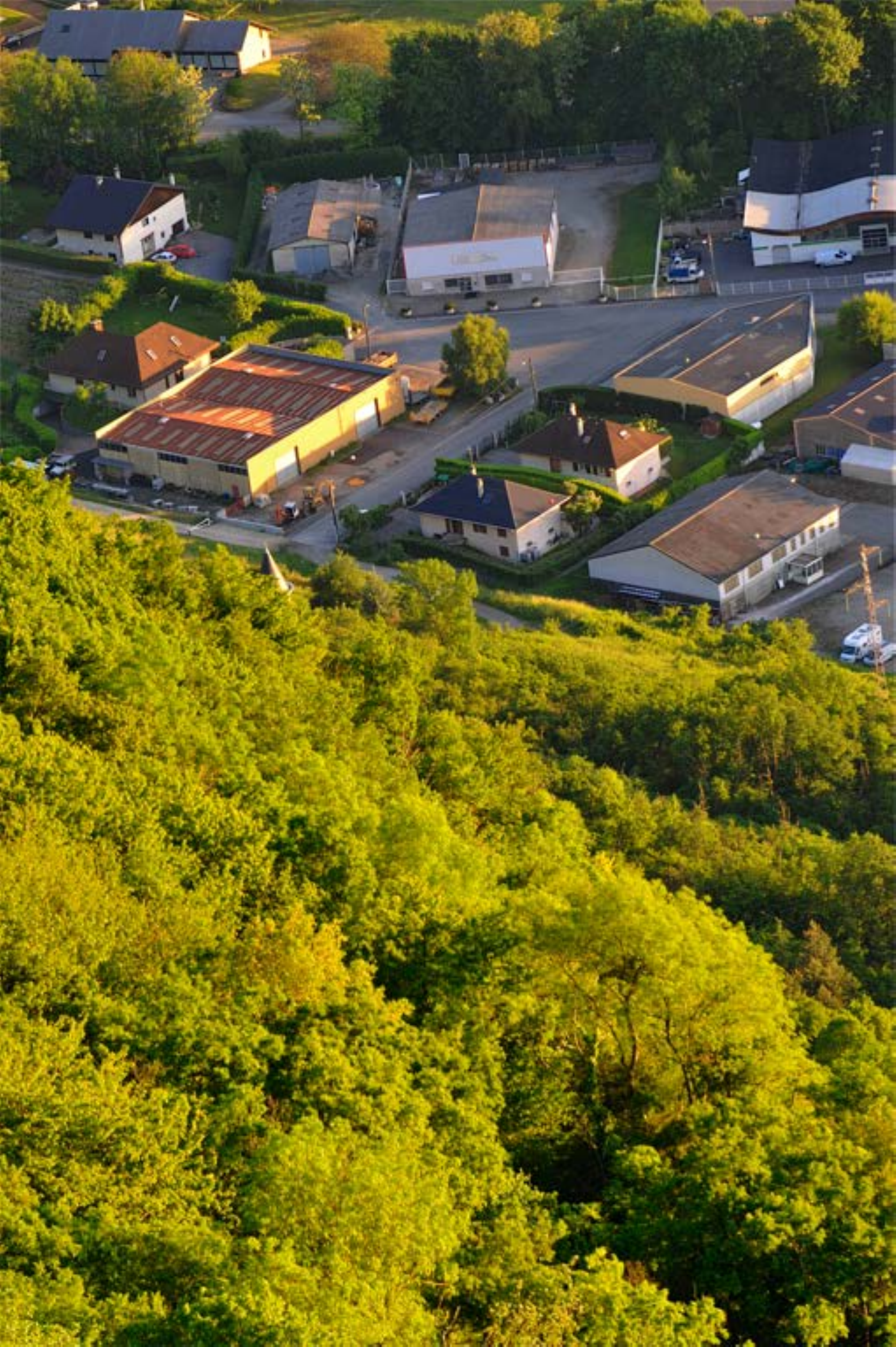} &
\includegraphics[width=0.1\textwidth]{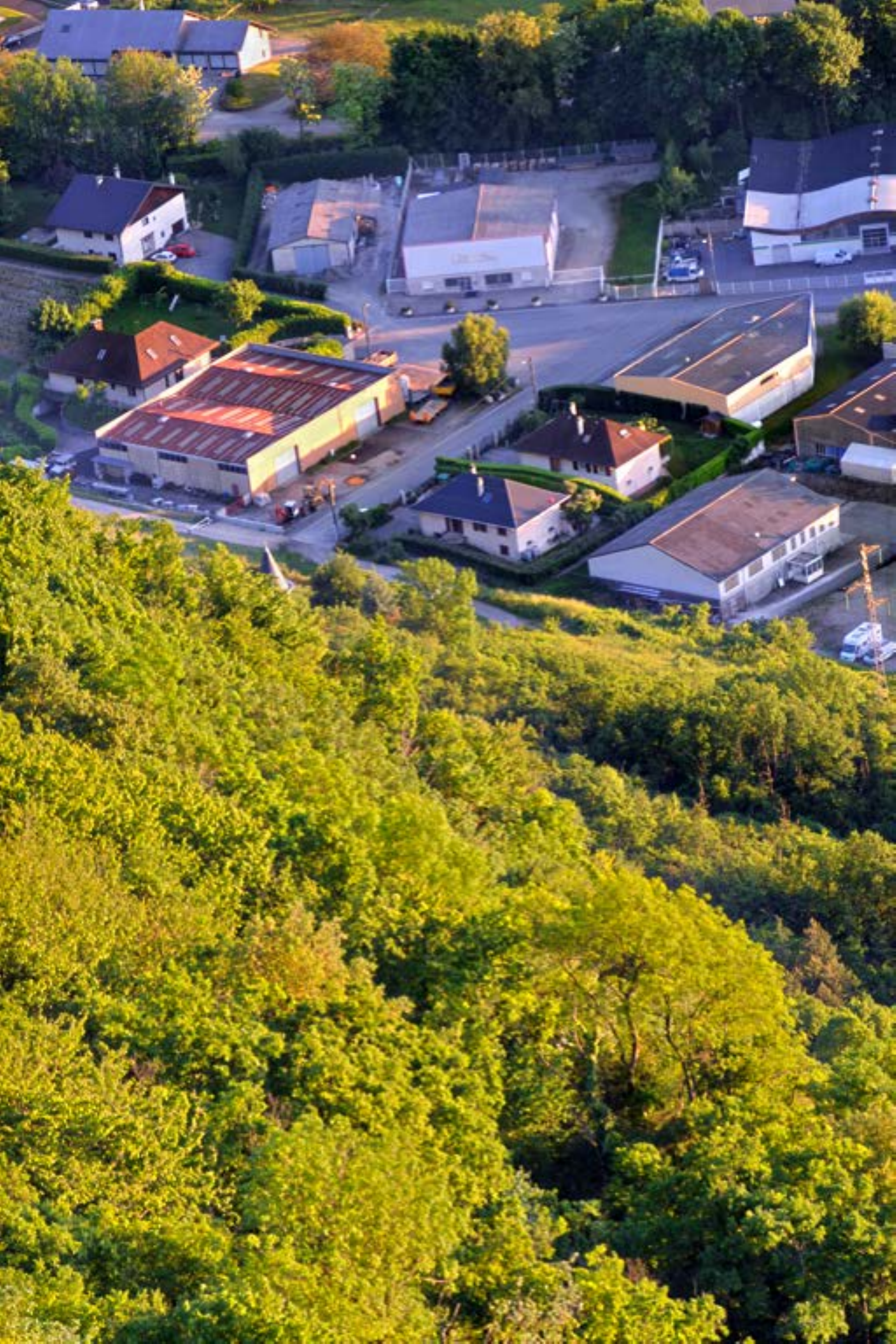} &
\includegraphics[width=0.1\textwidth]{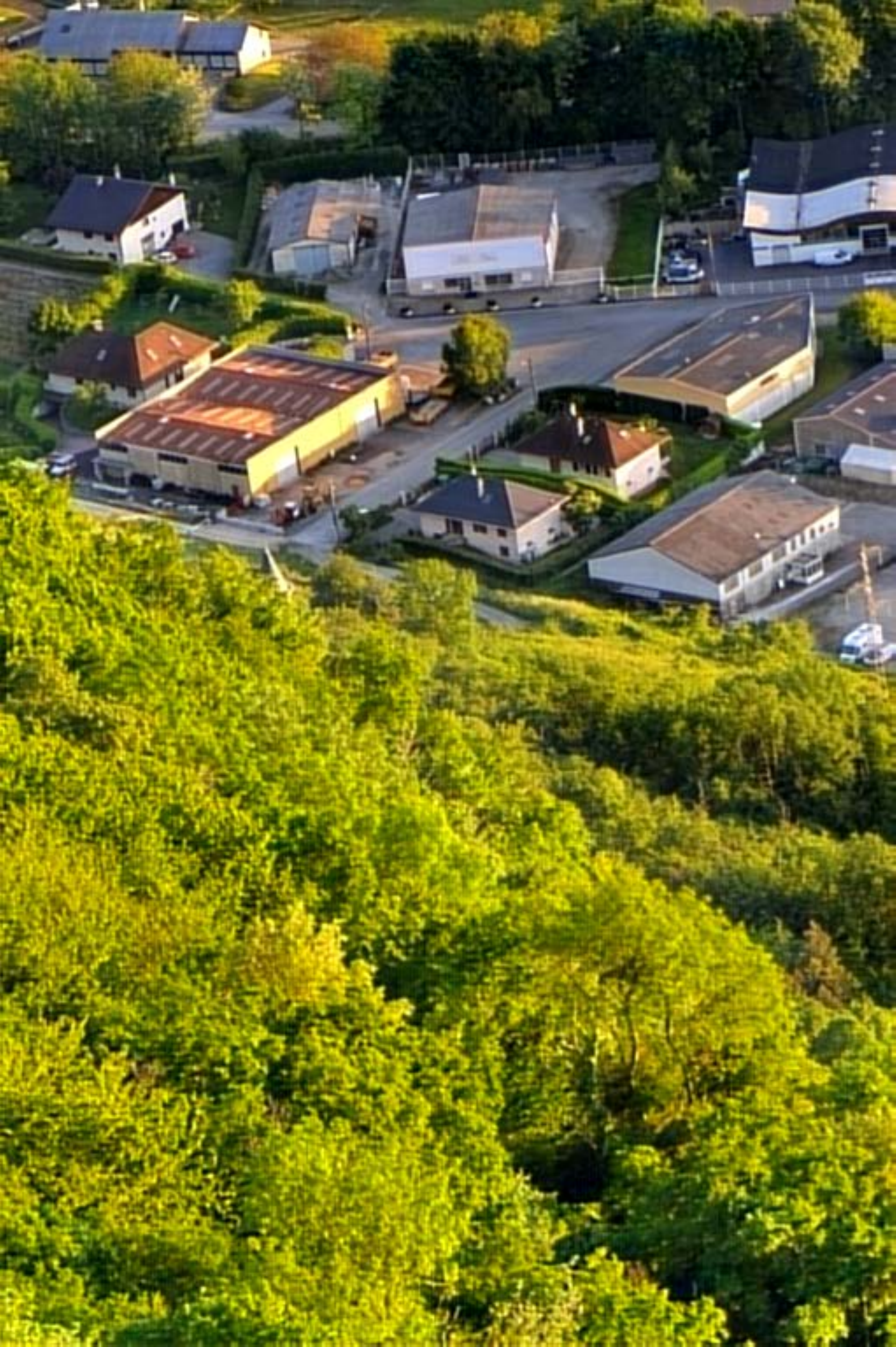} \\

\includegraphics[width=0.1\textwidth]{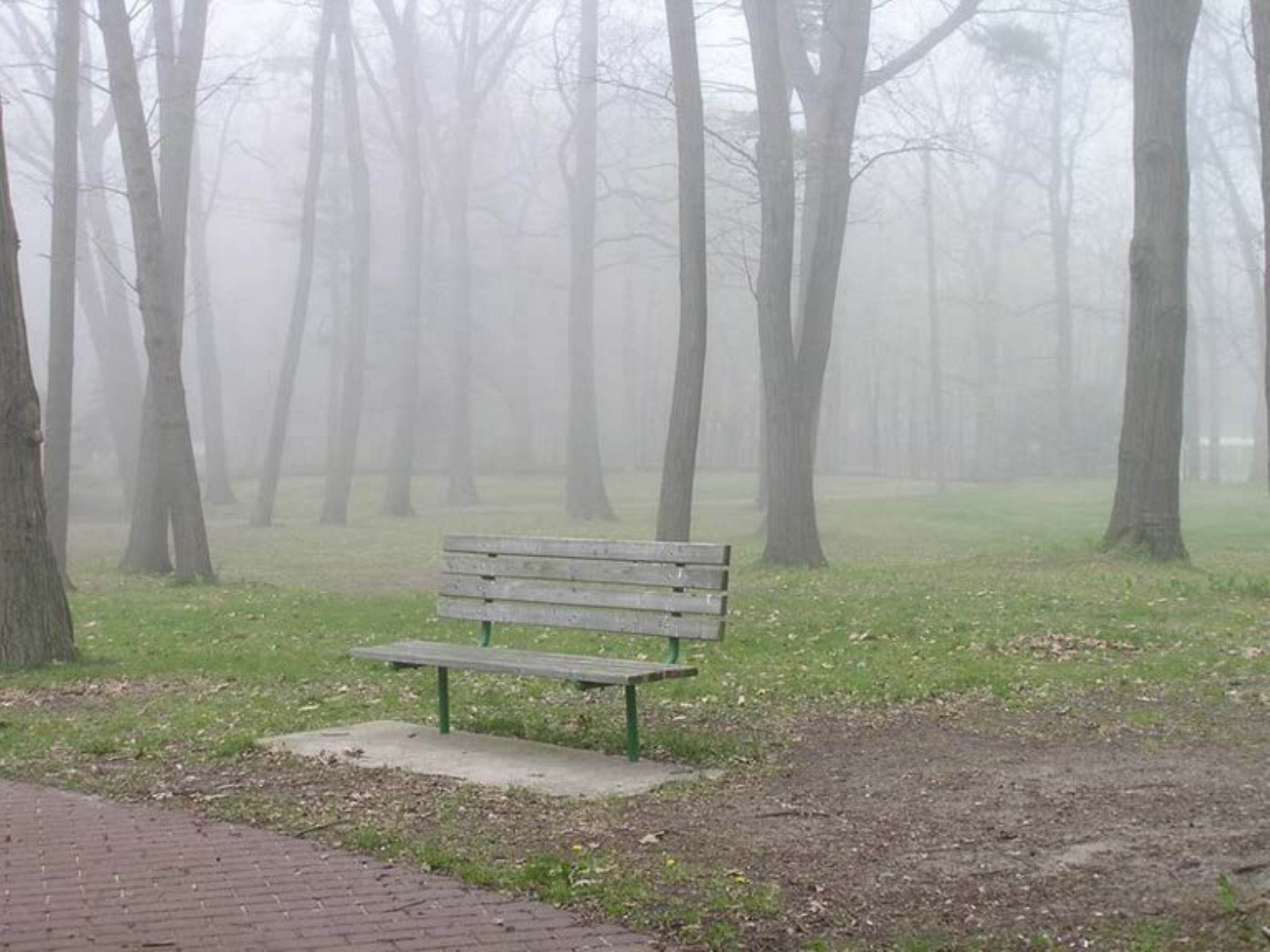} &
\includegraphics[width=0.1\textwidth]{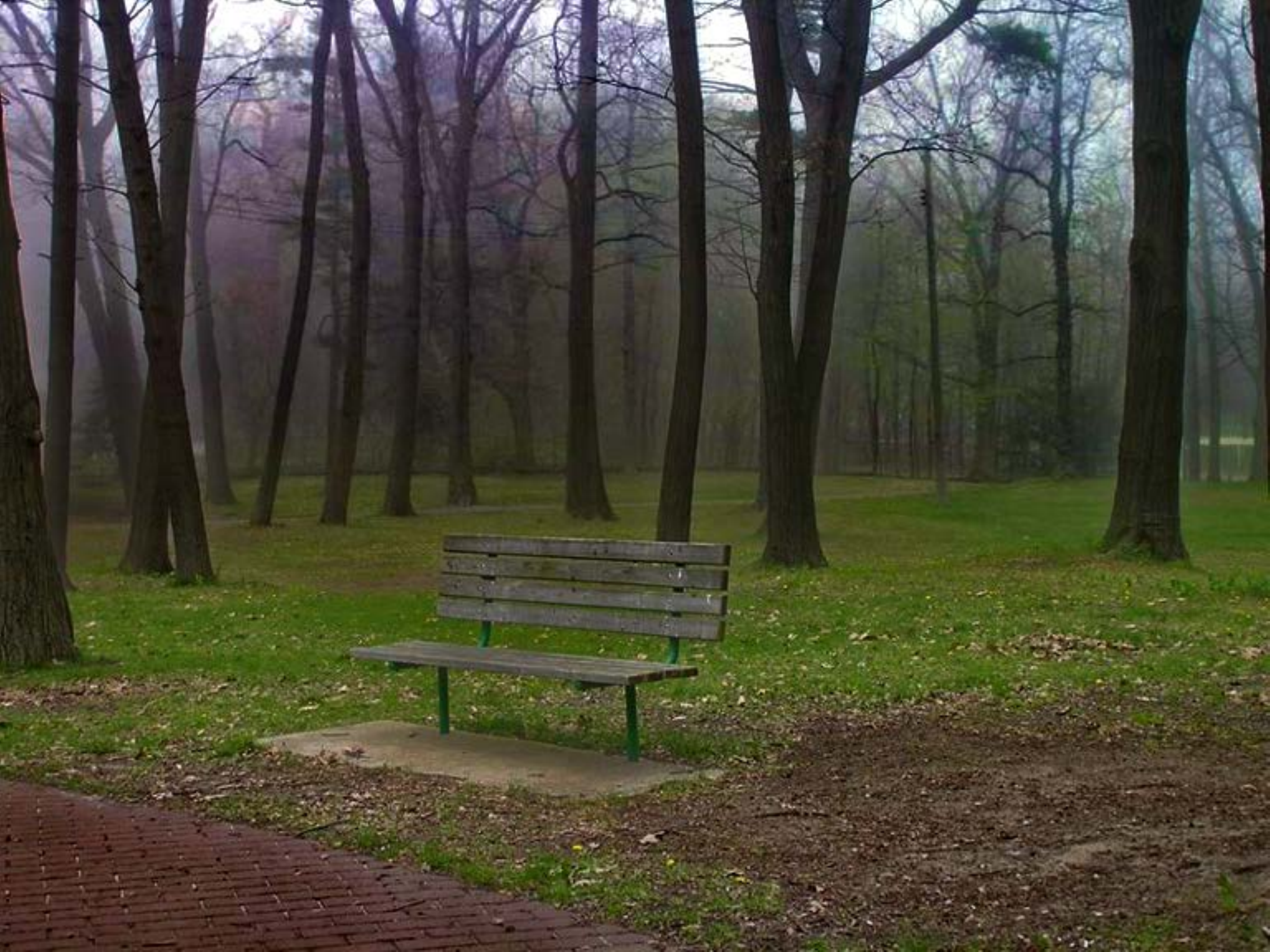} &
\includegraphics[width=0.1\textwidth]{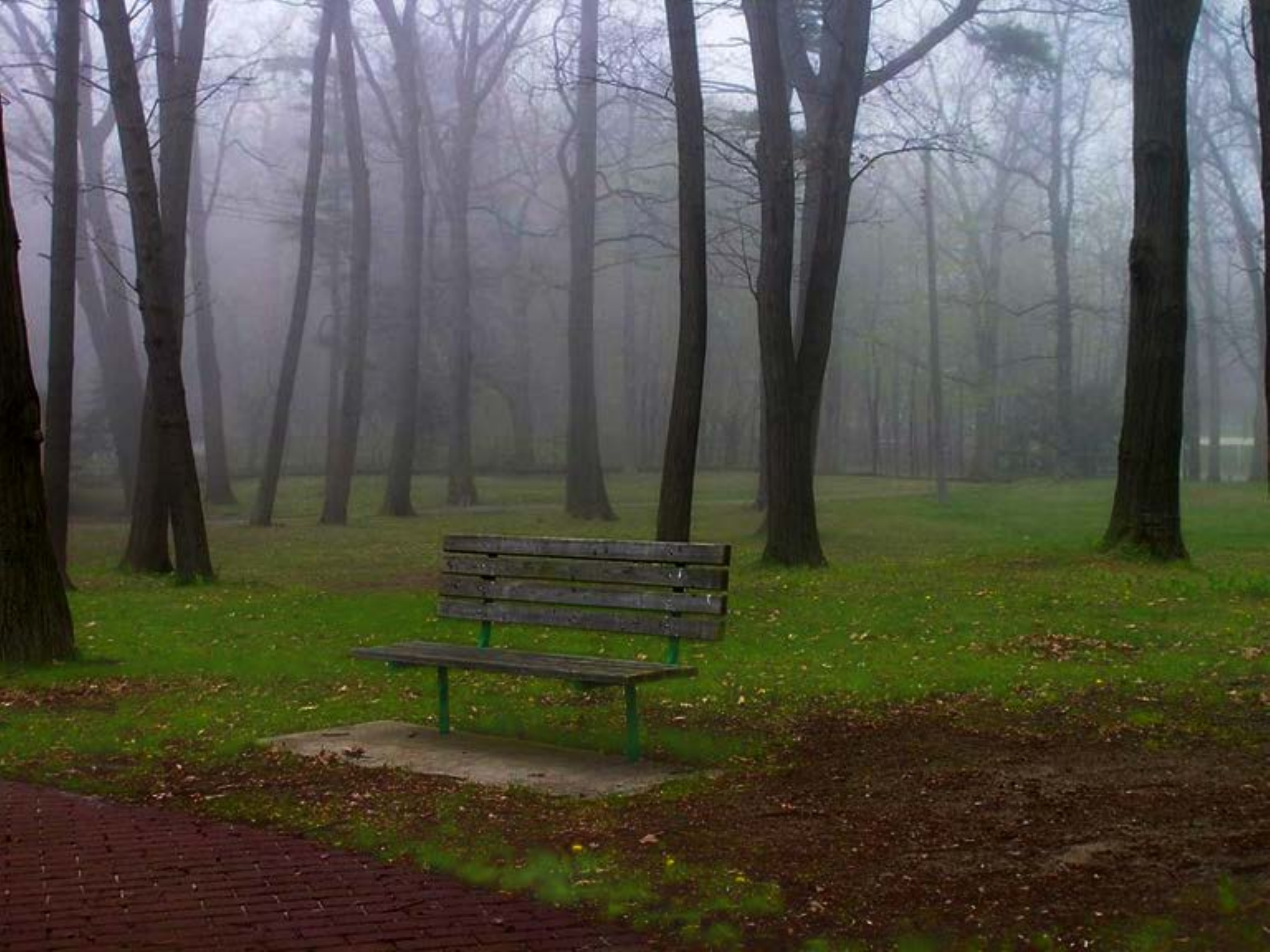} &
\includegraphics[width=0.1\textwidth]{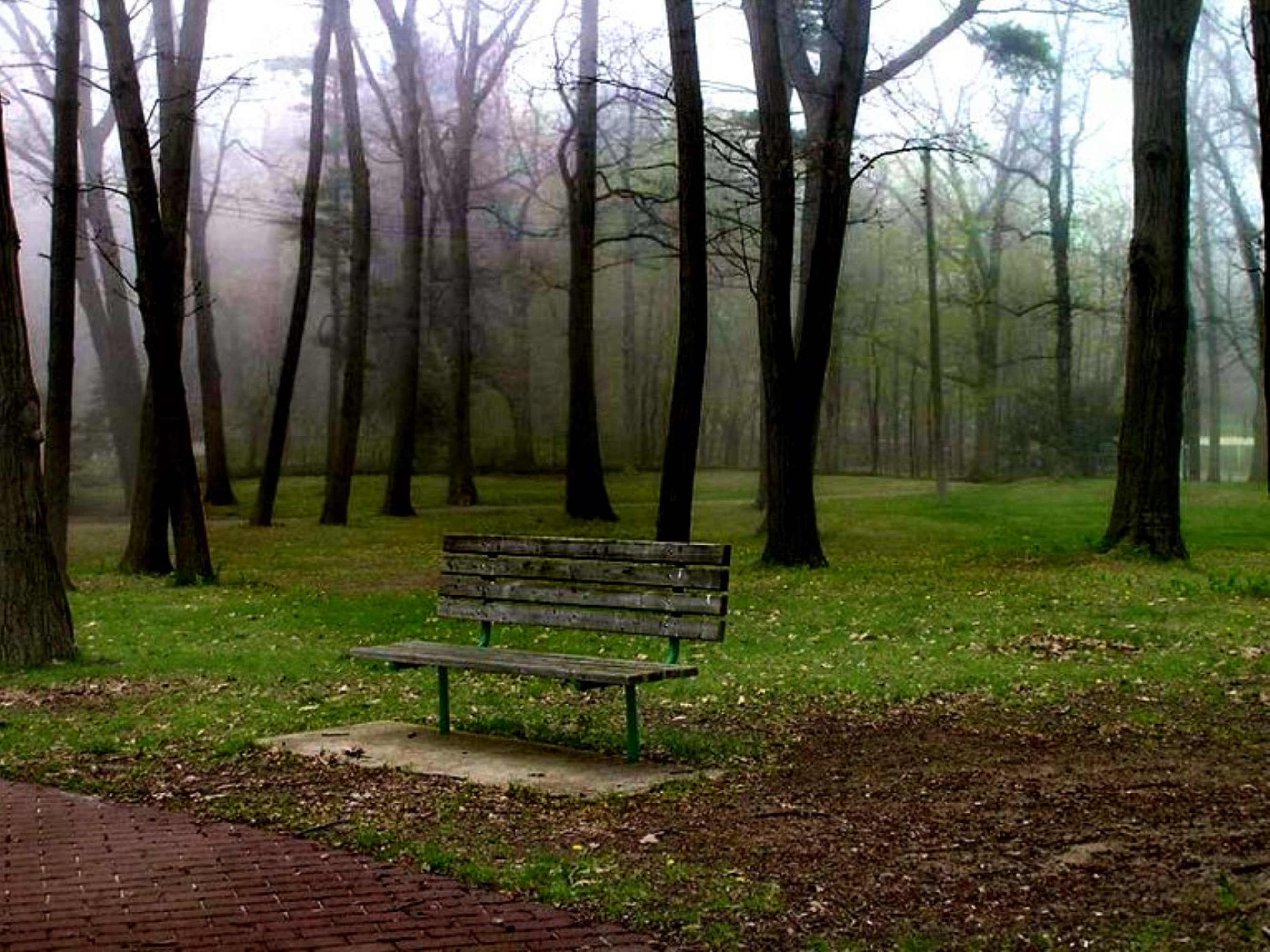} &
\includegraphics[width=0.1\textwidth]{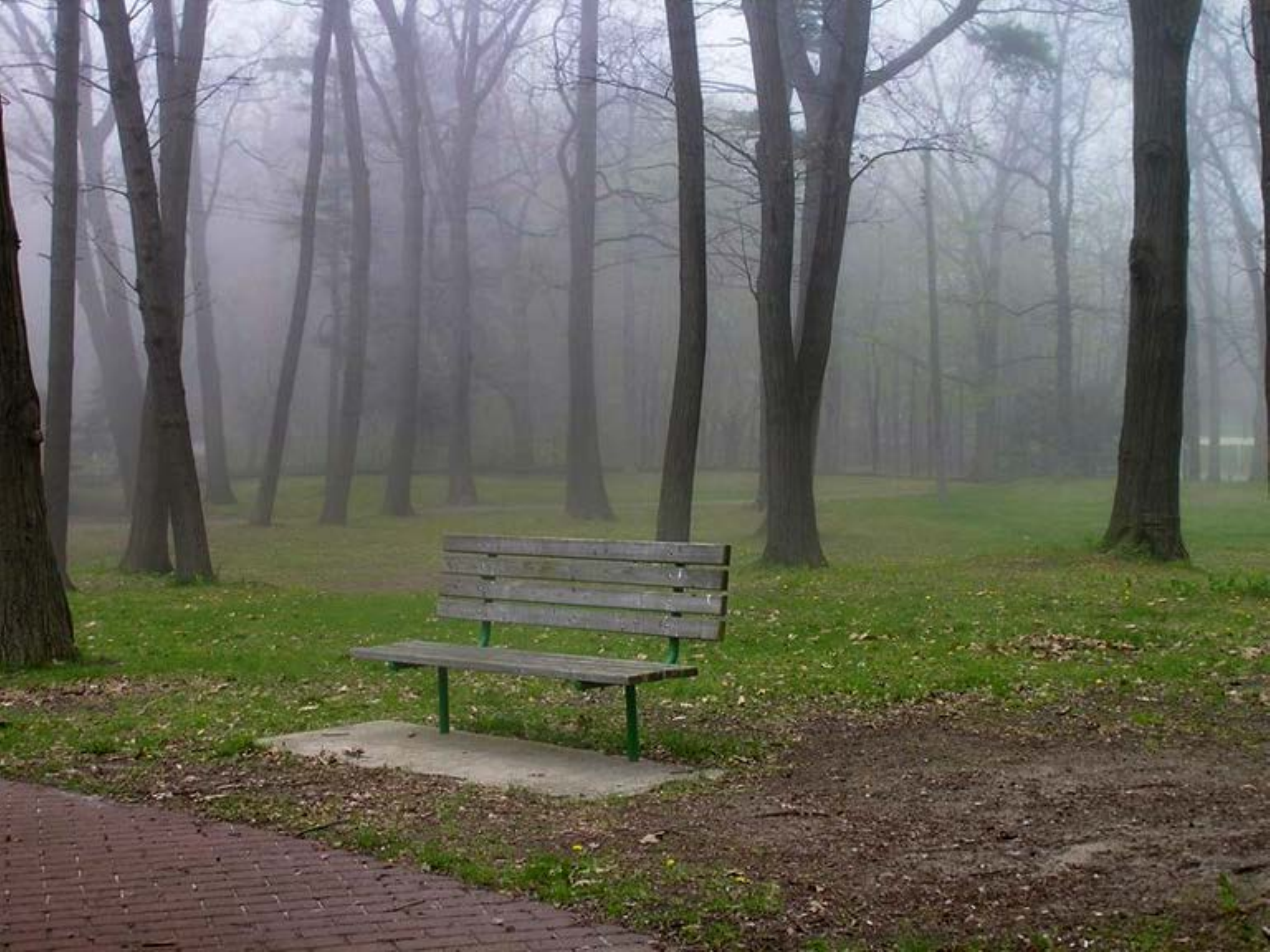} &
\includegraphics[width=0.1\textwidth]{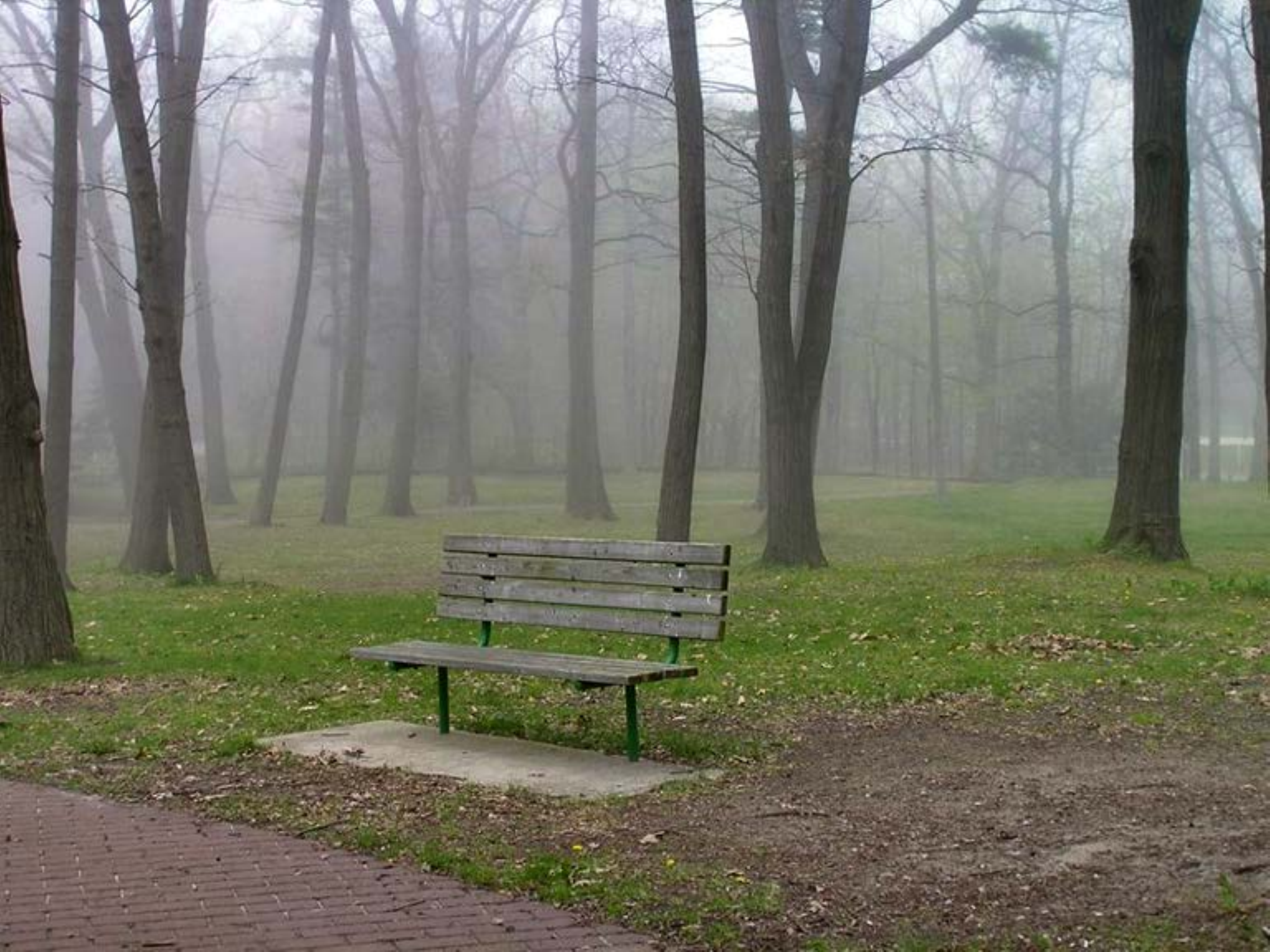} &
\includegraphics[width=0.1\textwidth]{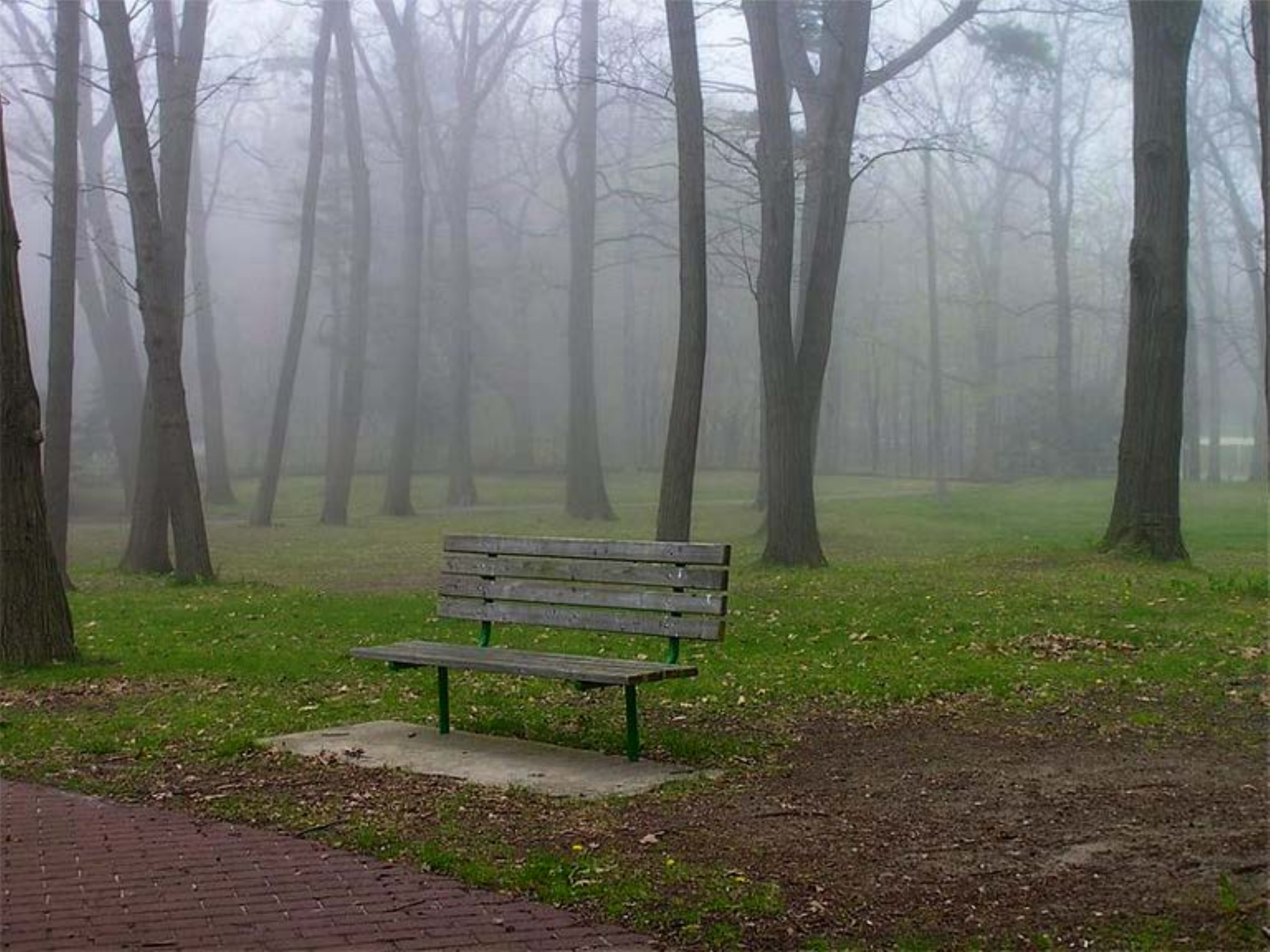} &
\includegraphics[width=0.1\textwidth]{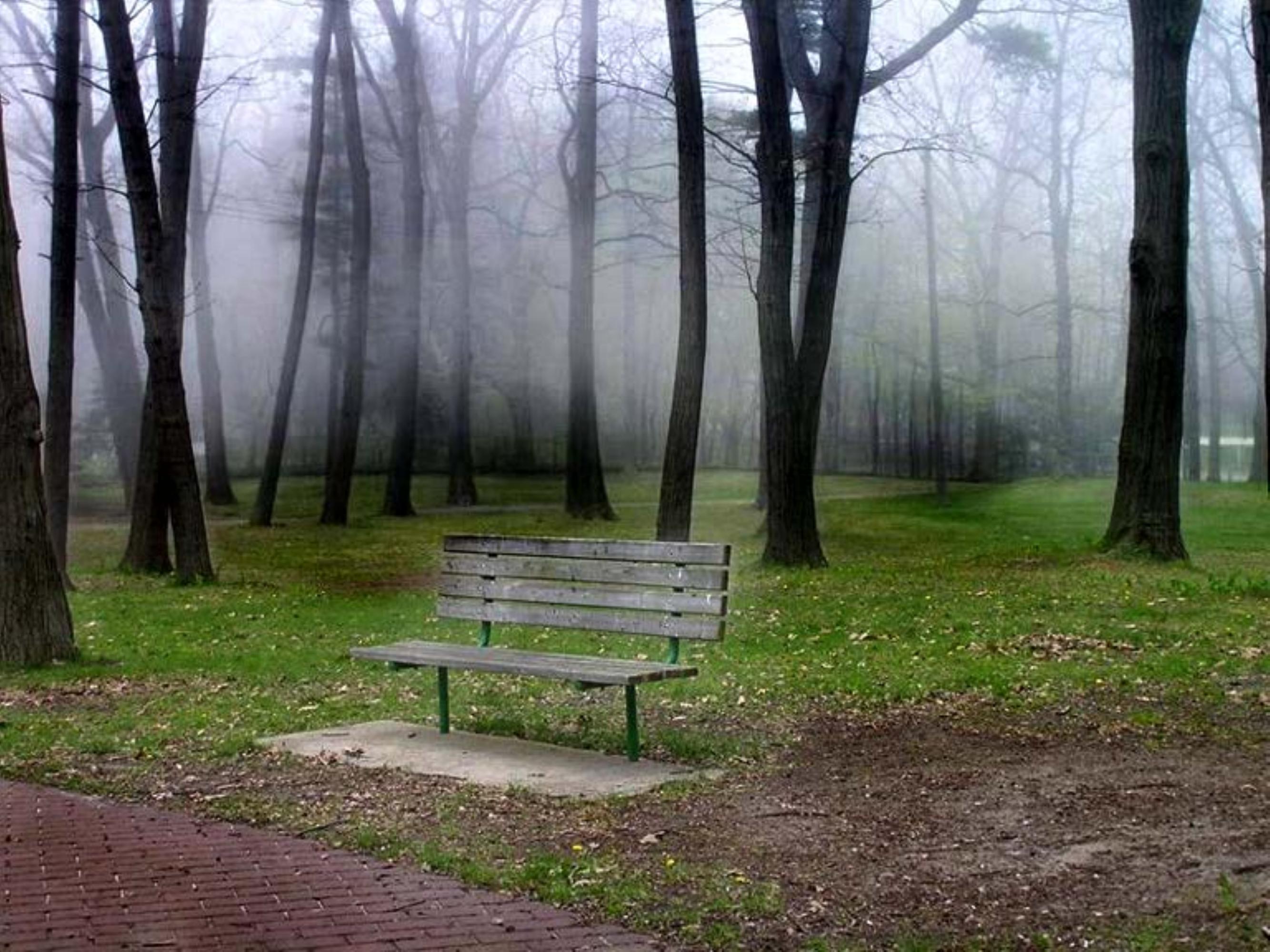} &
\includegraphics[width=0.1\textwidth]{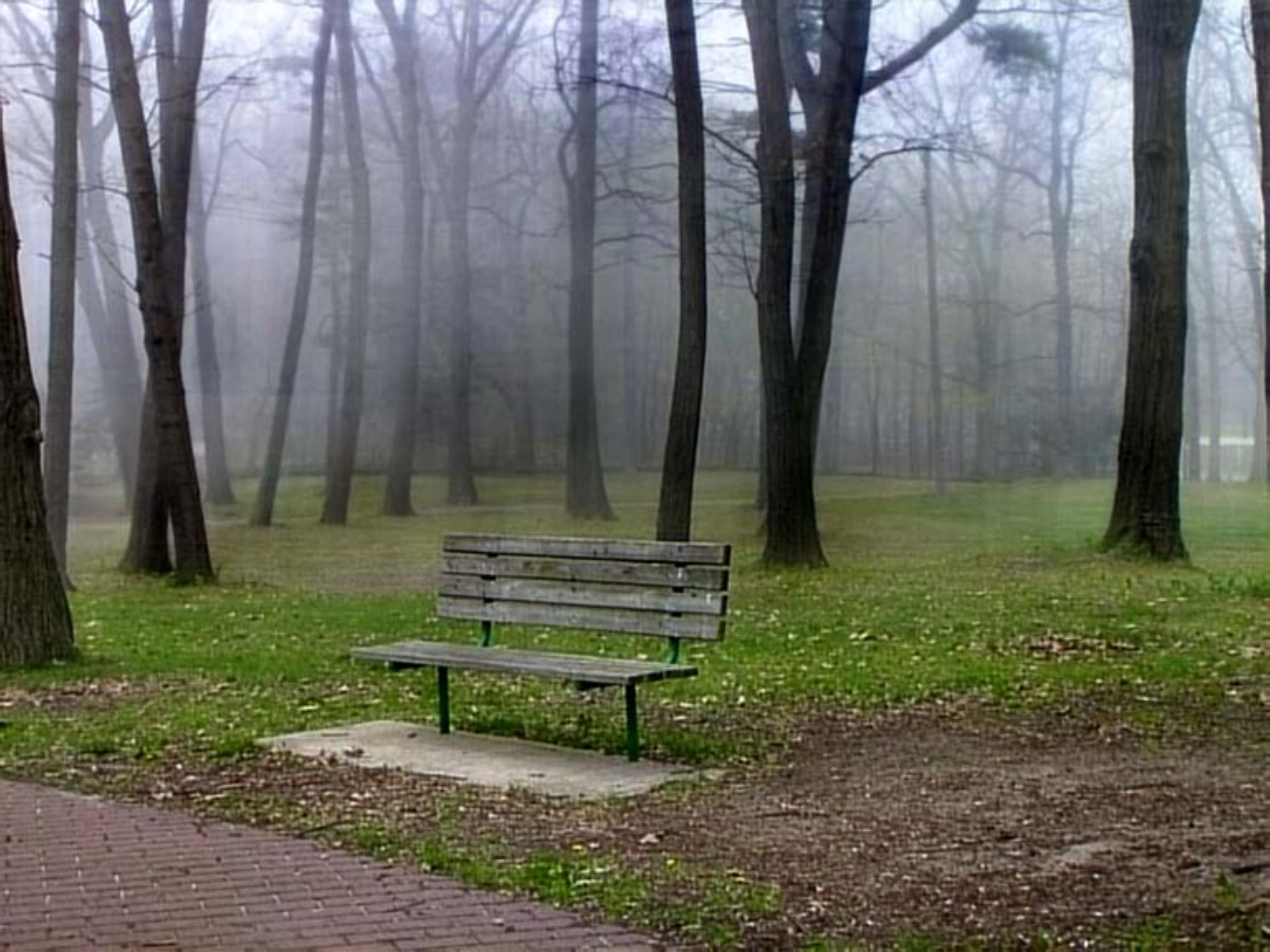} \\

\includegraphics[width=0.1\textwidth]{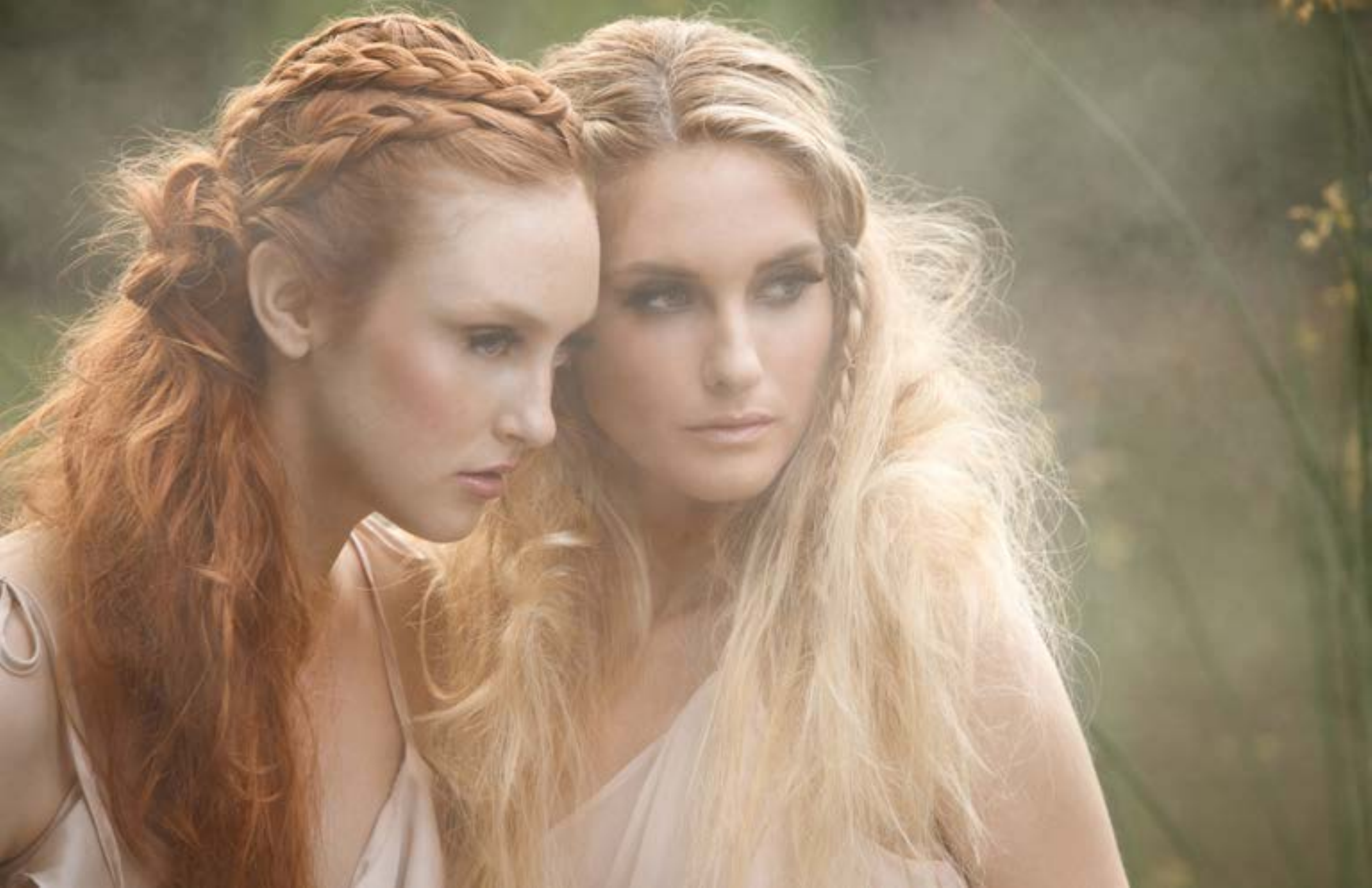} &
\includegraphics[width=0.1\textwidth]{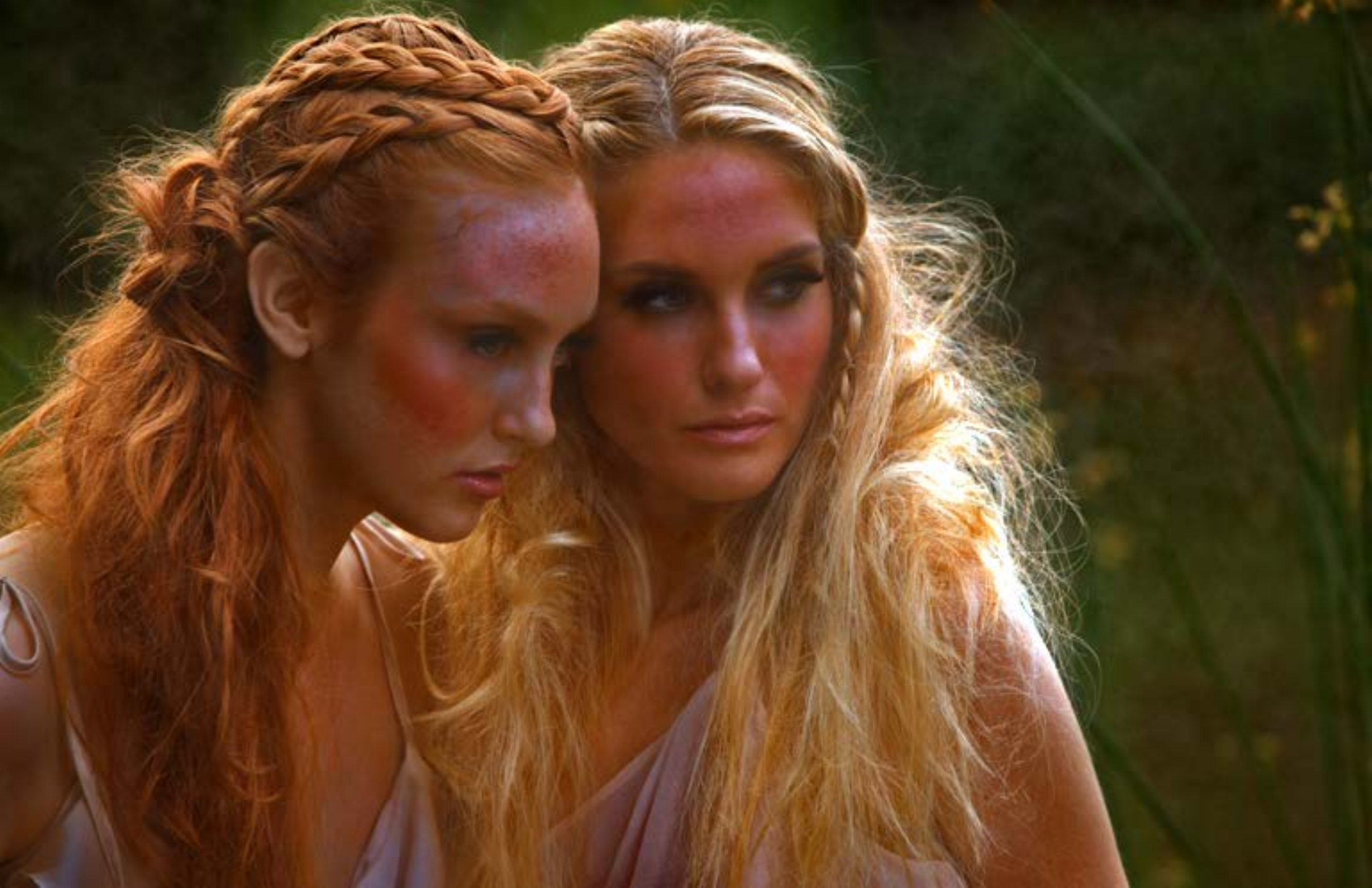} &
\includegraphics[width=0.1\textwidth]{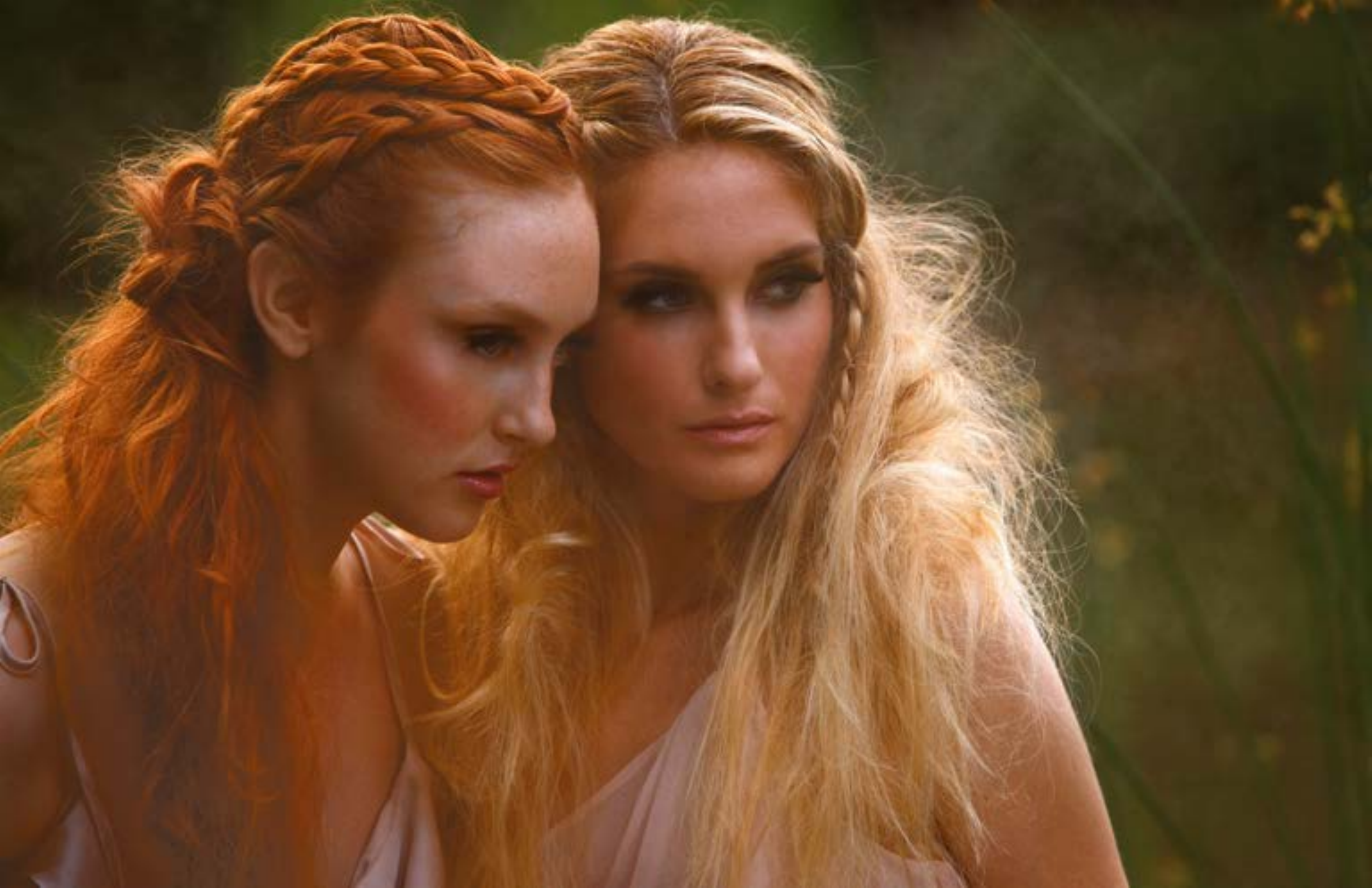} &
\includegraphics[width=0.1\textwidth]{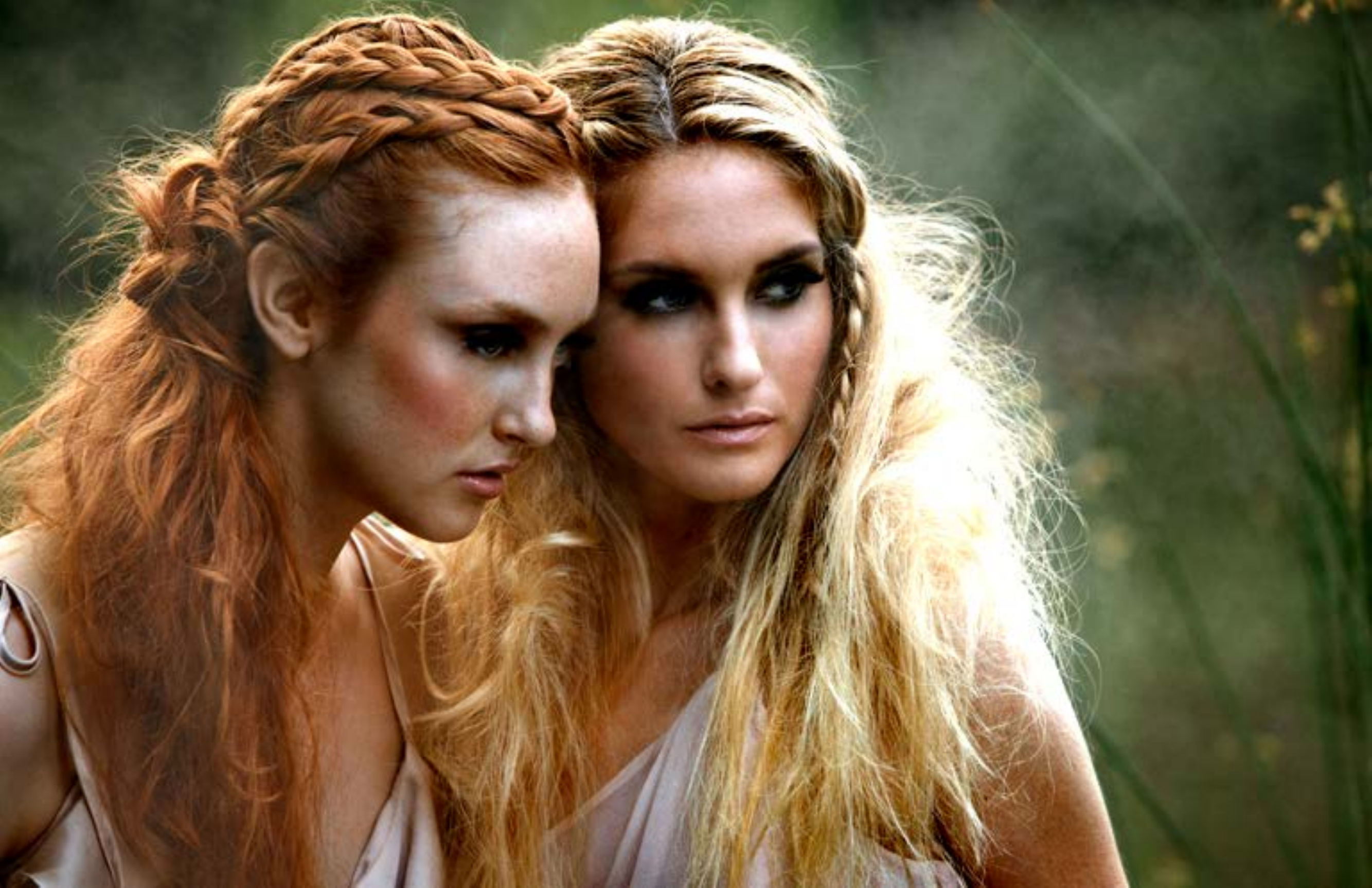} &
\includegraphics[width=0.1\textwidth]{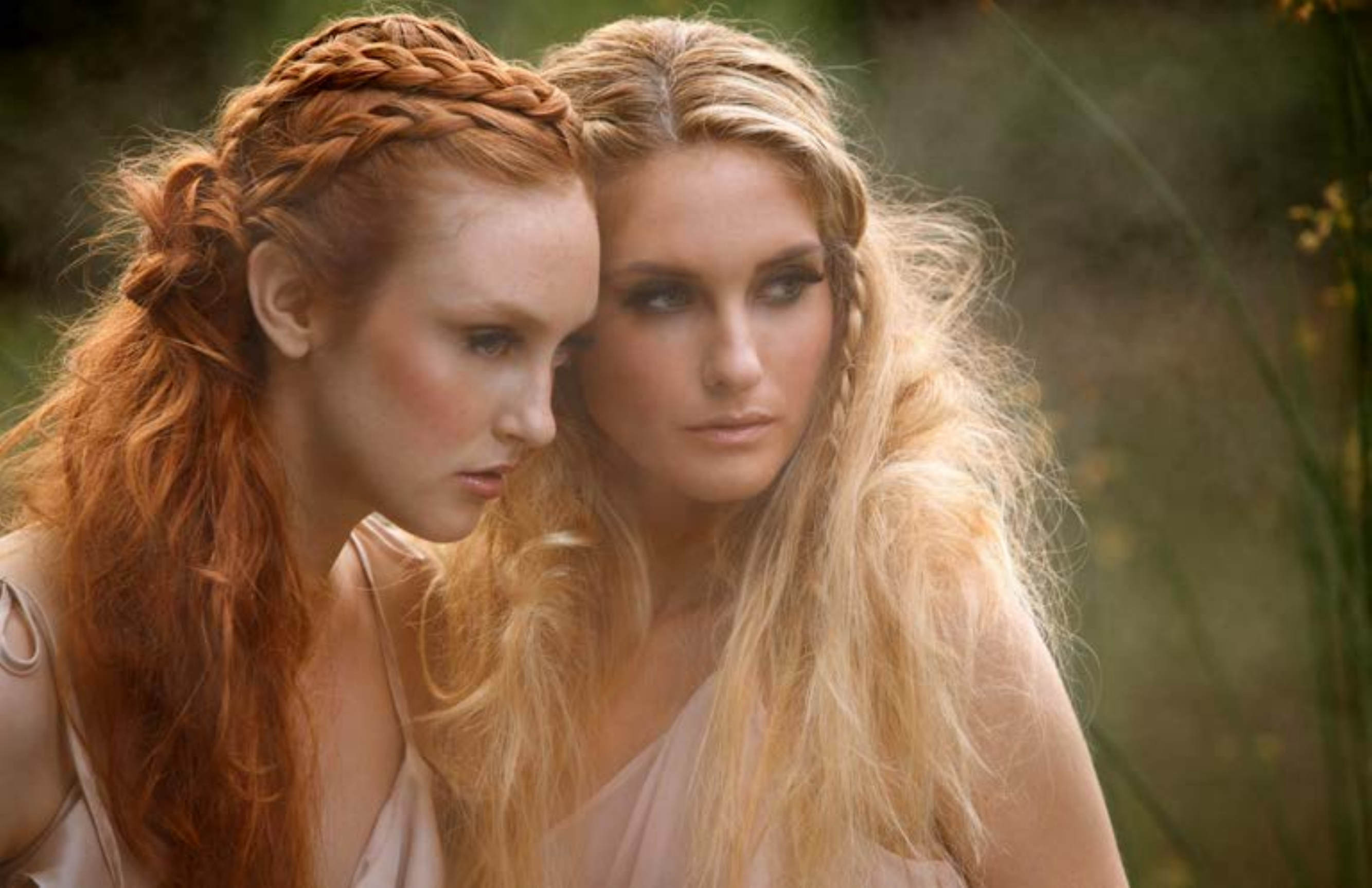} &
\includegraphics[width=0.1\textwidth]{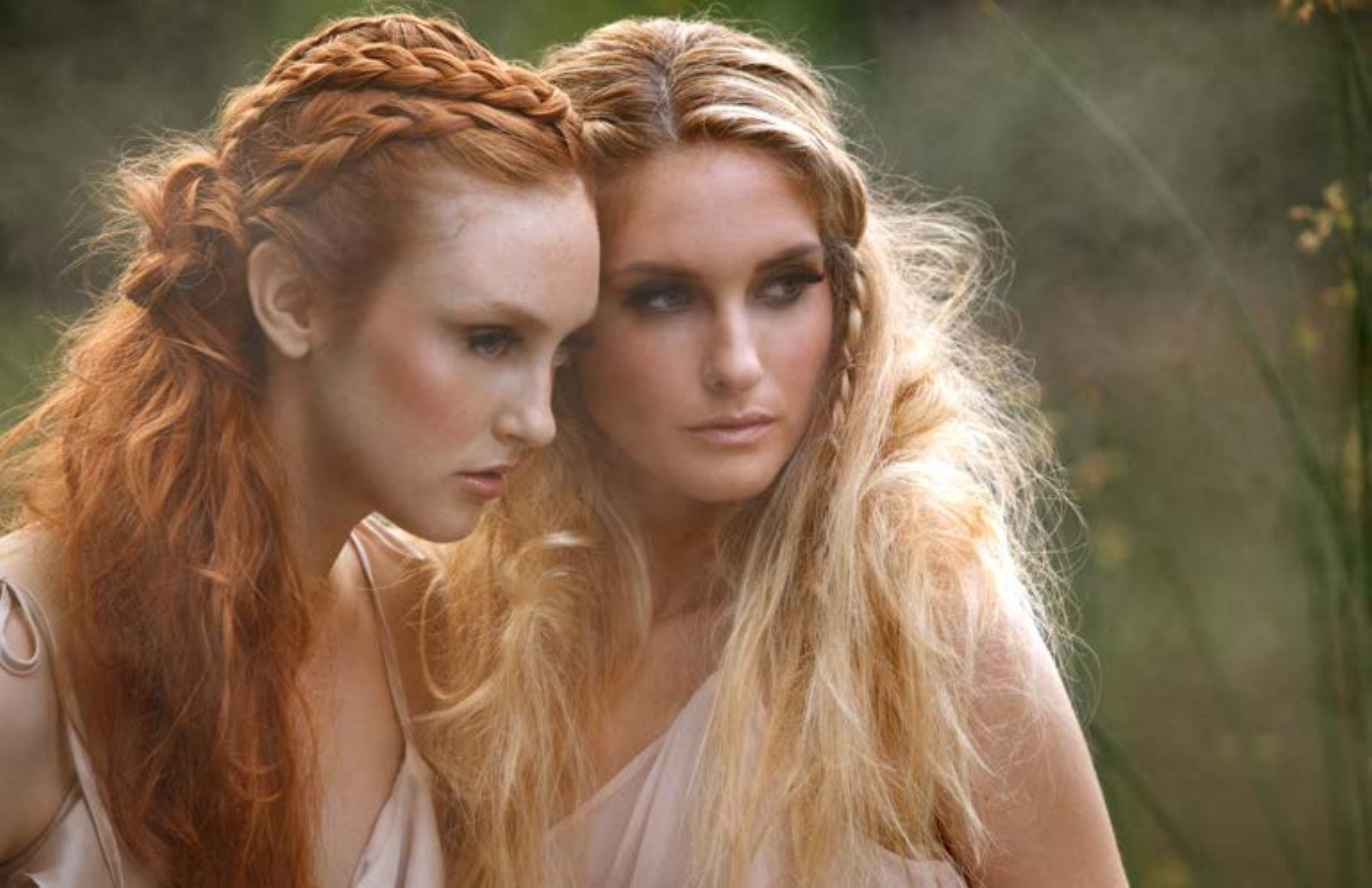} &
\includegraphics[width=0.1\textwidth]{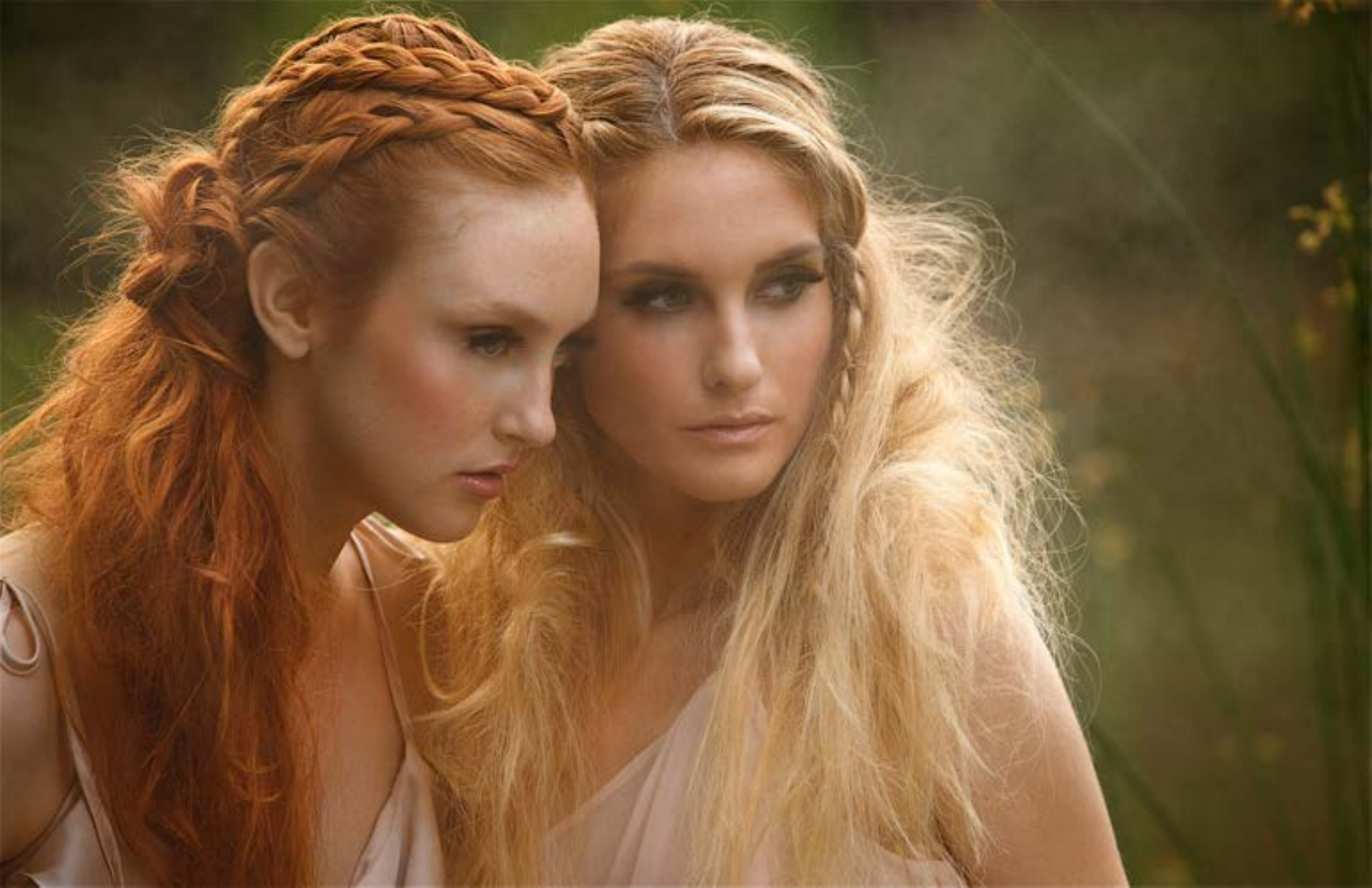} &
\includegraphics[width=0.1\textwidth]{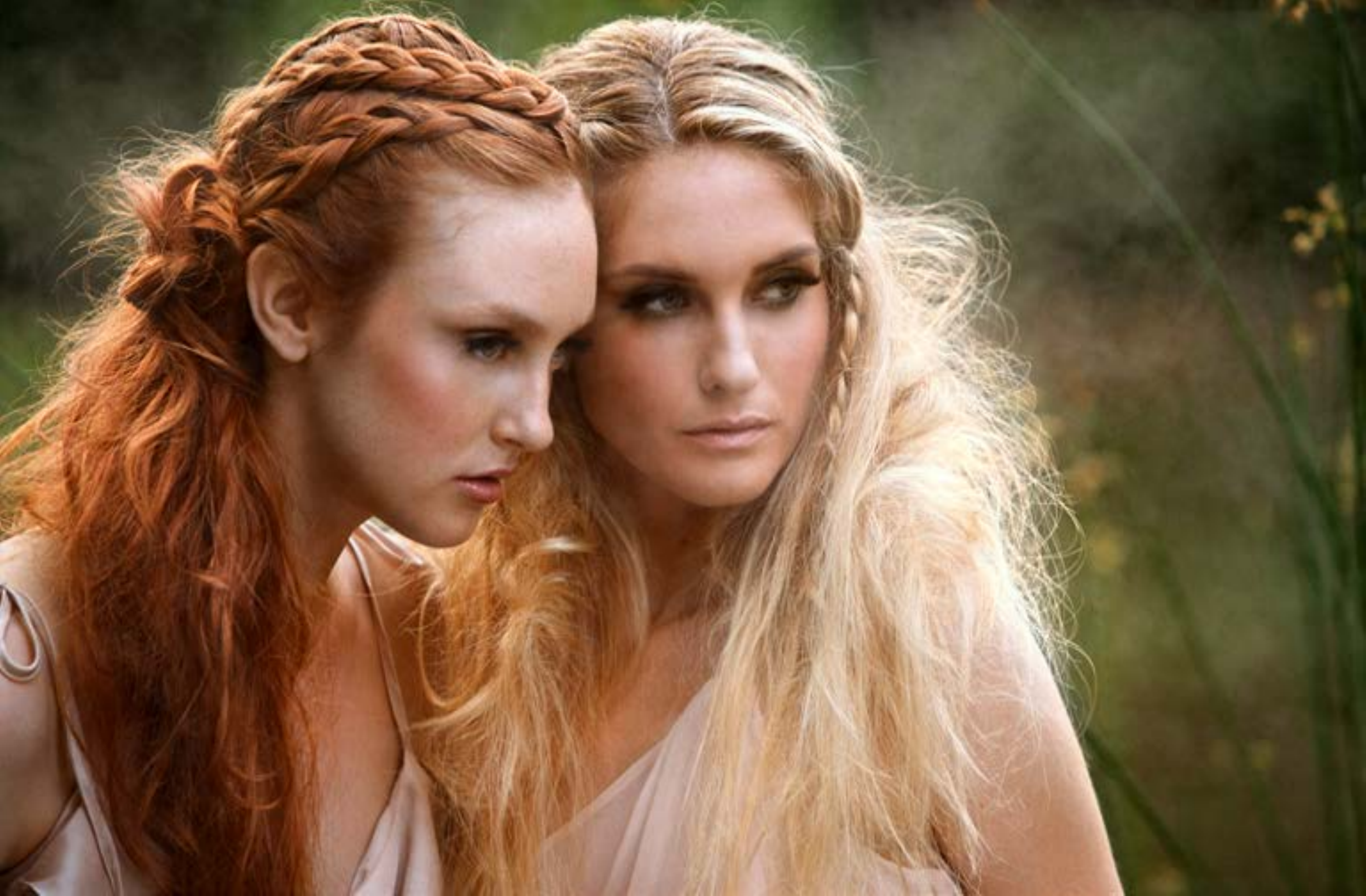} &
\includegraphics[width=0.1\textwidth]{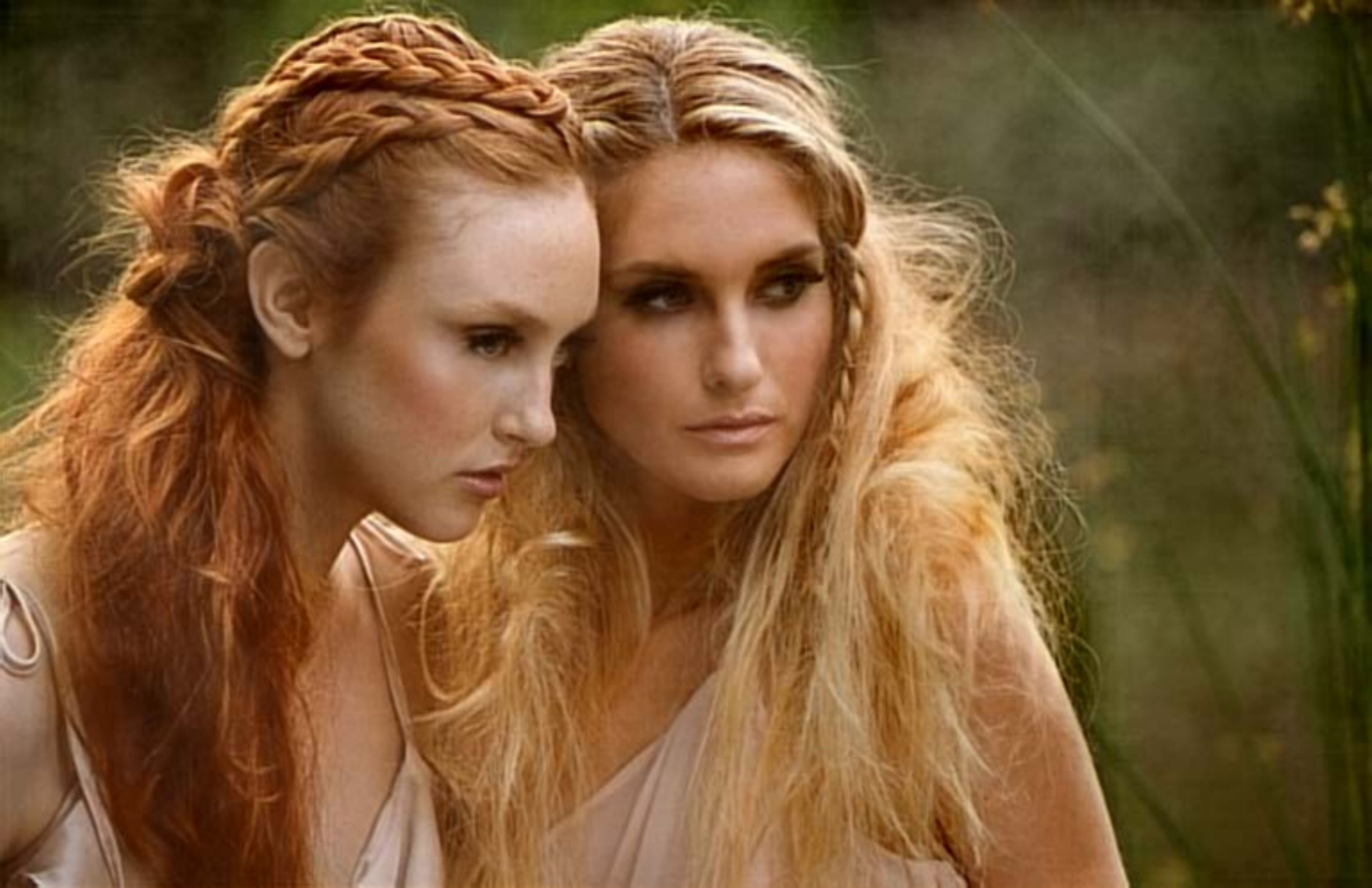} \\

\includegraphics[width=0.1\textwidth]{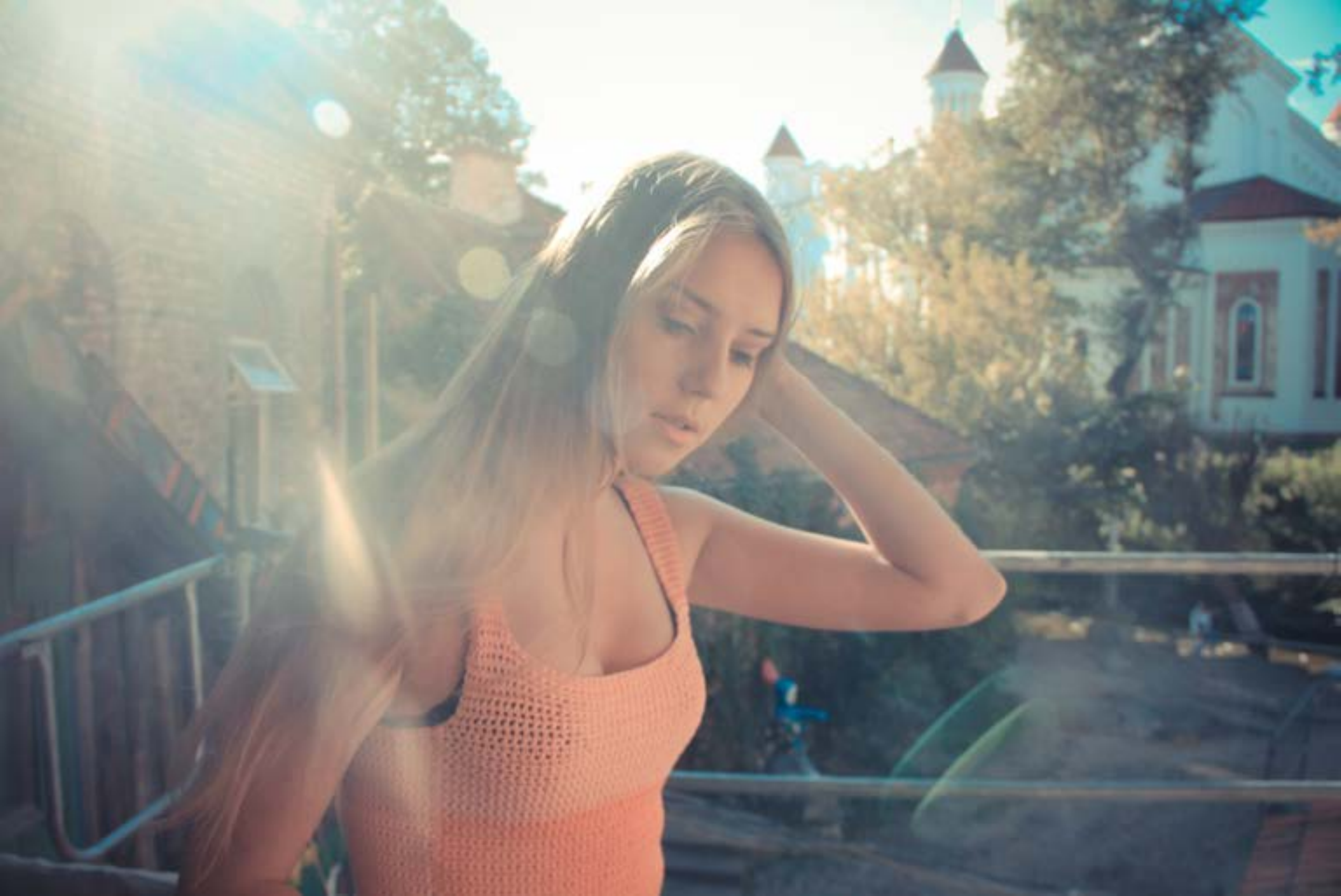} &
\includegraphics[width=0.1\textwidth]{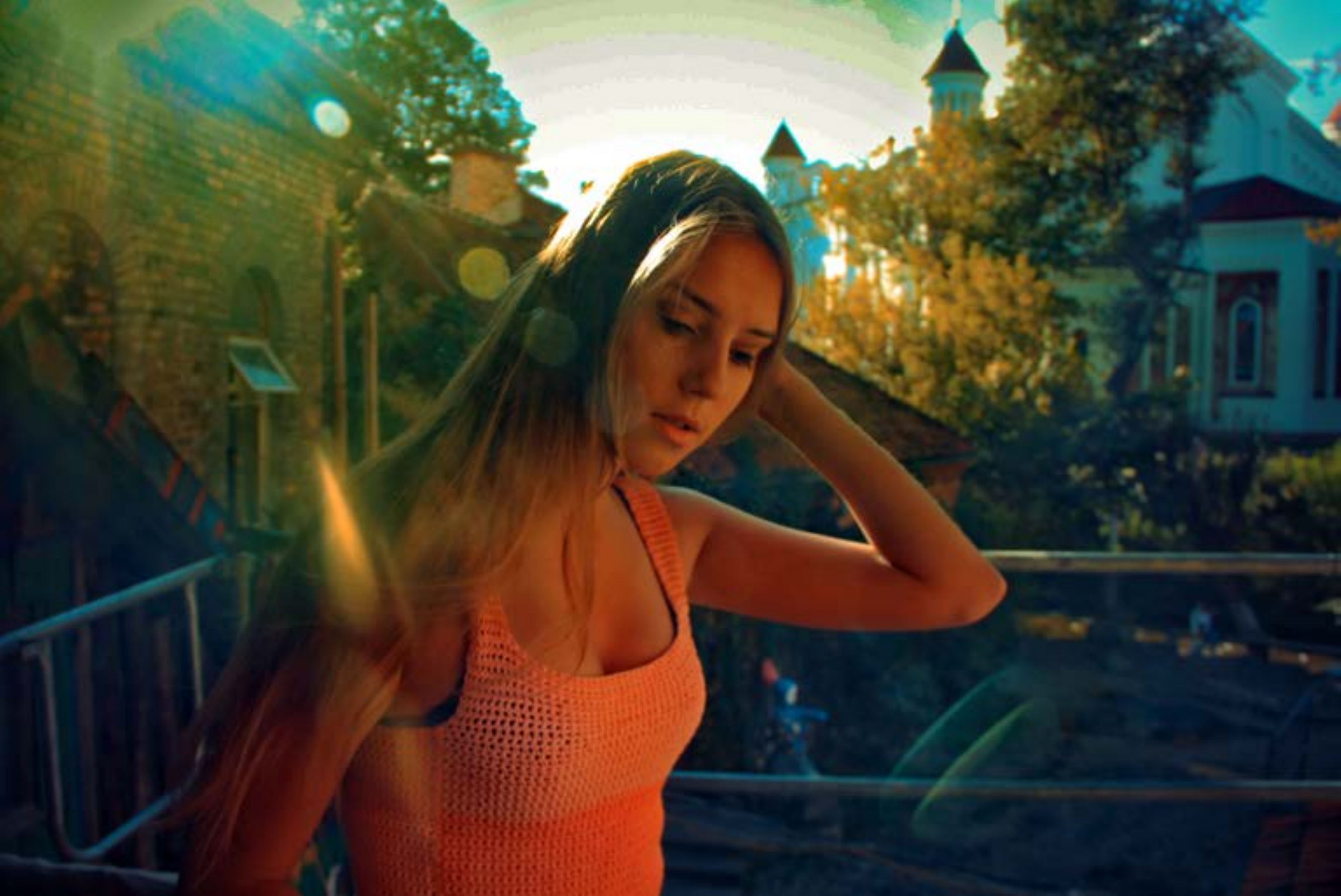} &
\includegraphics[width=0.1\textwidth]{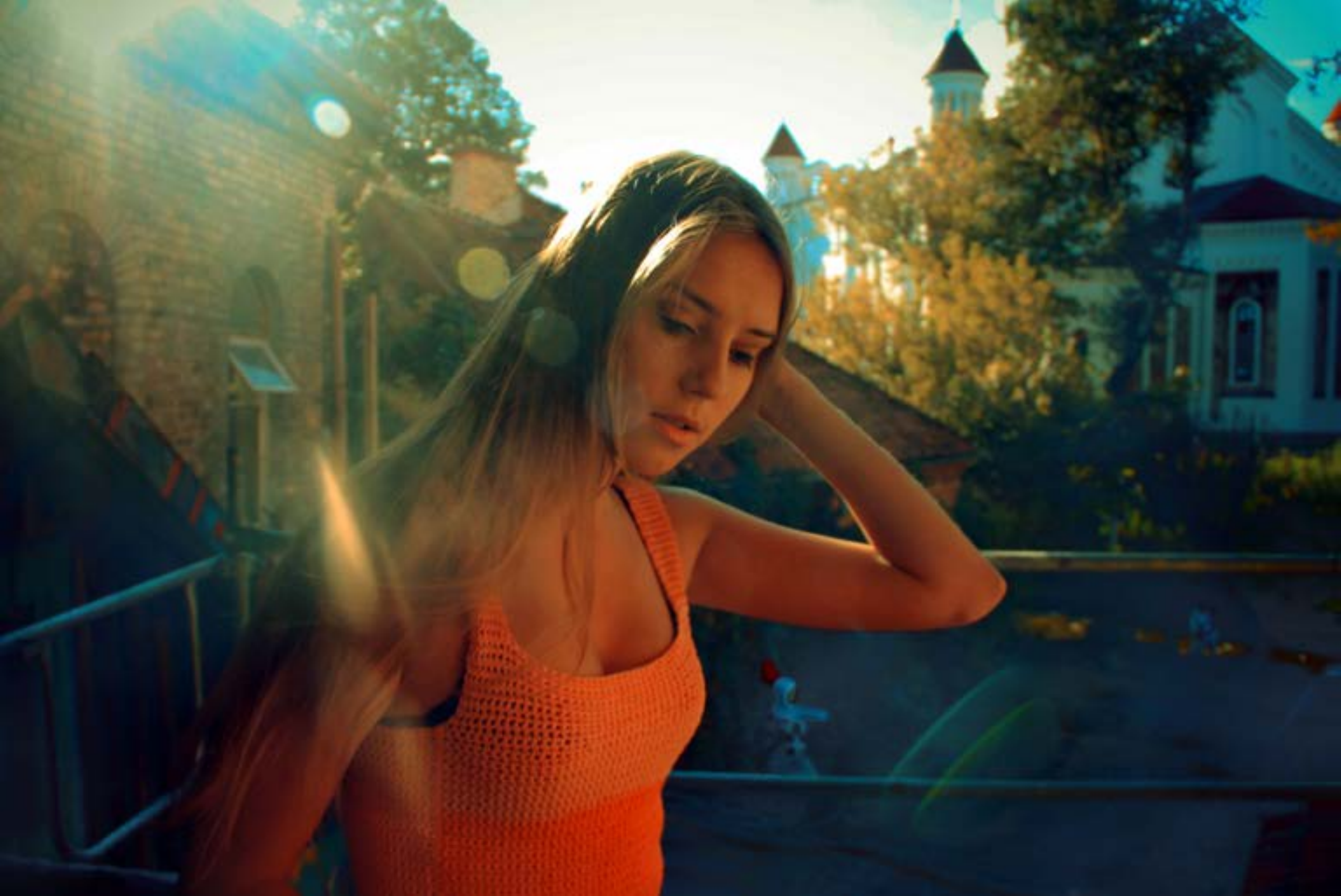} &
\includegraphics[width=0.1\textwidth]{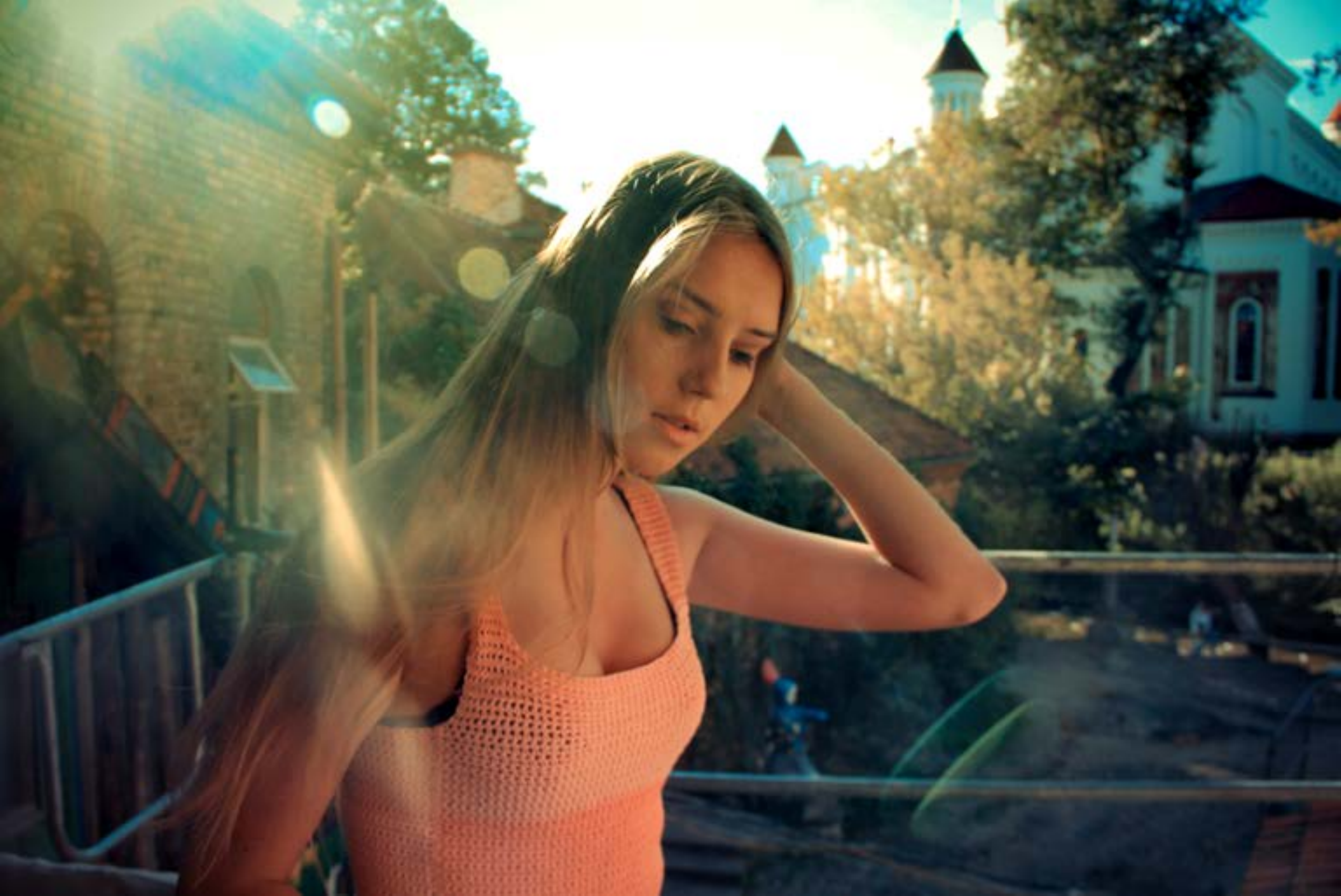} &
\includegraphics[width=0.1\textwidth]{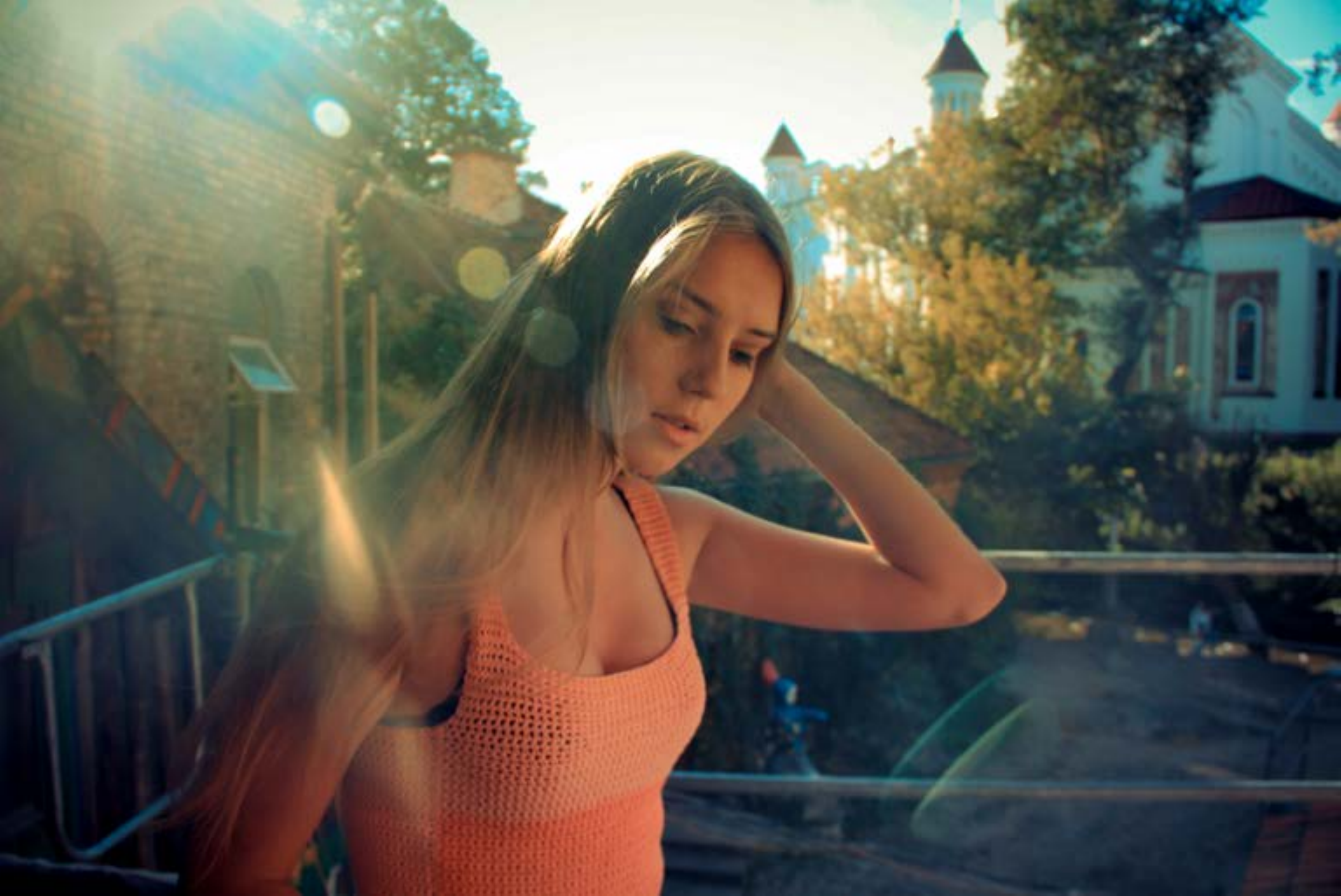} &
\includegraphics[width=0.1\textwidth]{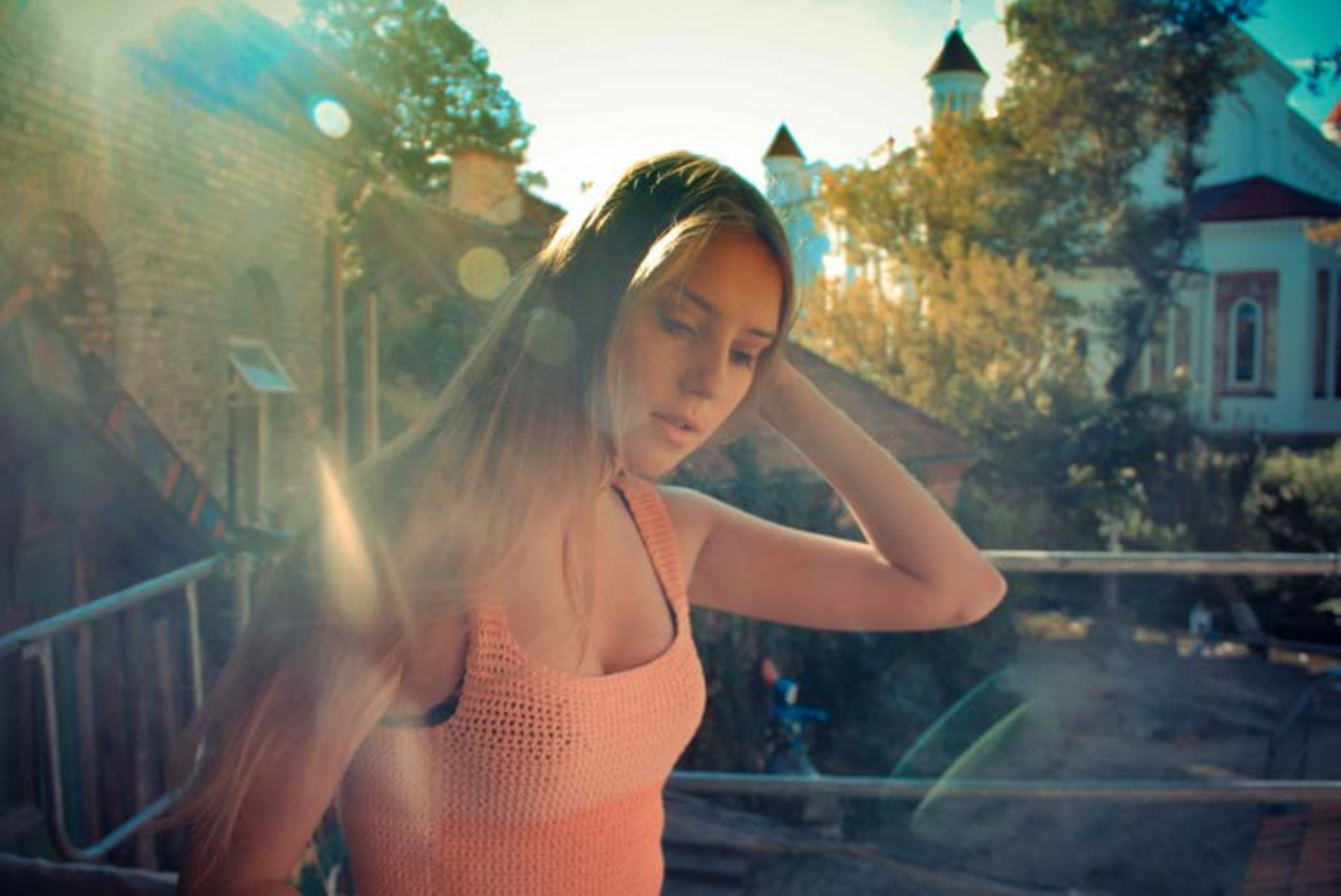} &
\includegraphics[width=0.1\textwidth]{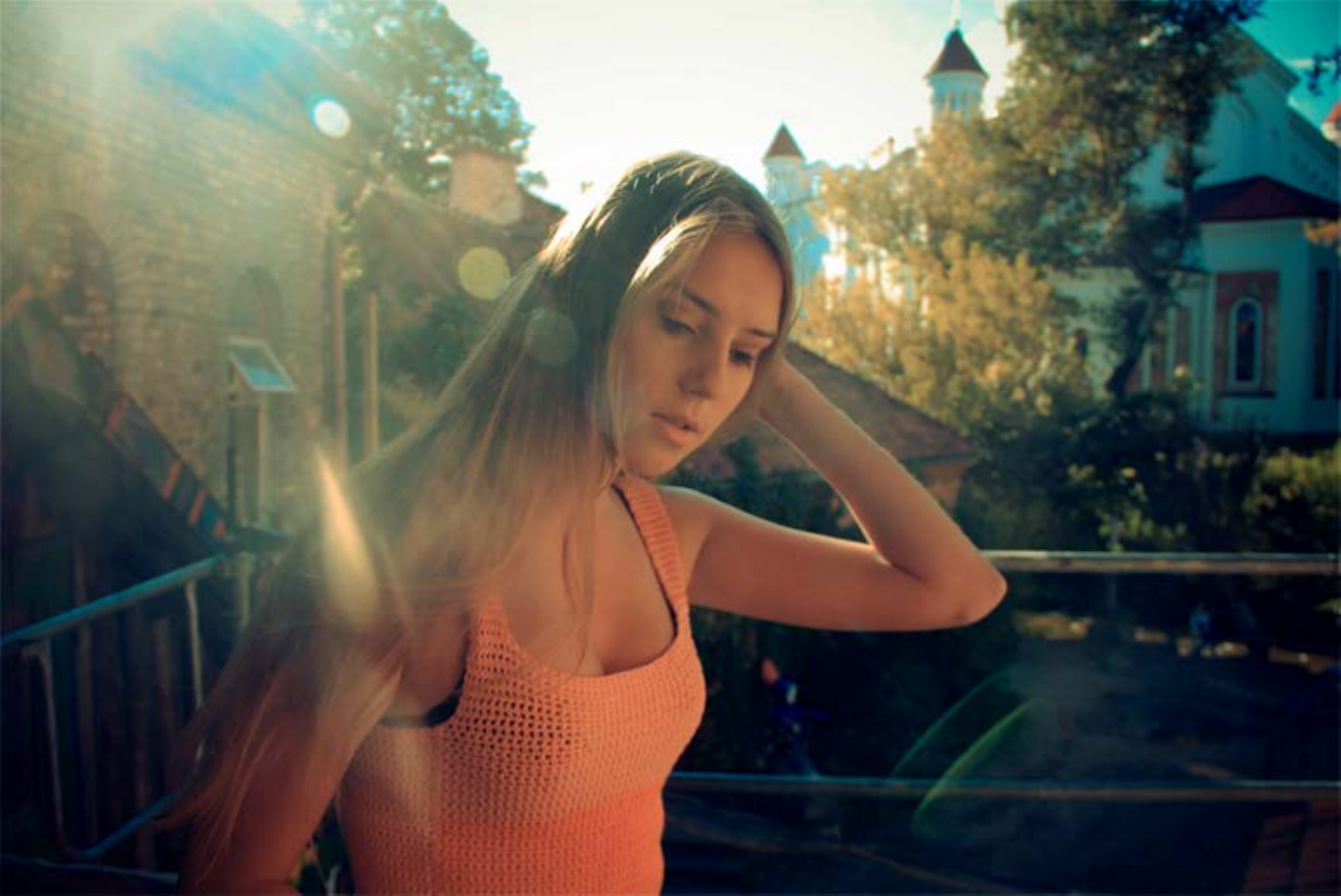} &
\includegraphics[width=0.1\textwidth]{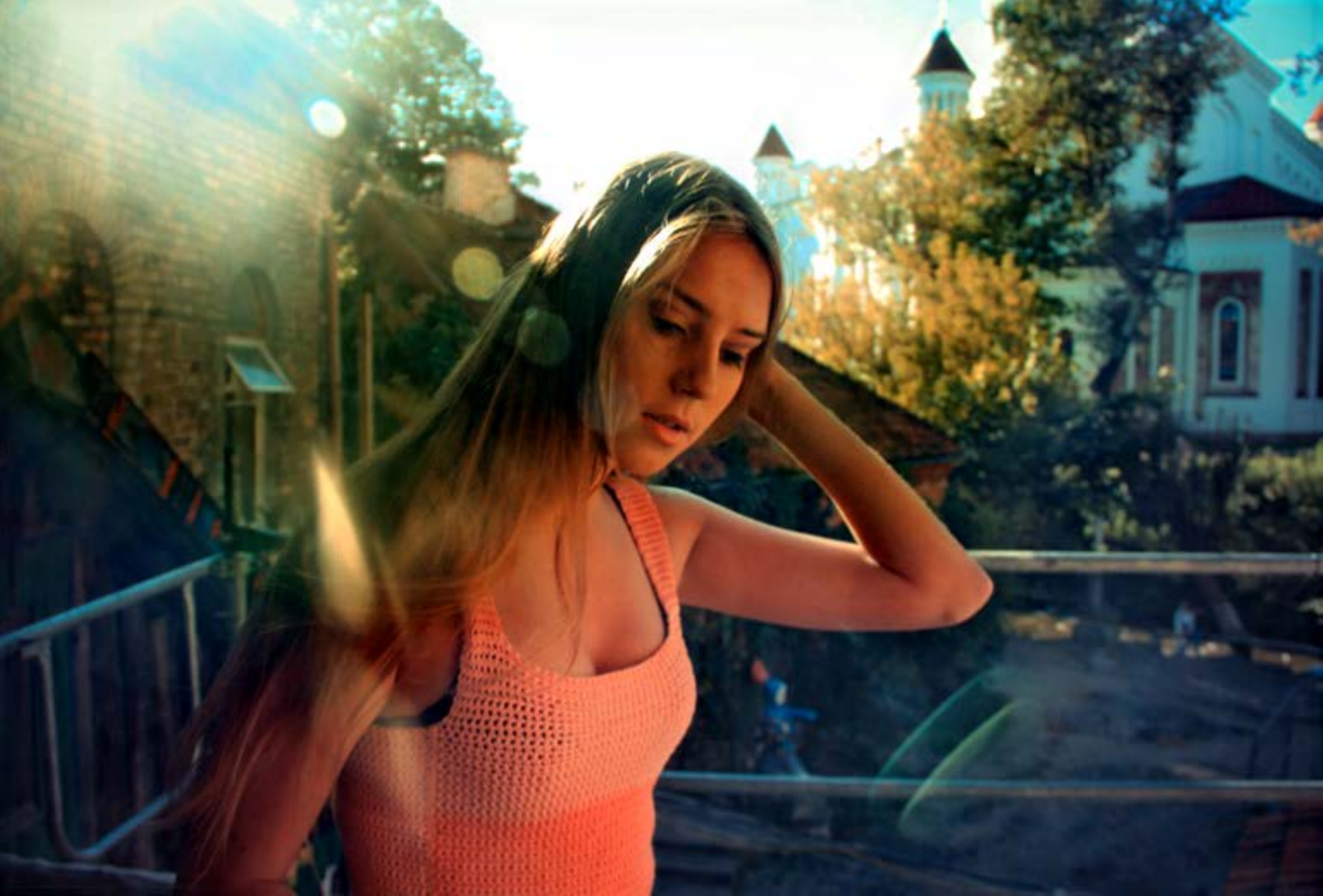} &
\includegraphics[width=0.1\textwidth]{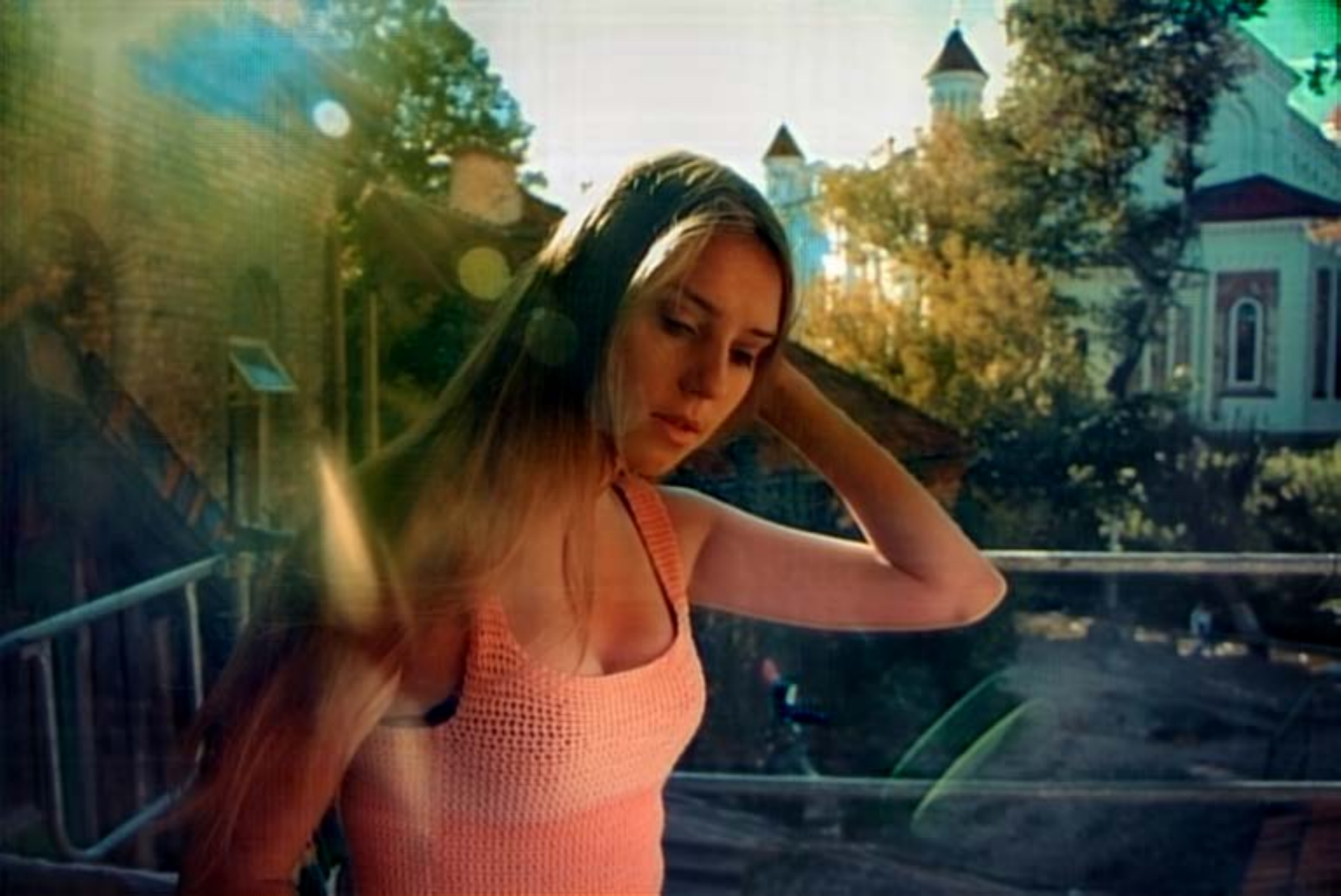} \\

\includegraphics[width=0.1\textwidth]{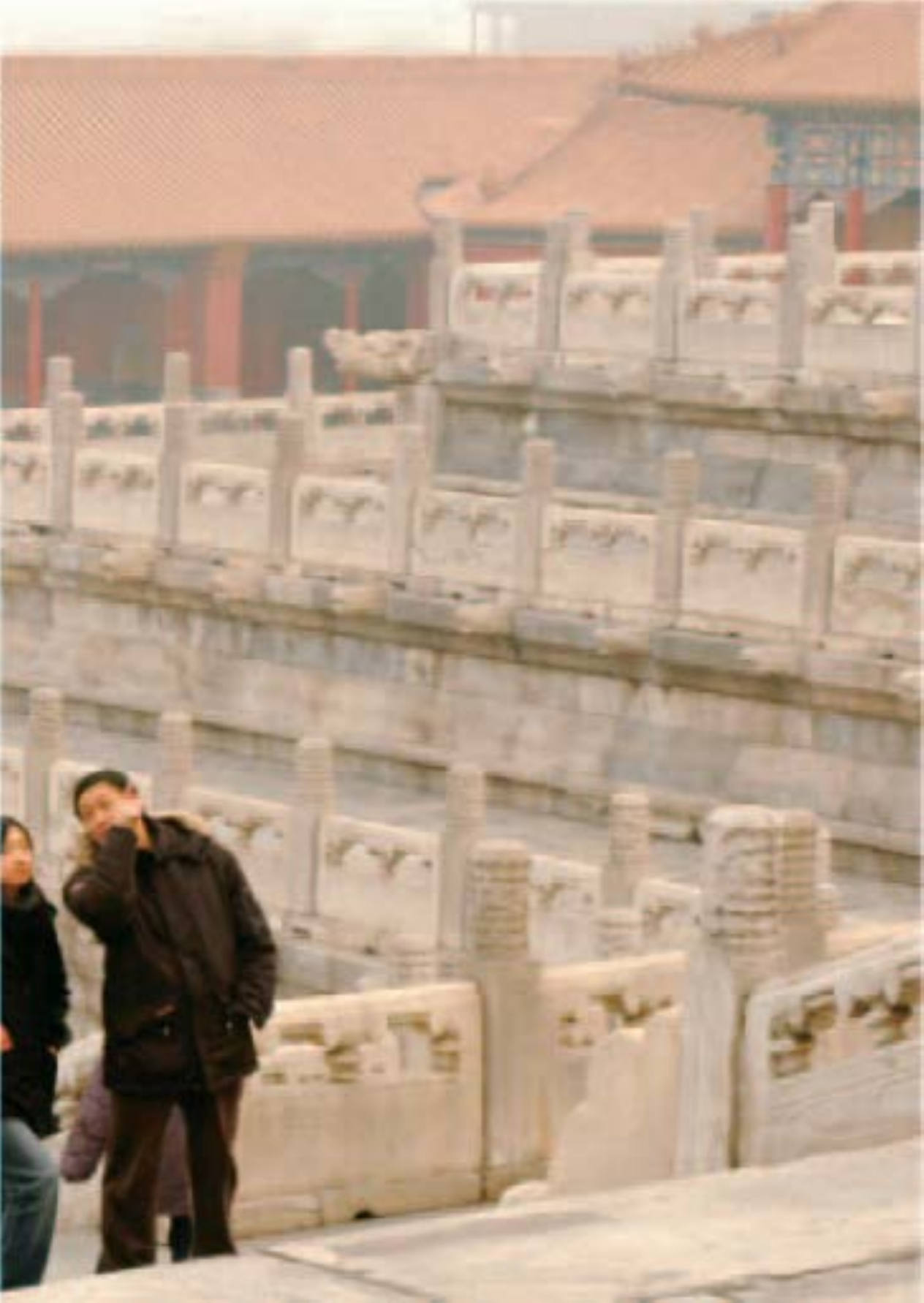} &
\includegraphics[width=0.1\textwidth]{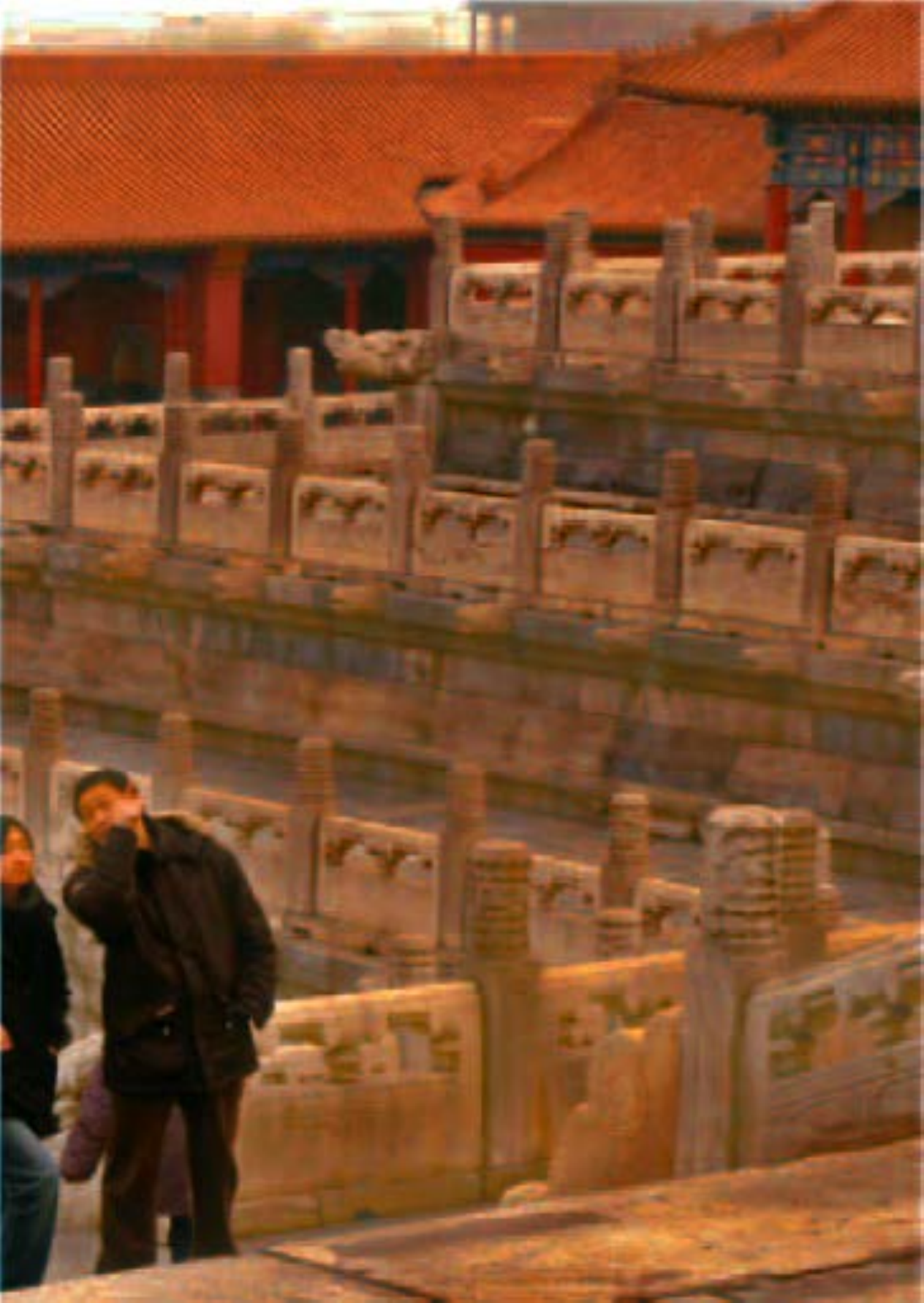} &
\includegraphics[width=0.1\textwidth]{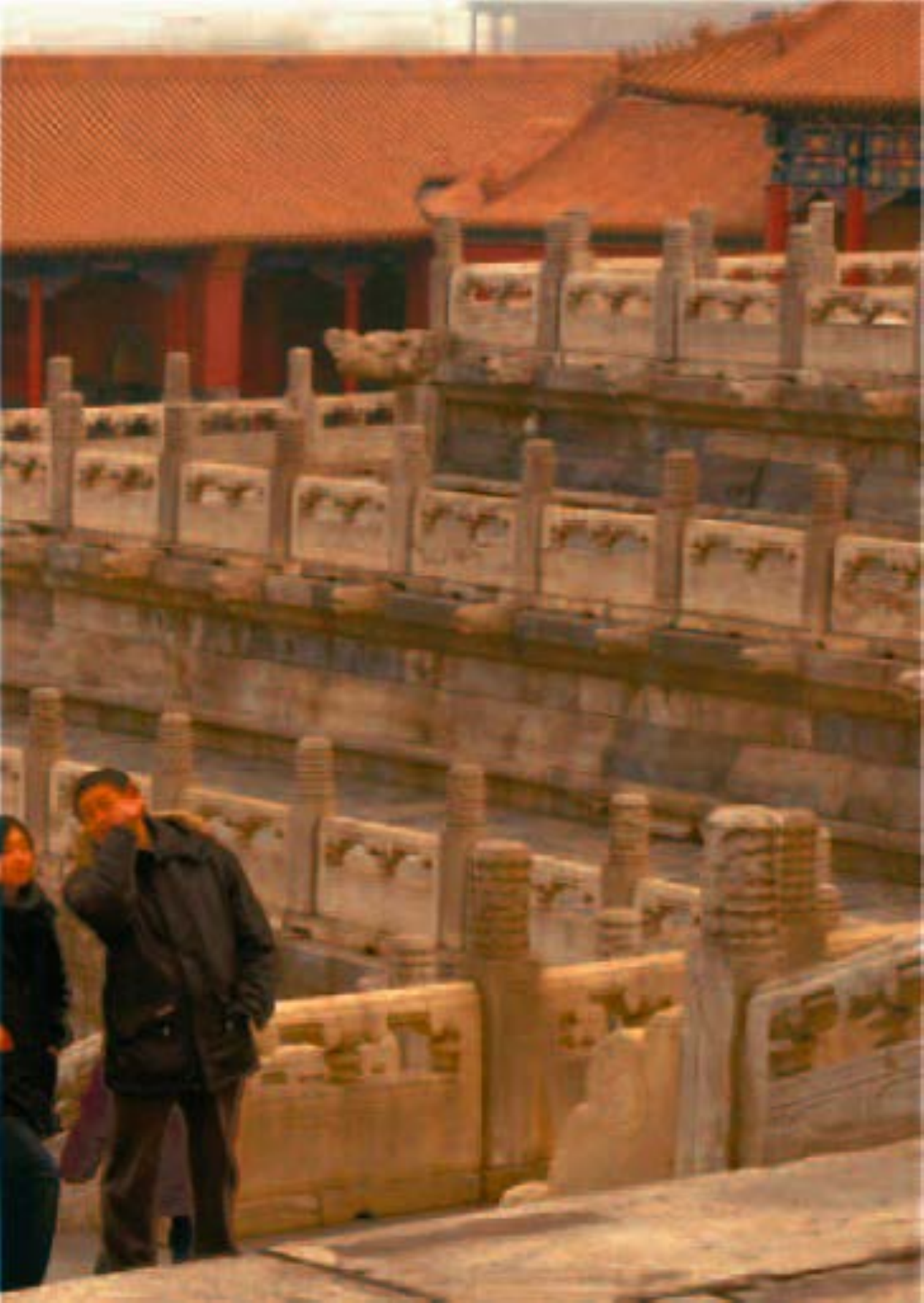} &
\includegraphics[width=0.1\textwidth]{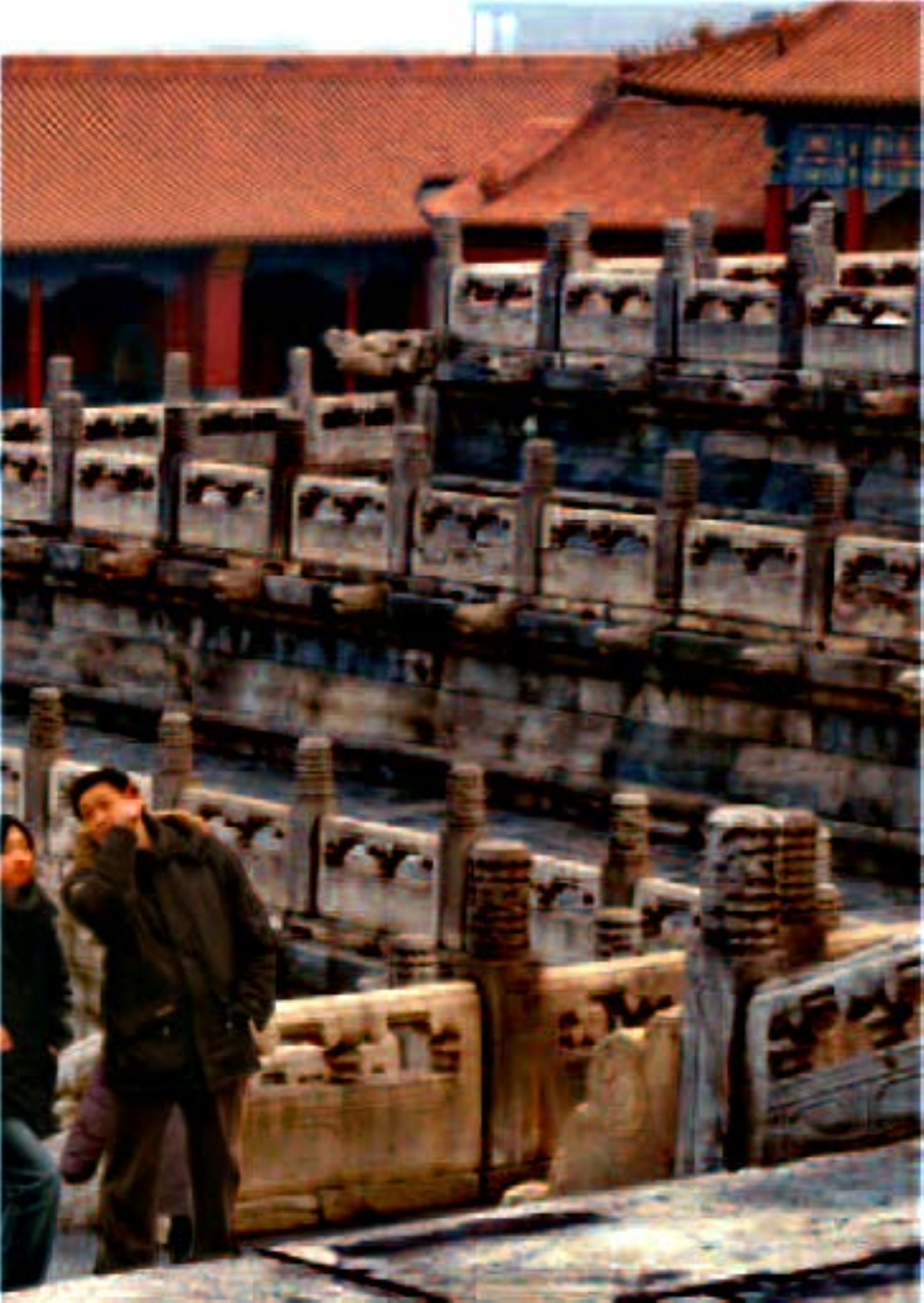} &
\includegraphics[width=0.1\textwidth]{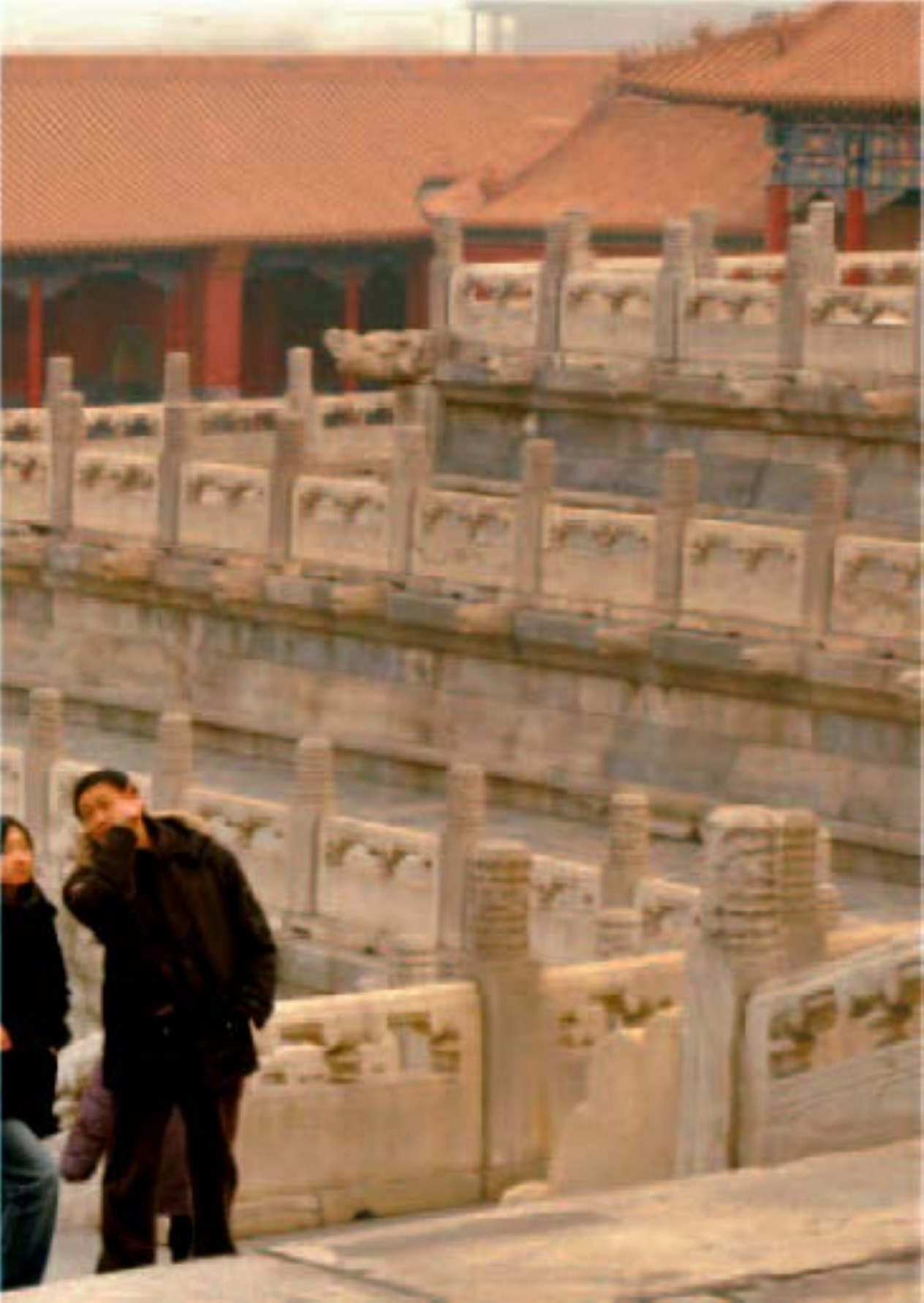} &
\includegraphics[width=0.1\textwidth]{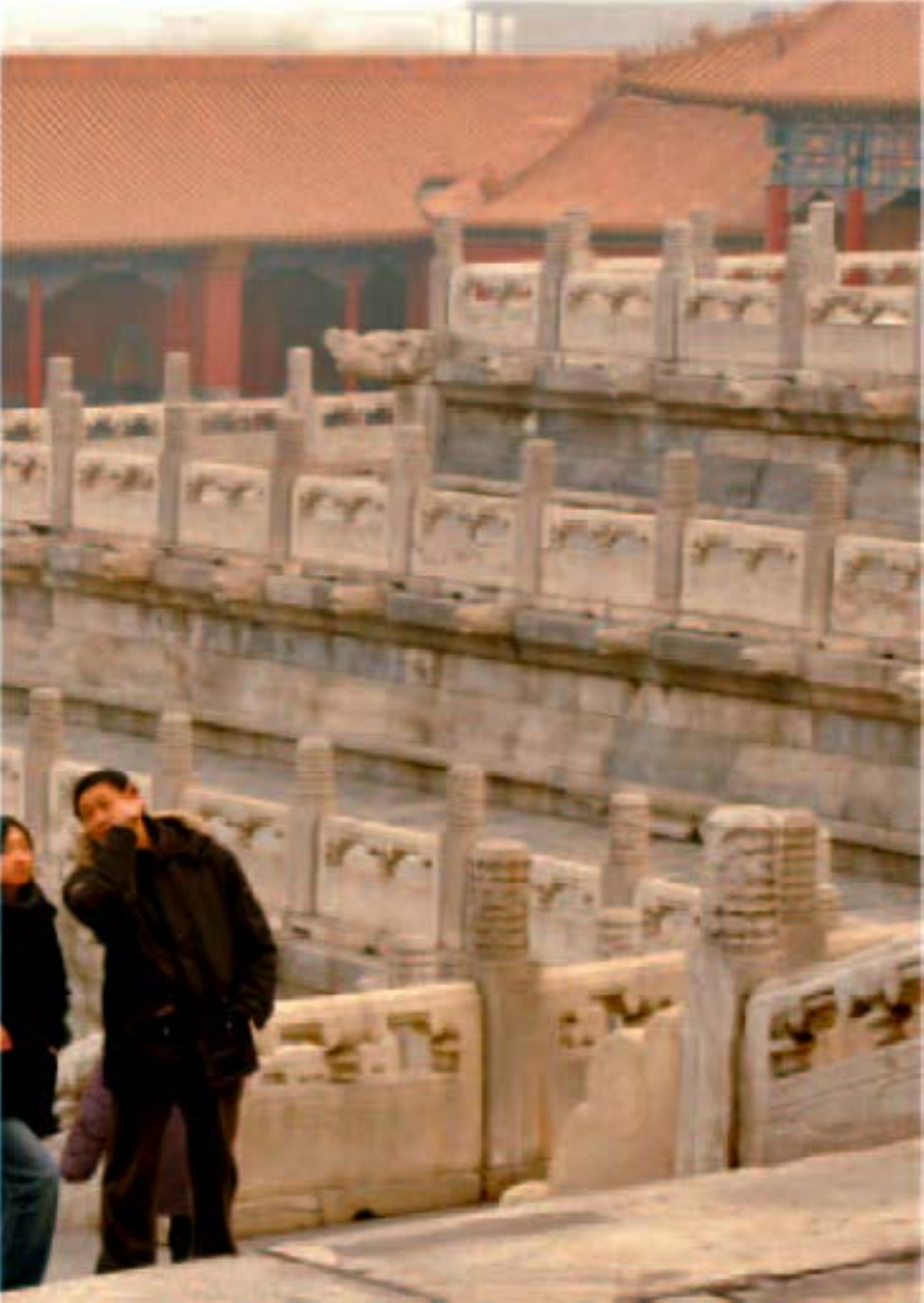} &
\includegraphics[width=0.1\textwidth]{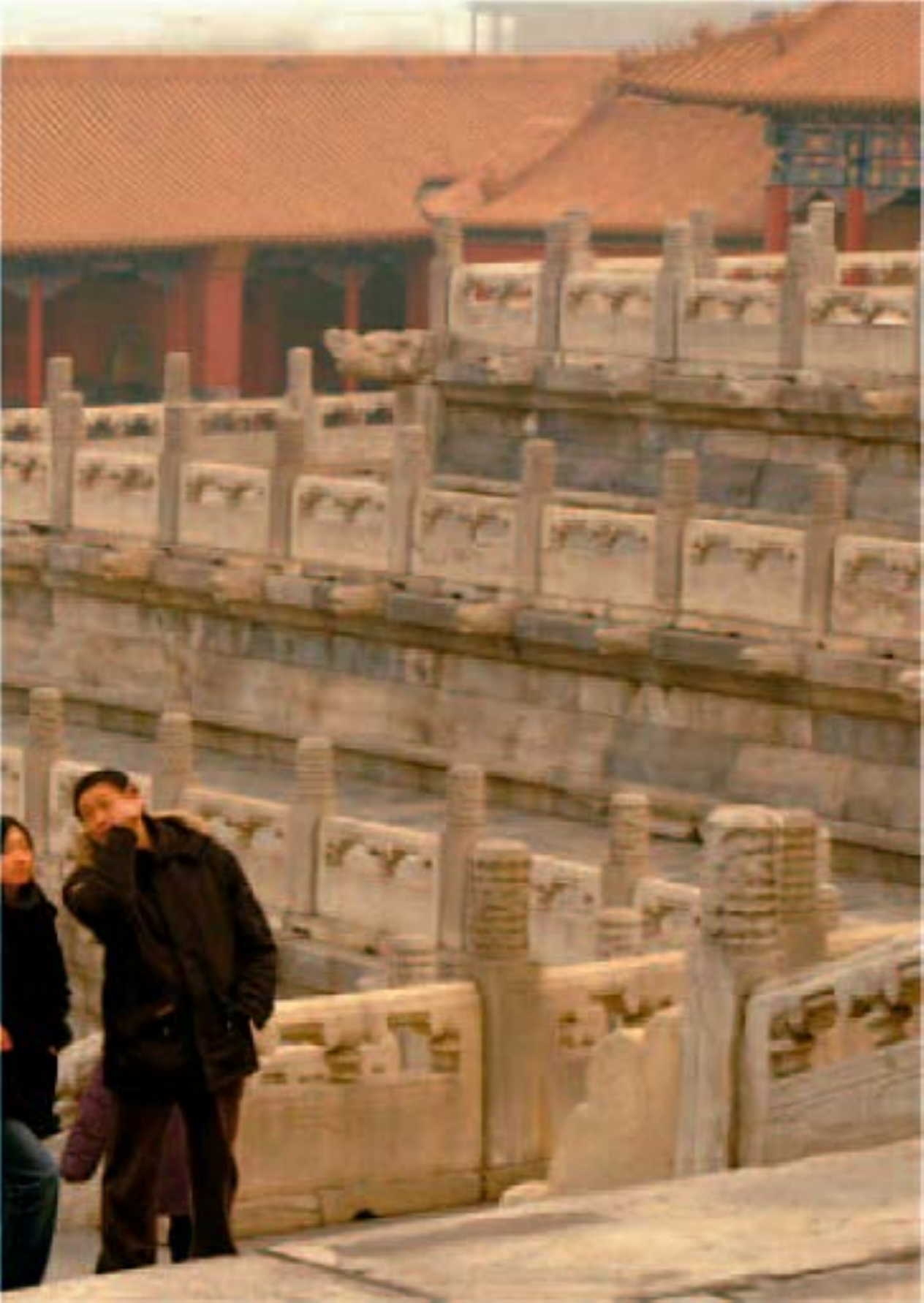} &
\includegraphics[width=0.1\textwidth]{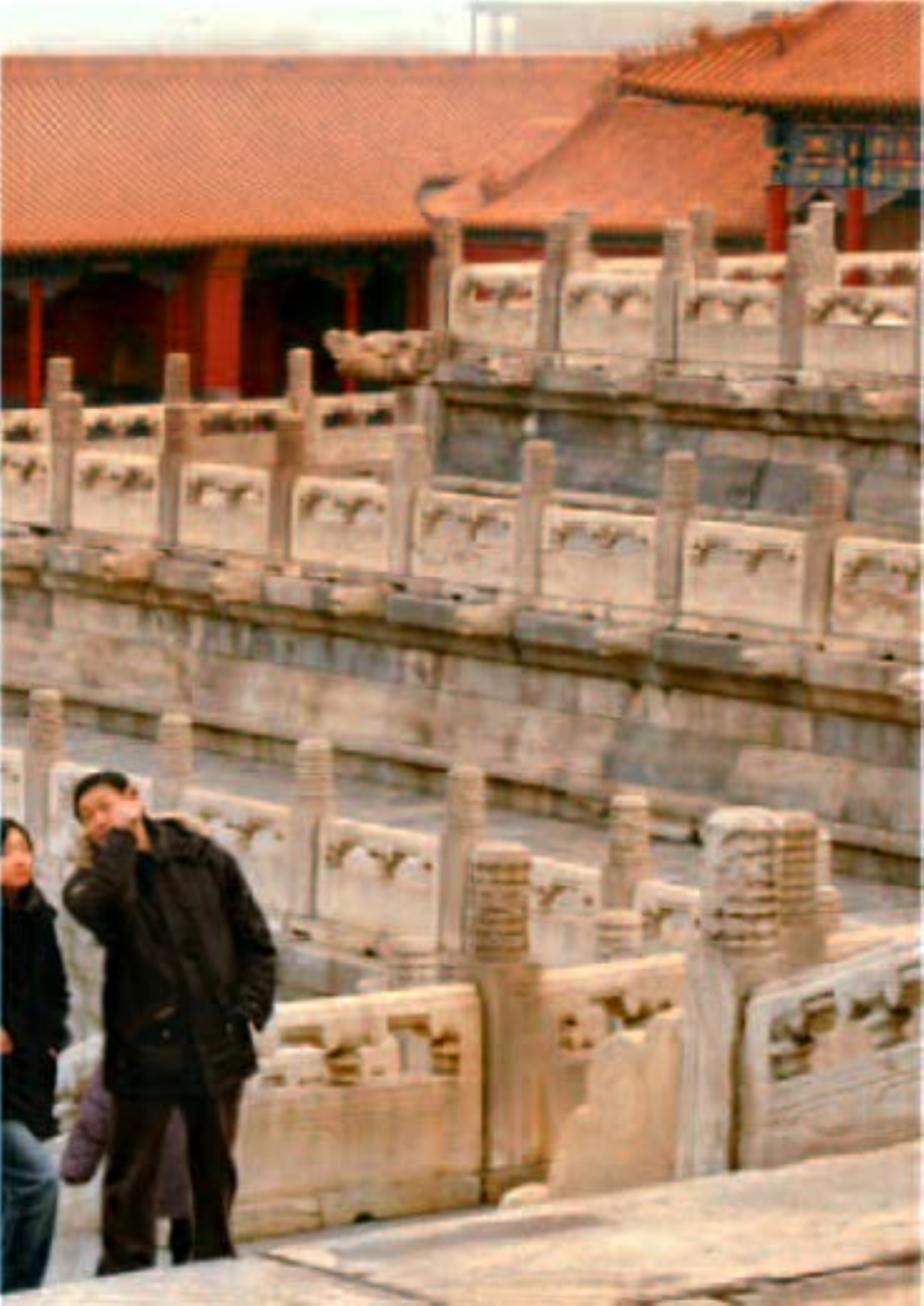} &
\includegraphics[width=0.1\textwidth]{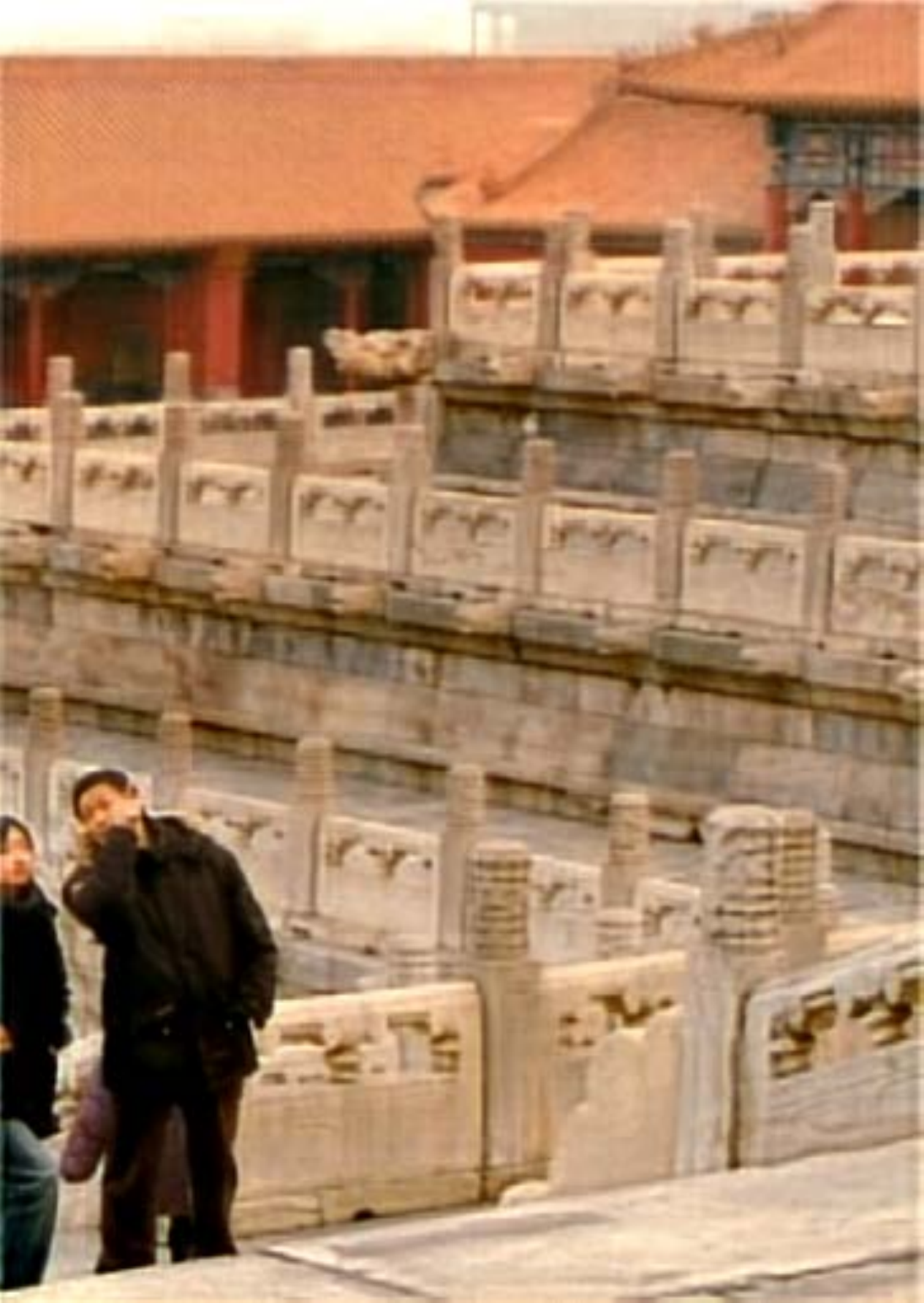} \\
\multirow{2}{*}{\begin{tabular}{c} Hazy \\ Image \end{tabular}} & DCP & CAP & NLD & DehazeNet & MSCNN & AOD-Net & GFN & PFF-Net \\
\end{tabular}
\end{center}
\caption{Comparisons with state-of-the-art methods on some real-world hazy images.}
\label{fig:realisticcomparisons}
\end{figure*}

Some qualitative comparisons on real-world hazy image are further shown in Figure~\ref{fig:realisticcomparisons}. These collected hazy images are at resolution around $500\times600$ pixels and captured from natural environment, best viewed on high-resolution display. As shown, the dehazed results from our method are clear and the details of the scenes are enhanced moderately better with natural perceptual qualities.

\section{Conclusion}
In this paper, we have proposed an effective trainable U-Net like end-to-end network for image dehazing. Progressive feature fusions are employed to learn input adaptive restoration model. Owing to the proposed U-Net like encoder-decoder architecture, our dehazing network has efficient memory usage and can directly recover ultra high definition hazed image up to 4K resolution. We evaluate our proposed network on two public dehazing benchmarks. The experimental results demonstrate that our network can achieve superior performance with great improvements when compared with several popular state-of-the-art methods.

\section*{Acknowledgment}
This work was supported by National Natural Science Foundation of China (Grant No. 61365002 and 61462045) and Provincial Natural Science Foundation of Jiangxi (Grant No. 20181BAB202013).
%
%
%
%

\end{document}